\documentclass{article}

\PassOptionsToPackage{numbers, compress}{natbib}

 \usepackage[main, final]{neurips_2026}

\usepackage[utf8]{inputenc} 
\usepackage[T1]{fontenc}    
\usepackage{xurl}
\usepackage{url}            
\usepackage{booktabs}       
\usepackage{amsfonts}       
\usepackage{nicefrac}       
\usepackage{microtype}      
\usepackage[table]{xcolor}          
\usepackage{amssymb}

\usepackage[most,skins]{tcolorbox}
\definecolor{rliableolive}{HTML}{BBCC33}
\definecolor{rliableblue}{HTML}{77AADD}
\definecolor{rliablered}{HTML}{EE8866}
\definecolor{LightCyan}{rgb}{0.88,1,1}
\definecolor{darkblue}{HTML}{2878D9}
\definecolor{navyblue}{HTML}{0000FF}
\definecolor{keycolor}{HTML}{B5605C}         
\definecolor{morandibluelight}{HTML}{E2ECF7} 
\definecolor{morandibluedeep}{HTML}{6E8AAD}  

\newtcolorbox{highlightbox}{
  enhanced,
  colback=morandibluelight,
  colframe=morandibluelight,
  boxrule=0pt, arc=0pt,
  left=10pt, right=8pt, top=6pt, bottom=6pt,
  borderline west={2pt}{0pt}{morandibluedeep}
}
\usepackage[mathscr]{eucal}
\usepackage{epsfig,epsf,psfrag}
\usepackage{amssymb,amsmath,amsthm,amsfonts,latexsym}
\usepackage{amsmath,graphicx,bm,xcolor,url,overpic}
\usepackage{array}
\usepackage{verbatim}
\usepackage{bm}
\usepackage{algorithmic}
\usepackage{algorithm}
\usepackage{enumitem}
\usepackage{verbatim}
\usepackage{textcomp}
\usepackage{mathrsfs}
\usepackage{hyperref}
\usepackage{epstopdf}
\usepackage{subfigure}
\usepackage{acronym}
\usepackage{diagbox}
\usepackage{tikz}
\usetikzlibrary{graphs, positioning, quotes, shapes.geometric}
\usepackage{thmtools, thm-restate}

\newcommand{\openone}{\leavevmode\hbox{\small1\normalsize\kern-.33em1}} 

\catcode`~=11 \def\UrlSpecials{\do\~{\kern -.15em\lower .7ex\hbox{~}\kern .04em}} \catcode`~=13 

\allowdisplaybreaks[4]

\newcommand{\calA}{\mathcal{A}}

\newcommand{\calD}{\mathcal{D}}

\newcommand{\calV}{\mathcal{V}}

\newcommand{\bbE}{\mathbb{E}}

\newcommand{\bbP}{\mathbb{P}}

\newcommand{\bbR}{\mathbb{R}}

\DeclareMathAlphabet{\mathbsf}{OT1}{cmss}{bx}{n}
\DeclareMathAlphabet{\mathssf}{OT1}{cmss}{m}{sl}

\DeclareSymbolFont{bsfletters}{OT1}{cmss}{bx}{n}  
\DeclareSymbolFont{ssfletters}{OT1}{cmss}{m}{n}
\DeclareMathSymbol{\bsfGamma}{0}{bsfletters}{'000}
\DeclareMathSymbol{\ssfGamma}{0}{ssfletters}{'000}
\DeclareMathSymbol{\bsfDelta}{0}{bsfletters}{'001}
\DeclareMathSymbol{\ssfDelta}{0}{ssfletters}{'001}
\DeclareMathSymbol{\bsfTheta}{0}{bsfletters}{'002}
\DeclareMathSymbol{\ssfTheta}{0}{ssfletters}{'002}
\DeclareMathSymbol{\bsfLambda}{0}{bsfletters}{'003}
\DeclareMathSymbol{\ssfLambda}{0}{ssfletters}{'003}
\DeclareMathSymbol{\bsfXi}{0}{bsfletters}{'004}
\DeclareMathSymbol{\ssfXi}{0}{ssfletters}{'004}
\DeclareMathSymbol{\bsfPi}{0}{bsfletters}{'005}
\DeclareMathSymbol{\ssfPi}{0}{ssfletters}{'005}
\DeclareMathSymbol{\bsfSigma}{0}{bsfletters}{'006}
\DeclareMathSymbol{\ssfSigma}{0}{ssfletters}{'006}
\DeclareMathSymbol{\bsfUpsilon}{0}{bsfletters}{'007}
\DeclareMathSymbol{\ssfUpsilon}{0}{ssfletters}{'007}
\DeclareMathSymbol{\bsfPhi}{0}{bsfletters}{'010}
\DeclareMathSymbol{\ssfPhi}{0}{ssfletters}{'010}
\DeclareMathSymbol{\bsfPsi}{0}{bsfletters}{'011}
\DeclareMathSymbol{\ssfPsi}{0}{ssfletters}{'011}
\DeclareMathSymbol{\bsfOmega}{0}{bsfletters}{'012}
\DeclareMathSymbol{\ssfOmega}{0}{ssfletters}{'012}

\DeclareMathOperator{\diag}{diag}

\DeclareMathOperator{\tr}{tr}

\DeclareMathOperator{\poly}{poly}

\def\##1\#{\begin{align}#1\end{align}}
\def\$#1\${\begin{align*}#1\end{align*}}

\def\BibTeX{{\rm B\kern-.05em{\sc i\kern-.025em b}\kern-.08em
    T\kern-.1667em\lower.7ex\hbox{E}\kern-.125emX}}
\usepackage{mathtools}
\usepackage{bbm}
\usepackage{multirow}
\usepackage{acronym}
\usepackage{makecell}

\usepackage{float}        
\usepackage{tabularx}     

\usepackage{hyperref}       
\usepackage{cleveref}

\acrodef{sft}[SFT]{Supervised Fine-Tuning}
\acrodef{rl}[RL]{Reinforcement Learning}
\acrodef{rft}[RFT]{Rejection Fine-Tuning}
\acrodef{ood}[OOD]{out-of-distribution}
\acrodef{llm}[LLM]{Large Language Model}
\acrodef{rlvr}[RLVR]{Reinforcement Learning with Verifiable Rewards}
\acrodef{grpo}[GRPO]{Group Relative Policy Optimization}
\acrodef{ppo}[PPO]{Proximal Policy Optimization}

\theoremstyle{plain}
\newtheorem{theorem}{Theorem}[section]

\newtheorem{lemma}[theorem]{Lemma}

\theoremstyle{definition}
\newtheorem{definition}[theorem]{Definition}
\newtheorem{assumption}[theorem]{Assumption}
\theoremstyle{remark}

\newcommand{\W}{W}
\newcommand{\Soft}{\mathtt{soft}}
\newcommand{\sft}{\mathrm{SFT}}
\newcommand{\rl}{\mathrm{RL}}
\newcommand{\ood}{\mathrm{OOD}}

\newcommand{\lm}{\mathrm{lm}}
\newcommand{\pt}{\mathrm{pt}}

\newcommand{\RM}[1]{\left(\romannumeral#1\right)}

\newcommand{\esb}{\mathcal{E}^{*}_{\mathrm{P}}}
\newcommand{\Lba}{L}
\newcommand{\Uba}{U}
\newcommand{\pit}{P_{\mathrm{in}}}
\newcommand{\cV}{\mathcal{V}}

\title{Rethinking ``RL Generalizes, SFT Memorizes'':\\ The Role of SFT Data}

\author{%
  \textbf{Yunlong~Hou}$^{1*}$ \quad
  \textbf{Fengzhuo~Zhang}$^{2*\dagger}$ \quad
  \textbf{Yuan Cheng}$^{1*}$ \\
  \textbf{Jiachun~Pan}$^{1}$ \quad
  \textbf{Xingyao~Li}$^{1}$ \quad
  \textbf{Zhuoran~Yang}$^{2,3}$ \\[0.5em]
  $^{1}$National University of Singapore, Singapore \\
  $^{2}$Yale University, USA, $^{3}$Cisco Foundation AI, USA
}

\begin{document}

\maketitle

\begin{abstract}

Large language models trained with Reinforcement Learning (RL) with verifiable rewards exhibit strong reasoning ability and broad generalization, whereas models trained with Supervised Fine-Tuning (SFT) are often viewed as more prone to memorization and limited transfer. This paper rethinks this distinction through the lens of SFT training data on widely used Qwen2.5 and Qwen3 series models. First, we study the role of data source and show that it is critical: a carefully mixed SFT dataset substantially outperforms data generated solely by a larger model. Second, we study the role of data scale and show that matching the number of correct rollouts between SFT and RL greatly improves SFT generalization, while matching the total rollout budget enables SFT to generalize as well as RL. Combining these two factors further enables SFT to generalize even better than RL. Third, by using LLM annotations to characterize
the solution methods in training rollouts, we show that larger datasets cover more tail methods and that these tail methods provide generalizable reasoning signals. Finally, we support these empirical findings theoretically by analyzing the training dynamics of shallow transformers under both RL and SFT.

\end{abstract}

\begin{NoHyper}
\begingroup
\let\thefootnote\relax
\footnotetext{$^{*}$Equal contribution. $^{\dagger}$Project lead.}
\endgroup
\end{NoHyper}

\section{Introduction}

\acp{llm} trained via \ac{rl} or \ac{sft} have achieved remarkable improvements in reasoning ability, as exemplified by the success of DeepSeek-R1~\citep{guo_deepseek-r1_2025} and s1~\citep{muennighoff2025s1}. A line of work compares the strengths and limitations of \ac{rl} and \ac{sft} for reasoning~\citep{chen2025sft,shenfeld2025rls,huan2026doesmath}. These studies suggest that ``\ac{rl} generalizes, whereas \ac{sft} memorizes''~\citep{chu2025sft,huan2026doesmath}, indicating that \ac{sft} often struggles to match \ac{rl} on reasoning tasks beyond the training domain.

However, it remains unclear whether the conclusion that ``\ac{rl} generalizes, \ac{sft} memorizes'' is drawn in a setting where the full potential of \ac{sft} is realized and \ac{sft} is given comparable data resources to \ac{rl}. 
For instance, the amount of training data for \ac{sft} is not aligned with that of \ac{rl} training.
In this work, we revisit this conclusion through the lens of the training dataset. Specifically, we study how data source and scale affect \ac{sft} generalization, and ask:
\begin{itemize}[leftmargin=2em]
    \item[1.] \emph{What kind of data source best unlocks the generalization potential of \ac{sft}?}
    \item[2.] \emph{How should we match the data consumed by \ac{sft} and \ac{rl} to compare their generalization in a fairer setting?}
\end{itemize}

To answer the first question, we perform \ac{sft} using training data from two sources and their mixtures, and evaluate the resulting models on general reasoning tasks to assess generalization. Specifically, we construct \ac{sft} datasets from rollouts of two models: the strong Qwen3-14B model and the \ac{rl}-tuned base model, pairing each question prompt with one correct rollout. Although both \ac{sft}-trained models generalize worse than the \ac{rl}-trained model, the \ac{rl}-tuned dataset yields much better generalization than the Qwen3-14B dataset. Moreover, interpolating between the two sources shows that adding an appropriate amount of more diverse data further improves general reasoning. These findings highlight the importance of data source for \ac{sft} generalization.

To answer the second question, we match the \ac{sft} training dataset size to that of \ac{rl} under different criteria. Rather than matching \ac{sft} and \ac{rl} training data prompt-wise, we scale the \ac{sft} dataset according to three criteria: (1) the number of solvable prompts, (2) the number of correct rollouts, and (3) the total number of rollouts. After training and evaluating models on these datasets, we find that matching the number of correct rollouts used in \ac{rl} greatly improves \ac{sft} generalization, while further matching the total number of rollouts enables \ac{sft} to achieve generalization comparable to \ac{rl}. Moreover, data source and data scale have a compounding effect: \ac{sft} with a mixture of the two data sources and a matched number of total rollouts achieves better generalization than \ac{rl}.

To further demystify the role of data scale, we study how scaling the \ac{sft} dataset affects the coverage of LLM-annotated solution methods in training rollouts and, in turn, general reasoning performance. Our analysis shows that larger datasets cover more tail methods in the long-tailed method distribution.
By further annotating evaluation rollouts on general reasoning tasks, we find that these tail methods are used more often and solve more questions as the dataset grows. These results suggest that data scaling improves the coverage of long-tailed reasoning patterns, thereby enhancing generalization.

Finally, we theoretically justify our insights through a training-dynamics analysis of shallow transformers. For long-tailed \ac{sft} data, we prove that both \ac{sft} and \ac{rl} achieve high accuracy on in-distribution problems. However, with insufficient method coverage, \ac{sft} generalizes worse than \ac{rl} on \ac{ood} problems. As more methods are covered, potentially through a larger data scale, \ac{sft} can match \ac{rl} generalization. These results justify the importance of data scale and tail-method coverage for \ac{sft} generalization.

In conclusion, our findings suggest that ``\ac{rl} generalizes, whereas \ac{sft} memorizes'' requires prerequisites on \ac{sft} training datasets. In particular, we show that both the \ac{sft} training data source and data scale are essential for unlocking the full potential of \ac{sft}. When trained with the same total number of samples as \ac{rl}, \ac{sft} with a appropriate mixture of data sources can achieve generalization comparable to, or even better than, that of \ac{rl}.

\section{Preliminaries}\label{sec:prelim}

\textbf{\ac{rlvr}} fine-tunes \acp{llm} by maximizing the expected reward of model responses, where the reward is automatically computed from verifiable outcomes~\citep{guo_deepseek-r1_2025}. Given a set of question prompts $\calD_{\mathrm{RL}}\subseteq \calV^*$, where $\calV^*$ denotes the set of all finite sequences over the alphabet $\calV$, the regularized \ac{rlvr} objective for model parameter $\theta\in\Theta$ is
\begin{align*}
    J_\mathrm{RL}(\theta)= \bbE_{X\sim\mathcal{D}_{\mathrm{RL}}, Y\sim\bbP_{\theta}(\cdot\mid X)}\big[r(X,Y)\big] - \beta \bbE_{X\sim\mathcal{D}_{\mathrm{RL}}, Y\sim\bbP_{\theta}(\cdot\mid X)}\Big[\log \big(\bbP_{\theta}(Y\mid X)/\bbP_{\textrm{ref}}(Y\mid X)\big)\Big].
\end{align*}
The first term is the expected reward of model rollouts $Y\sim\bbP_{\theta}(\cdot\mid X)$ on the question-prompt set, where $r(X,Y)$ evaluates the correctness of the response $Y$ to the question $X$, and $X\sim\calD_{\mathrm{RL}}$ means that $X$ is sampled uniformly from $\calD_{\mathrm{RL}}$. The second term regularizes the model $\bbP_{\theta}$ to remain close to the reference model $\bbP_{\mathrm{ref}}$, typically the pre-\ac{rlvr} model. Directly optimizing the model parameters with $J_{\mathrm{RL}}(\theta)$ often leads to poor performance due to large gradient variance and coverage issues~\citep{schulman2015trust,schulman2017proximal}.
Instead of maximizing $J_{\mathrm{RL}}$ directly, \ac{ppo}~\citep{schulman2017proximal} and its recent powerful variant \ac{grpo}~\citep{guo_deepseek-r1_2025} maximize the following surrogate objective at each step:
\begin{align*}
    J_{\mathrm{GRPO}}(\theta) =
    \mathbb{E}
    \bigg[
    \frac{1}{G} \sum_{i=1}^G \frac{1}{|Y_i|} \sum_{t=1}^{|Y_i|}
    \Big( 
    \min\big( w_{i,t}(\theta)A_i, \; \mathrm{clip}_{\varepsilon}(w_{i,t}(\theta)A_i \big)
    -\beta \textrm{KL}(\bbP_\theta \,\|\, \bbP_{\mathrm{ref}})
    \Big)
    \bigg],
\end{align*}
where the expectation is taken over questions $X\sim\calD_{\mathrm{RL}}$ and
$G$ rollouts $\{Y_i\}_{i=1}^G\sim\bbP_{\theta_{\mathrm{old}}}(\cdot\mid X)$.
Here, $
    w_{i,t}(\theta)
    =
    \bbP_\theta(Y_{i,t}\mid X,Y_{i,<t})/
    \bbP_{\theta_{\mathrm{old}}}(Y_{i,t}\mid X,Y_{i,<t})
$
is the importance-sampling ratio between the current model and the previous-step model,
$\mathrm{clip}_{\varepsilon}(x)$ clips $x$ to $[1-\varepsilon,1+\varepsilon]$,
and
$
    A_i
    =
    (r_i-\mathrm{mean}(\{r_j\}_{j\in[G]}))/
    \mathrm{std}(\{r_j\}_{j\in[G]})
$
is the group-normalized advantage of the $i$-th rollout. \ac{grpo} maximizes $J_{\mathrm{GRPO}}$ to update $\theta$, typically using Adam.
Empirically, \citet{guo_deepseek-r1_2025} show that \ac{grpo} substantially improves multi-step reasoning performance on mathematics, logical deduction, and scientific reasoning tasks. Subsequent studies further show that \ac{rl}-induced gains on the training domain, such as mathematics, can generalize to other reasoning domains, including physics and broader scientific topics~\citep{chu2025sft,huan2026doesmath}. See more related works in Appendix~\ref{app:relatedworks}.

\textbf{\ac{sft}} equips pretrained \acp{llm} with long-form reasoning abilities by directly minimizing the cross-entropy loss on curated long-reasoning data~\citep{muennighoff2025s1,wang2026survey}. Let $\calD_{\sft}$ denote a long-reasoning dataset consisting of pairs $(X,Y)$, where $X$ is a question prompt and $Y$ is the corresponding long-reasoning answer. The objective of long-reasoning \ac{sft} is $L_\sft(\theta)= -\mathbb{E}_{(X,Y)\sim\mathcal{D}_{\sft}}[ \sum_{t=1}^T\log\bbP_{\theta}(Y_t\mid X,Y_{<t})].$ In practice, this optimization is typically solved with first-order optimizers such as Adam~\citep{kingma2017adam} or AdamW~\citep{loshchilov2018decoupled}.
Because dense supervision is available at every token position, \ac{sft} is often sample-efficient~\citep{zhao2026selfdistilled}. However, prior work~\citep{chu2025sft,huan2026doesmath} suggests that \ac{sft} mainly improves model reasoning ability on tasks that are in distribution with respect to the training data, and that the resulting models exhibit weaker generalization than \ac{rlvr}-trained models.

\section{Experiments}\label{sec:exp}

In this section, we conduct extensive experiments to compare the generalization ability of \ac{rl} and \ac{sft}. Following \citet{huan2026doesmath}, we train models with \ac{sft} and \ac{rl} on mathematical reasoning datasets and evaluate them on other general reasoning tasks to assess out-of-domain generalization. To further understand what affects the generalization of \ac{sft} models, we study two critical properties of the training data: (1) its source and (2) the dataset scale. We first describe the basic training setups used throughout the experiments.

\textbf{Experimental Setup.} 
We use Qwen2.5-7B~\citep{qwen2025qwen25technicalreport} and Qwen3-8B-Base~\citep{yang2025qwen3} as the base models and train them on mathematical problems from the DAPO dataset~\citep{yu2026dapo}. We present the Qwen2.5-7B results in the main paper and defer the Qwen3-8B-Base results to Appendix~\ref{app:qwen3_8b_results}.
For \ac{rl}, we use the questions in the DAPO dataset with a reasoning prompt prefix, and optimize the model with \ac{grpo}~\citep{guo_deepseek-r1_2025} implemented in VeRL~\citep{Sheng2025Hybrid} under the standard training setup (see Appendix~\ref{app:rl_train_config}).
For \ac{sft}, we use rollout models, Qwen3-14B~\citep{yang2025qwen3technicalreport} by default, to generate training responses for the same DAPO prompts, and optimize the model with LlamaFactory~\citep{zheng2024llamafactory}; detailed hyperparameters are provided in Appendix~\ref{app:sft_train_config}. We evaluate the trained models on both in-distribution mathematical reasoning tasks and general reasoning tasks. 
For mathematical reasoning, we primarily use the SimpleRL evaluation pipeline~\citep{zeng2025simplerlzoo}, covering AIME24~\citep{aime_2024}, Minerva Math~\citep{lewkowycz2022solving}, GSM8K~\citep{cobbe2021training}, MATH500~\citep{hendrycksmath2021}, and OlympiadBench~\citep{he2024olympiadbench}. For general reasoning, we use twelve tasks from the Language Model Evaluation Harness~\citep{eval-harness}, including GPQA-Diamond~\citep{rein2024gpqa} and MMLU-Pro~\citep{wang2024mmlupro}, with the full list deferred to Appendix~\ref{app:exp_eval_detail}. These tasks span science, action planning, deep semantic understanding, multilingual reasoning, and related domains beyond mathematics. For each task, we normalize the metric to $[0,100]$ and report the average score across tasks.

\subsection{Training Data Source Matters for SFT Generalization}\label{sec:training_data_matters}
\begin{figure}[t]
    \centering
    \begin{minipage}[t]{0.49\textwidth}
        \centering
        \includegraphics[width=\linewidth]{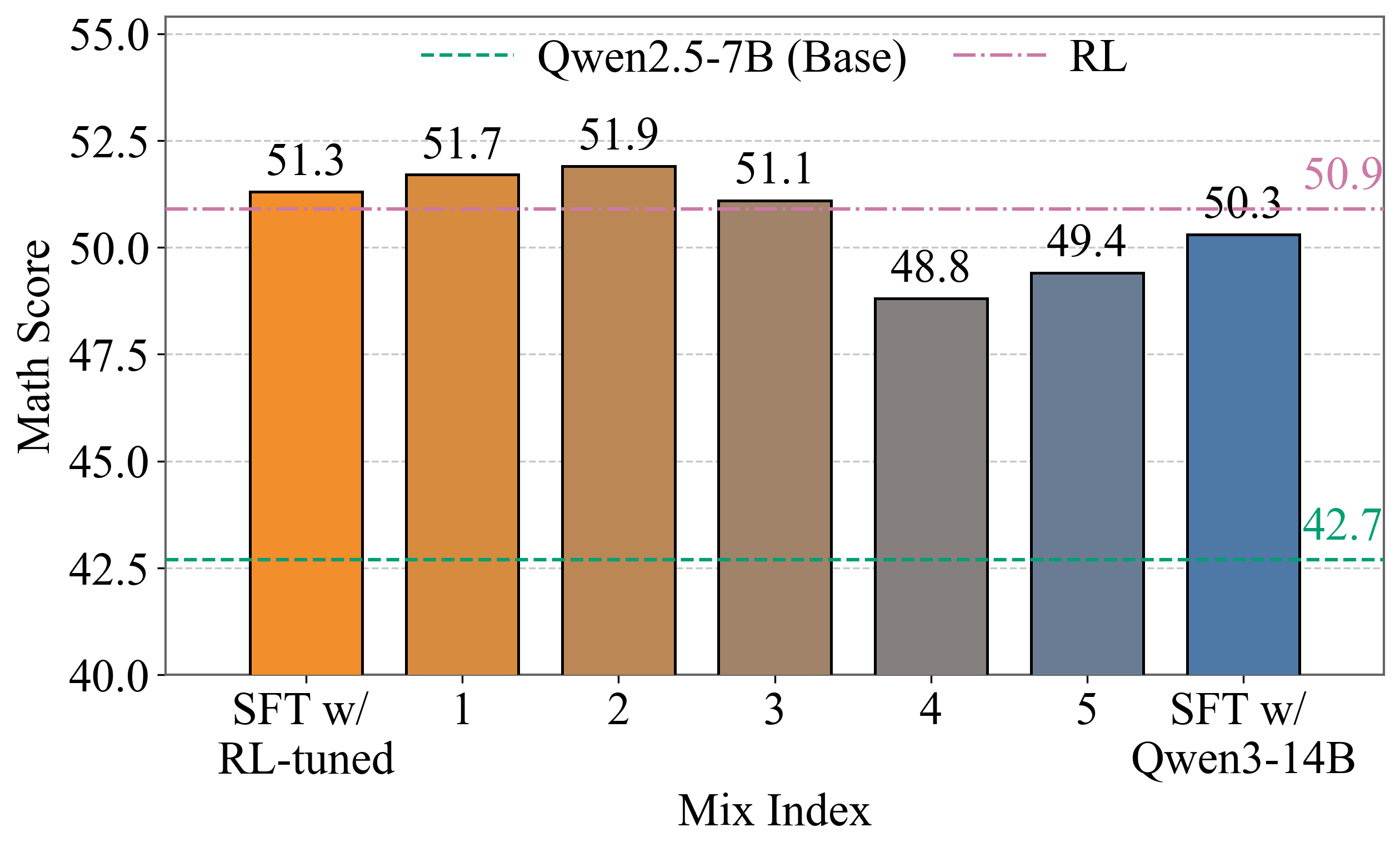}
        \caption*{(a) Math benchmarks}
    \end{minipage}
    \hfill
    \begin{minipage}[t]{0.49\textwidth}
        \centering
        \includegraphics[width=\linewidth]{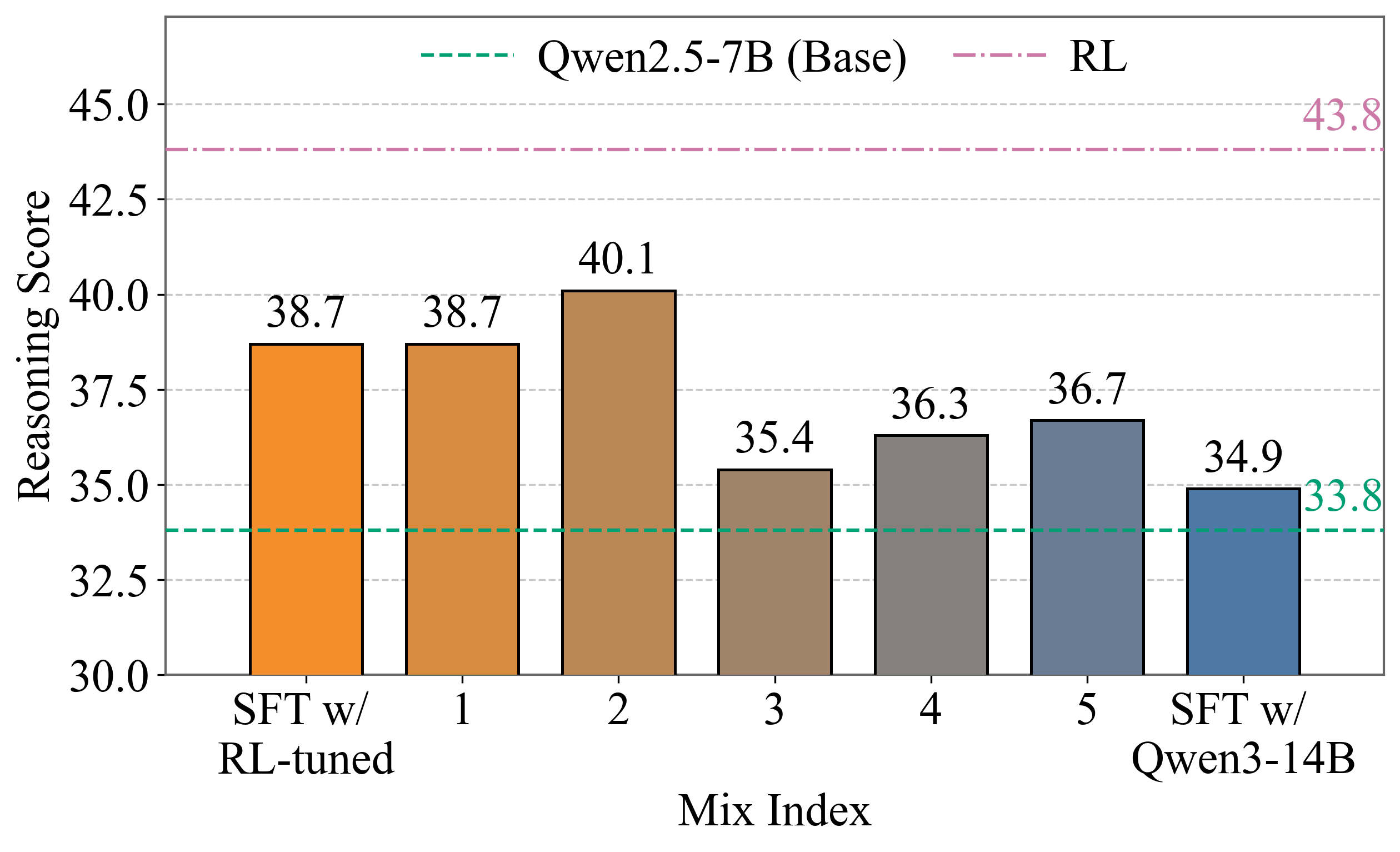}
        \caption*{(b) General reasoning benchmarks}
    \end{minipage}

    \caption{Performance comparison of \ac{sft}-trained models using datasets with different mixing ratios. Mix-$k$ consists of $(11-2k)\times 10\%$ data from the \ac{rl}-tuned dataset and $(2k-1)\times 10\%$ data from the Qwen3-14B dataset. The results report performance on both mathematical reasoning and broader reasoning tasks. The mixture with $70\%$ \ac{rl}-tuned data and $30\%$ Qwen3-14B data achieves the best performance across both tasks.}
    \vspace{-1em}
    \label{fig:plot_2_mix1}
\end{figure}

\begin{table}[htbp]
    \centering
    \scriptsize
    \setlength{\tabcolsep}{5pt}
    \begin{tabular}{l ccccc ccccc}
        \toprule
        & \multicolumn{5}{c}{\textbf{Math}} & \multicolumn{5}{c}{\textbf{General Reasoning}} \\
            \cmidrule(lr){2-6} \cmidrule(lr){7-11}
            \textbf{Model} & \makecell{AIME24\\(Avg@32)} & \makecell{Minerva\\Math} & GSM8k & MATH500 & \makecell{Olympiad\\Bench} & \makecell{ACP\\Bench} & \makecell{BBH} & C-Eval & \makecell{Ground\\Cocoa} & \makecell{MMLU\\-Pro} \\
            \midrule
            Qwen2.5-7B       & 4.7  & 26.5 & 88.6 & 63.6 & 30.2 & 23.3 & 10.9 & 40.1 & 34.7 & 9.2 \\
            RL-tuned         & 14.2 & 34.9 & 90.8 & 74.8 & 39.7 & 50.1 & 42.9 & 43.6 & 34.8 & 45.7 \\
            \midrule
            SFT w/ RL-tuned  & \underline{13.8} & 37.1 & 91.4 & 74.8 & \textbf{39.3} & \textbf{58.0} & 33.8 & 42.4 & 35.6 & \underline{19.4} \\
            SFT w/ Mix-1      & \textbf{13.9} & 37.5 & 91.3 & \textbf{76.6} & \underline{39.0} & \underline{56.6} & 34.4 & 36.4 & \underline{36.8} & 17.9 \\
            \rowcolor{blue!10}
            SFT w/ Mix-2      & 13.4 & \textbf{41.2} & \textbf{91.9} & \underline{76.2} & 36.9 & 53.4 & 37.3 & 47.8 & \textbf{37.0} & \textbf{21.3} \\
            SFT w/ Mix-3      & 12.4 & \underline{38.6} & \underline{91.6} & 75.0 & 37.8 & 50.7 & 35.6 & 28.2 & 36.7 & 3.8 \\
            SFT w/ Mix-4      & 10.7 & 31.2 & 90.8 & 73.8 & 37.6 & 43.7 & 34.1 & \underline{52.9} & 34.6 & 0.7 \\
            SFT w/ Mix-5      & 11.0 & 34.2 & 90.2 & 73.2 & 38.2 & 46.6 & \underline{37.7} & \textbf{53.8} & 34.2 & 1.2 \\
            SFT w/ Qwen3-14B & \textbf{13.9} & 33.5 & 90.5 & 75.0 & 38.5 & 45.3 & \textbf{42.1} & 36.5 & 33.2 & 1.1 \\
            \bottomrule
        \end{tabular}
        \vspace{1em}
        \caption{Evaluation results on math and selected general reasoning tasks of models trained from mixed data sources.
        The highest and second highest number within the \ac{sft}-tuned models are highlighted in \textbf{bold} and \underline{underlined}, respectively.
        SFT w/ Mix-2 achieves the best performance among the \ac{sft}-tuned models.}
        \label{tab:datasource}
\end{table}

In this section, we investigate how the training data source influences the generalization ability of \ac{sft}. We first introduce baselines and two data sources.

\textbf{Baselines and Data Sources.} Throughout the paper, we use the base model, i.e., the model without \ac{rl} or \ac{sft} tuning, and the model trained by \ac{rl} as baselines. We consider two data sources for \ac{sft}. The first is constructed by rejection sampling from a strong teacher model, Qwen3-14B~\citep{dong2023raft,yuan2023scaling,xiong2025minimalist}: for each question, we generate at most $128$ rollouts and retain one correct rollout if available. We refer to the resulting dataset, of size $15{,}607$, as the \emph{Qwen3-14B dataset}.
The second data source uses the model trained by \ac{rl} for
rejection sampling under the same protocol. Since the \ac{rl}-tuned model is obtained from the base model, its trajectories are expected to be more aligned with the base model parameters. We refer to this dataset, of size $14{,}512$, as the \emph{\ac{rl}-tuned dataset}. We highlight that both the Qwen3-14B and \ac{rl}-tuned datasets match the
\ac{rl} dataset in the number of prompts. 
To ablate the effect of data source more finely, we further interpolate between these two datasets with $5$ mixing ratios. The $k$-th mixed dataset, denoted by Mix-$k$, contains $(11-2k)\times 10\%$ data from the \ac{rl}-tuned dataset and $(2k-1)\times 10\%$ data from the Qwen3-14B dataset, where $k\in\{1,\ldots,5\}$.

Table~\ref{tab:datasource} reports results on all mathematical reasoning datasets and a representative subset of general reasoning datasets, with the full results deferred to Appendix~\ref{app:exp_results_data_source_exp}. 
Figure~\ref{fig:plot_2_mix1} shows the average performance over the mathematical and general reasoning datasets.

First, for the two baselines, our results show that \ac{rl} tuning improves the base model on all mathematical reasoning tasks: the average math score increases from $42.7$ to $50.9$ after \ac{rl} training on DAPO. Notably, this improvement also transfers to general reasoning tasks, where the average score increases from $33.8$ to $43.8$. 
Second, comparing \ac{sft} with Qwen3-14B and \ac{rl}-tuned data, Figure~\ref{fig:plot_2_mix1} shows that both data sources improve the base model's mathematical reasoning ability, reaching performance comparable to or even higher than the model trained by \ac{rl}. However, both \ac{sft} variants perform worse than the \ac{rl}-tuned model on general reasoning tasks, while \ac{sft} with the \ac{rl}-tuned dataset still outperforms \ac{sft} with the Qwen3-14B dataset. 
Finally, comparing different mixtures of Qwen3-14B and \ac{rl}-tuned data, Table~\ref{tab:datasource} and Figure~\ref{fig:plot_2_mix1} show that pure \ac{rl}-tuned data or strong model-generated data does not always yield the best general-reasoning performance. Instead, Mix-$2$, which consists of $70\%$ data from the \ac{rl}-tuned model and $30\%$ from Qwen3-14B, achieves the highest average score on general reasoning tasks. These lead to the following observation.
\begin{highlightbox}
\textbf{Observation 1:} 
The training data source is crucial for generalization: data from a model closer to the one being fine-tuned can be more effective than data from a stronger model, while data from a single model is not necessarily optimal; incorporating \textit{an appropriate amount of more diverse data} can further improve general reasoning ability.
\end{highlightbox}

\subsection{Matched Rollout Budgets Enable \ac{sft} to Generalize Like \ac{rl}}\label{sec:scale}

While the previous section shows that a suitable \ac{sft} data source improves general reasoning, the resulting model still lags behind the model trained by \ac{rl}. We next study an orthogonal factor: how \ac{sft} training-set size affects generalization.

In Section~\ref{sec:training_data_matters}, \ac{rl} and \ac{sft} are matched at the question-prompt level, as both use questions from the DAPO dataset. However, \ac{rl} generates $8$ rollouts per prompt, whereas \ac{sft} observes only one correct rollout per prompt. As a result, \ac{rl} trains on $8\times$ more rollouts than \ac{sft}. This motivates us to increase the \ac{sft} dataset size under different rollout-budget matching criteria.
\begin{itemize}[leftmargin=1em]
    \item We first match \textit{the number of solvable prompts}, defined as prompts for which at least one correct rollout is obtained during \ac{rl} training. Under this criterion, the corresponding Qwen3-14B and \ac{rl}-tuned datasets each contain one correct rollout for every solvable prompt. This yields a training set of about $10{,}000$ prompt--rollout pairs, fewer than the one used in Section~\ref{sec:training_data_matters}.
    \item We then match \textit{the number of correct rollouts} per prompt in the \ac{sft} training dataset to that observed during \ac{rl} training. Specifically, for each prompt, we record the number of correct rollouts generated during \ac{rl} and construct the \ac{sft} dataset with the same number of correct rollouts for that prompt. This yields a dataset with about $39{,}000$ prompt--rollout pairs.
    \item Lastly, we match \textit{the total number of rollouts} for each prompt. Since \ac{rl} training uses both correct and incorrect rollouts, we augment the $39{,}000$ correct training samples with about $98{,}000$ incorrect-rollouts. This criterion also matches the gradient computation cost of \ac{sft} to that of \ac{rl}. The resulting \ac{sft} datasets contain about $138{,}000$ prompt--rollout pairs.
\end{itemize}

\begin{figure}[t]
    \centering

    \begin{minipage}[t]{0.49\textwidth}
        \centering
        \includegraphics[width=\linewidth]{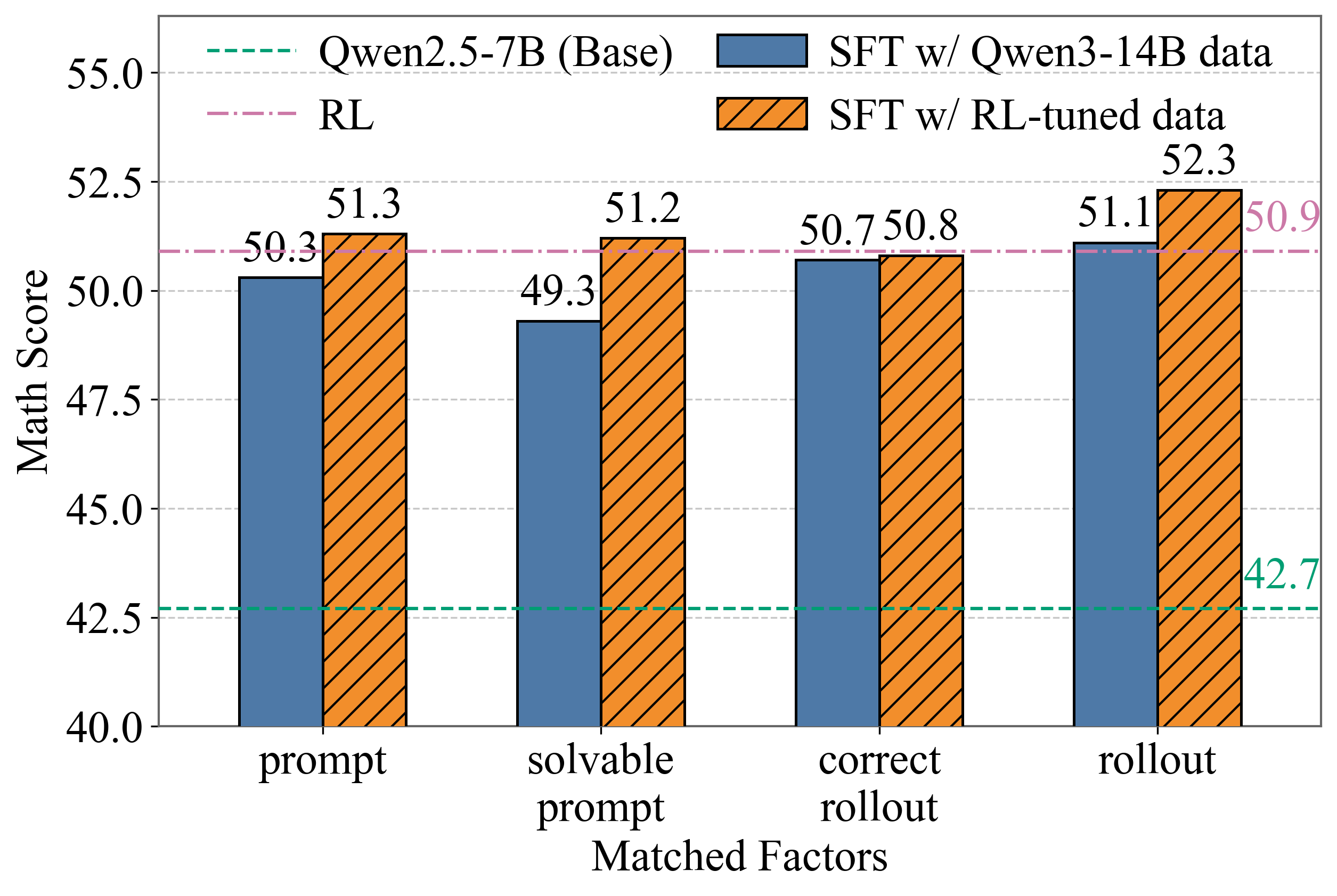}
        \caption*{(a) Math benchmarks}
    \end{minipage}
    \hfill
    \begin{minipage}[t]{0.49\textwidth}
        \centering
        \includegraphics[width=\linewidth]{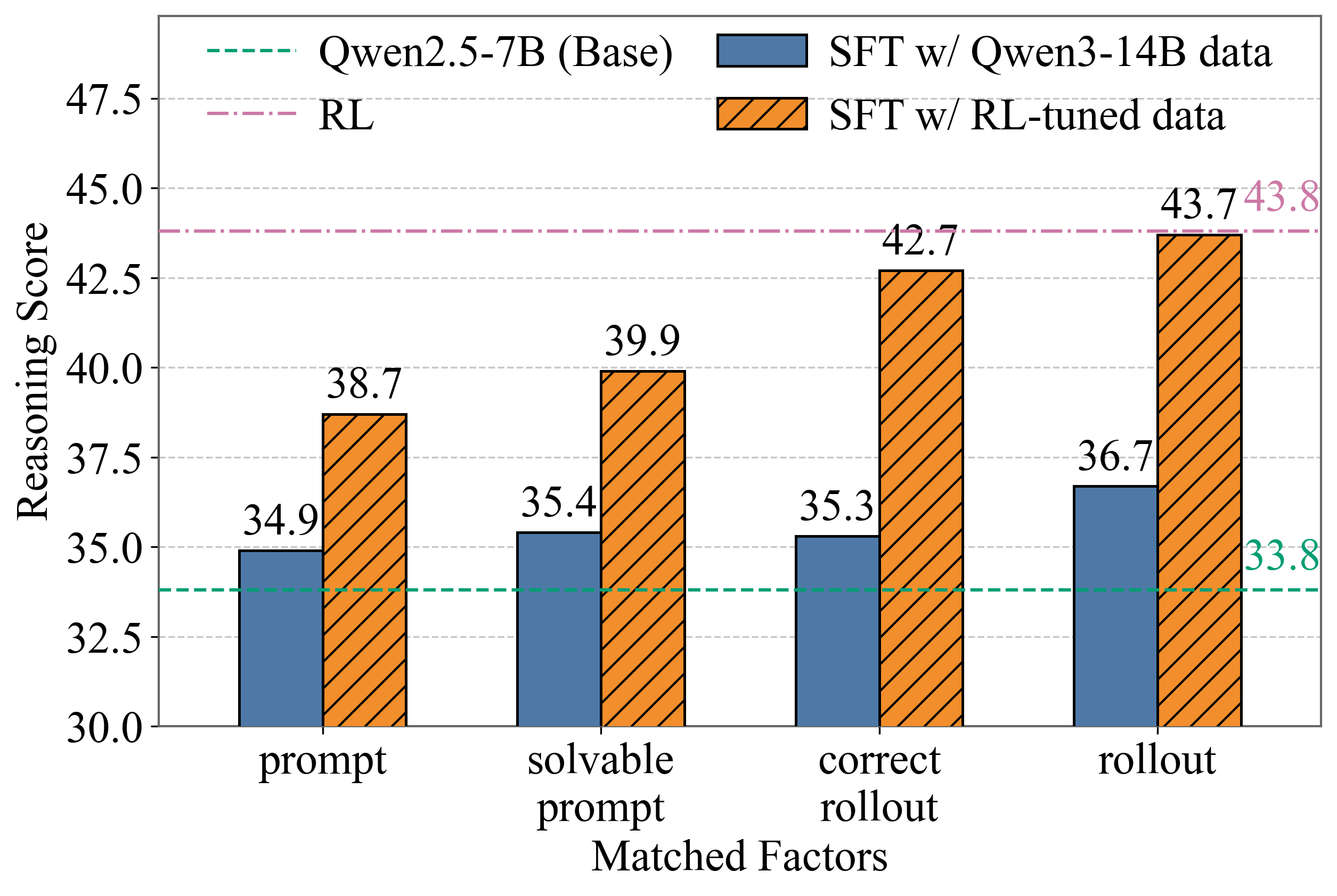}
        \caption*{(b) General reasoning benchmarks}
    \end{minipage}

    \caption{Performance comparison of \ac{sft}-trained models on mathematical and broader reasoning benchmarks when different training factors are matched to \ac{rl}, including the number of prompts, solvable prompts, correct rollouts, and total rollouts. Matching the total rollout budget of \ac{sft} to that of \ac{rl} substantially improves general reasoning performance, while preserving strong mathematical reasoning ability.
    }
    \label{fig:plot_1_exp1_4}
\end{figure}

More details of constructing these \ac{sft} datasets are defered to Appendix~\ref{app:sft_train_config}. Evaluation results are presented in Figure~\ref{fig:plot_1_exp1_4}.
First, comparing the prompt-matched and solvable-prompt-matched \ac{sft} datasets shows that the solvable-prompt-matched dataset achieves comparable mathematical reasoning ability but better general reasoning ability. One possible explanation is that forcing the model to learn from prompts that are not solvable during \ac{rl} training can hurt generalization. 
Second, comparing the prompt-matched and correct-rollout-matched datasets shows that matching the number of correct rollouts substantially improves \ac{sft} generalization, yielding general reasoning performance only slightly below that of the \ac{rl}-trained model.
Finally, comparing the correct-rollout-matched and rollout-matched datasets shows that including rollouts with incorrect final answers can further improve \ac{sft} generalization. This may be because such rollouts still contain correct intermediate reasoning steps that can benefit \ac{sft}~\citep{tian2026learningmistakesnegativereasoning}. Moreover, \ac{sft} with the rollout-matched dataset achieves comparable or better mathematical and general reasoning ability than the \ac{rl}-trained model. These results lead to the following observation.

\begin{highlightbox}
\textbf{Observation 2:} The amount of training data substantially influences \ac{sft} generalization:
matching the number of correct rollouts greatly improves \ac{sft}
generalization, while further matching the total rollout budget enables \ac{sft} to achieve comparable generalization to \ac{rl}.
\end{highlightbox}

We also conducted an ablation study to exclude the effect of the compute in Appendix~\ref{app:exp_ablation_compute}, which further supports our observations on the training-set size.

\textbf{Compounding Effects of Data Source and Scale.} Observations~1 and~2 show that choosing an appropriate data source and increasing the training-set size can each improve the generalization of \ac{sft}. We next investigate whether combining these two factors enables \ac{sft} to generalize even better than \ac{rl}.
Specifically, we mix the rollout-matched training dataset across different data sources. Since Section~\ref{sec:training_data_matters} suggests that the best mixture ratio is around $70\%$ from the \ac{rl}-tuned model and $30\%$ from Qwen3-14B, we focus on ratios near this setting to reduce computation. We consider mixture ratios $(11-2k)\times 10\%$ of the \ac{rl}-tuned dataset and $(2k-1)\times 10\%$ of the Qwen3-14B dataset for $k\in\{1,2,3\}$, yielding three training datasets, each with about $138{,}000$ prompt--rollout pairs.

\begin{figure}[t]
    \centering

    \begin{minipage}[t]{0.49\textwidth}
        \centering
        \includegraphics[width=\linewidth]{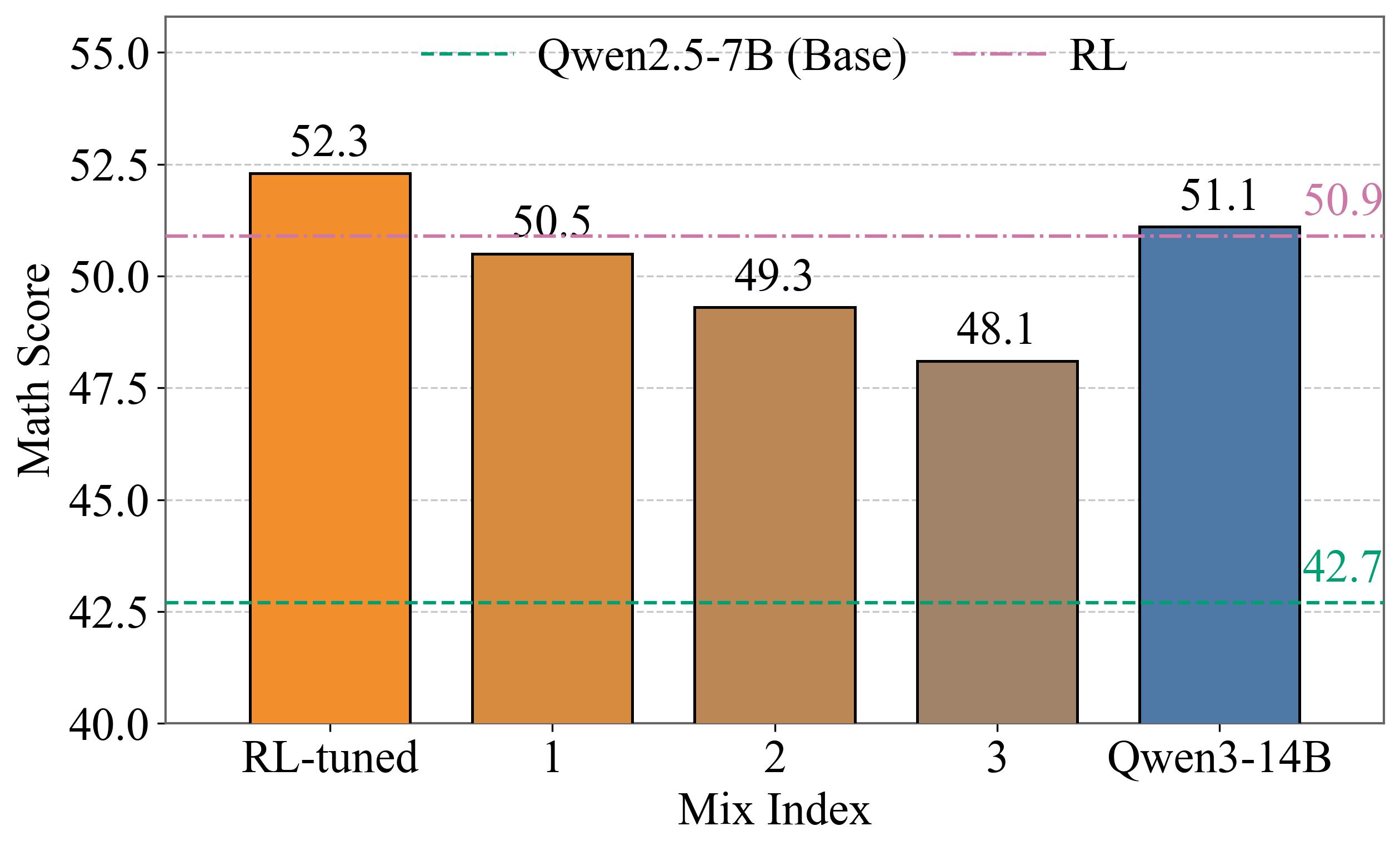}
        \caption*{(a) Math benchmarks}
    \end{minipage}
    \hfill
    \begin{minipage}[t]{0.49\textwidth}
        \centering
        \includegraphics[width=\linewidth]{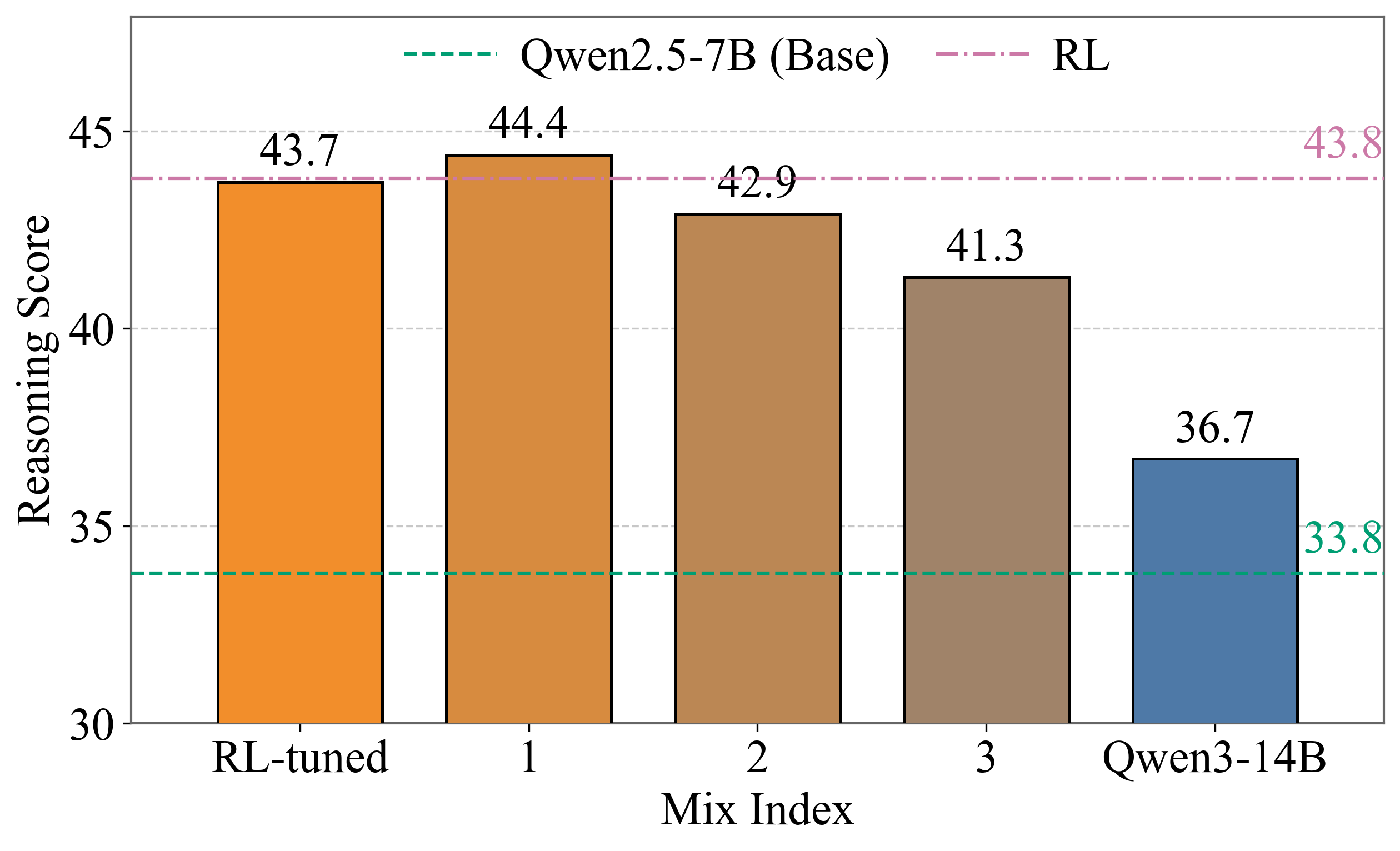}
        \caption*{(b) General reasoning benchmarks}
    \end{minipage}

    \caption{Performance comparison of \ac{sft}-trained models using scaled training datasets with different mixing ratios. The datasets are mixtures of \ac{rl}-tuned data and Qwen3-14B data. After scaling the training data, the mixture with $90\%$ \ac{rl}-tuned data and $10\%$ Qwen3-14B data achieves the best performance.
}
    \label{fig:plot_3_mix2}
\end{figure}

The experimental results are reported in Figure~\ref{fig:plot_3_mix2}, with the standard error reported in Appendix~\ref{app:with_std}. It shows that the best \ac{sft} performance is achieved by the dataset mixing $90\%$ \ac{rl}-tuned dataset with $10\%$ Qwen3-14B dataset, which yields better generalization than the \ac{rl}-trained model. 
Thus, we reach the following observation.

\begin{highlightbox}
\textbf{Observation 3:} With a carefully mixed dataset and a rollout budget matched to \ac{rl}, \ac{sft} achieves promisingly better general reasoning ability than \ac{rl}, undermining ``\ac{rl} generalizes, \ac{sft} memorizes.''
\end{highlightbox}

\subsection{Long-Tailed Rollouts Provide Generalizable Reasoning Signals}\label{sec:longtail}

Our experiments in Section~\ref{sec:scale} show that training-set size plays a critical role in enabling \ac{sft} to match \ac{rl} generalization. In this section, we investigate what additional information in larger datasets drives this effect. As the training set scales, it naturally includes more diverse rollouts, including rare tail rollouts that are less likely to appear in smaller datasets. We conjecture that these rare but informative rollouts are essential for \ac{sft} generalization, and we verify this hypothesis below.

We first characterize the rollout distribution by the \textit{methods} used to solve each math problem.
To study how dataset size changes the contained information, we construct training datasets with $k\in\{1,2,4,8\}$ rollouts per prompt, where $k=8$ matches the total rollout budget of \ac{rl}. We then prompt Qwen3-32B to annotate the methods used in each rollout, allowing multiple labels per rollout. After consolidating the annotations, we identify $138$ unique methods for solving math problems. More experimental details are in Appendix~\ref{app:rollout_analysis}. In the following, for robustness, we say that a method is \textit{covered} by a dataset if it appears at least $100$ times in that dataset.

\begin{figure}[t]
    \centering
    \subfigure[Distribution of method usage (top-50). ]{ \includegraphics[width=0.44\textwidth]{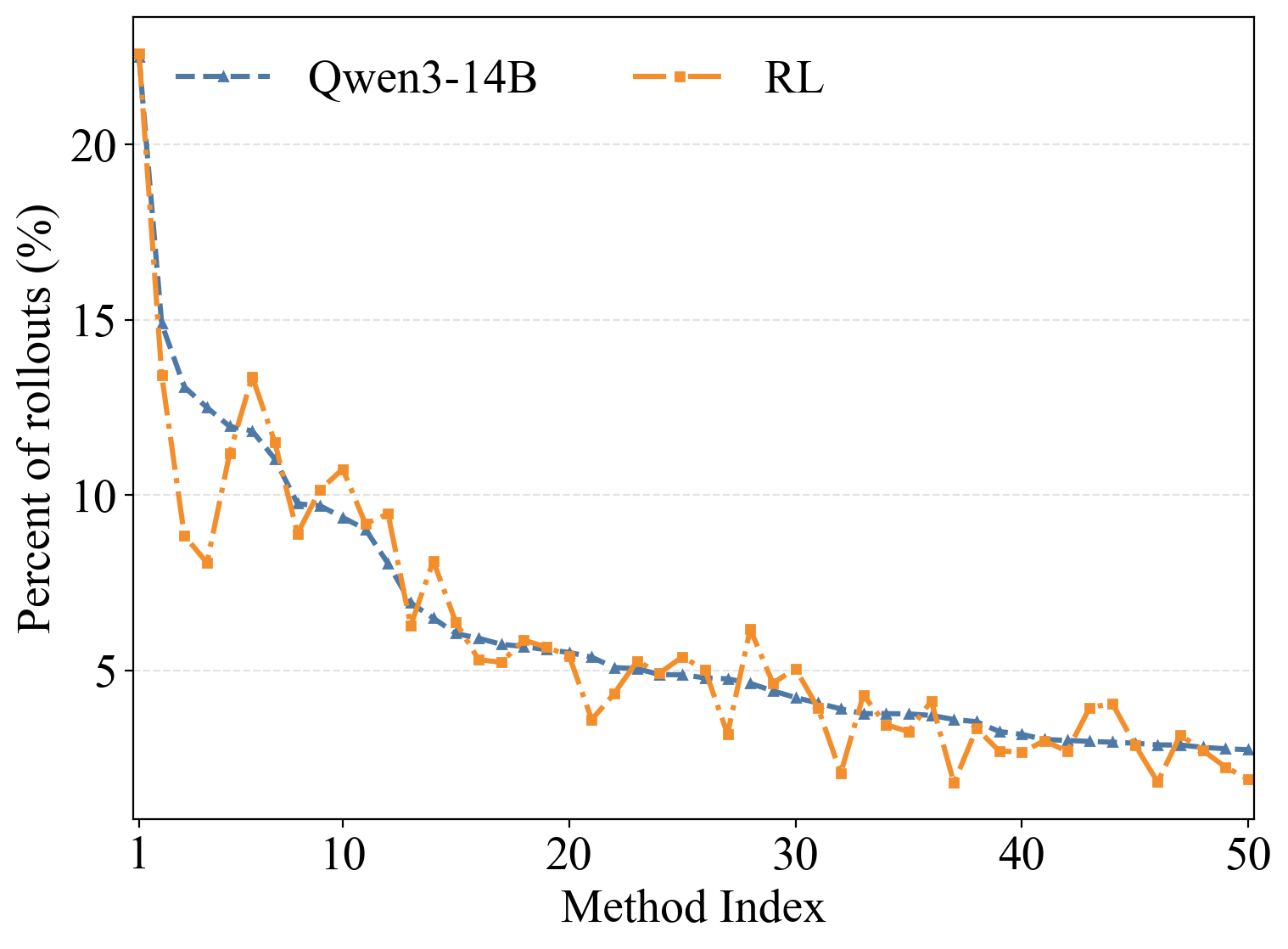}
    \label{fig:longtail}}
    \hfill
    \subfigure[Method coverage ]{ \includegraphics[width=0.52\textwidth]{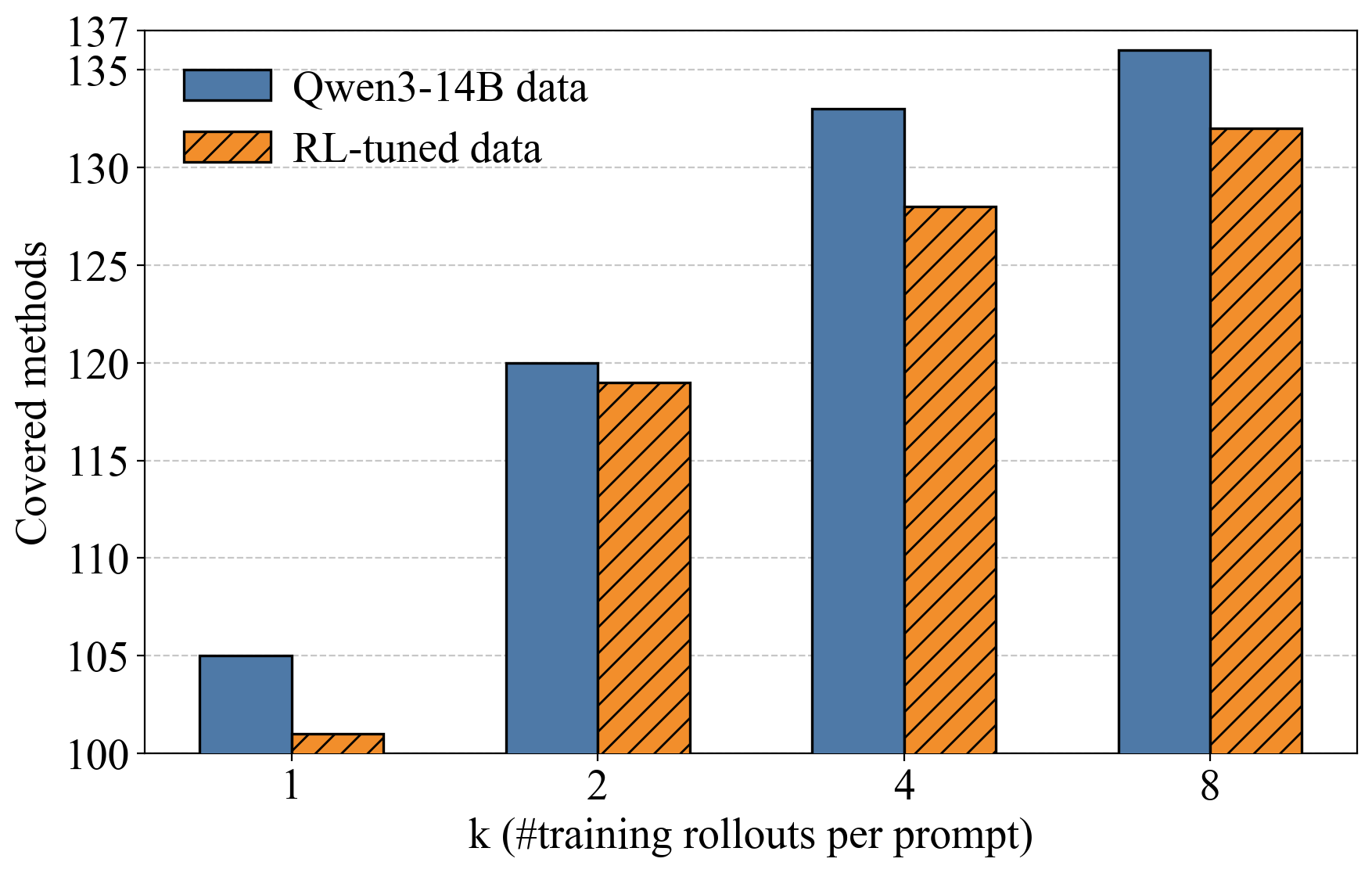}\label{fig:tailwithk}}
    
    \caption{Characterization of LLM-annotated method distribution and coverage. (a) plots the occurrence frequencies of solution methods in the \ac{rl}-tuned dataset, showing that the method distribution is long-tailed. (b) plots the number of covered methods in datasets of different sizes, showing that method coverage increases as the training data scale grows.
}
    \label{fig:covered_methods_training_dataset}
\end{figure}

Figure~\ref{fig:longtail} plots the top-50 method frequencies in the datasets with $k=8$, with methods ranked by their frequencies in the Qwen3-14B dataset. The figure reveals a clear long tail: many methods appear only infrequently. For each dataset, we define \textit{tail methods} as those in the bottom $20\%$ by frequency among all identified methods. Figure~\ref{fig:tailwithk} further plots the number of covered tail methods as $k$ varies. For both the Qwen3-14B and RL-tuned datasets, larger datasets cover more tail methods, exposing SFT models to a broader set of rare but potentially generalizable reasoning patterns.

\begin{figure}[t]
    \centering

    \begin{minipage}[t]{0.49\textwidth}
        \centering
        \includegraphics[width=\linewidth]{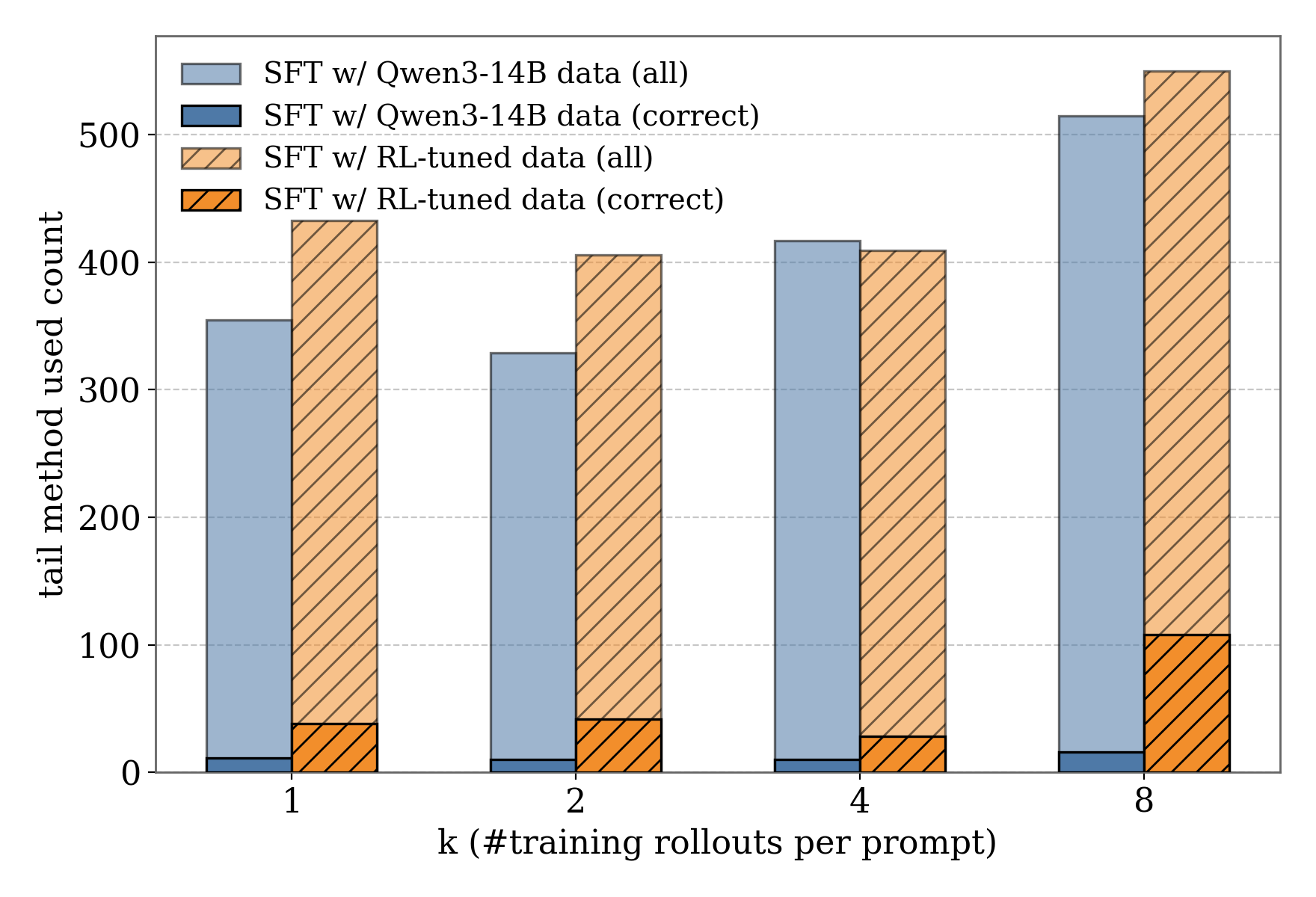}
        \caption{(a) MMLU-Pro}
    \end{minipage}
    \hfill
    \begin{minipage}[t]{0.49\textwidth}
        \centering
        \includegraphics[width=\linewidth]{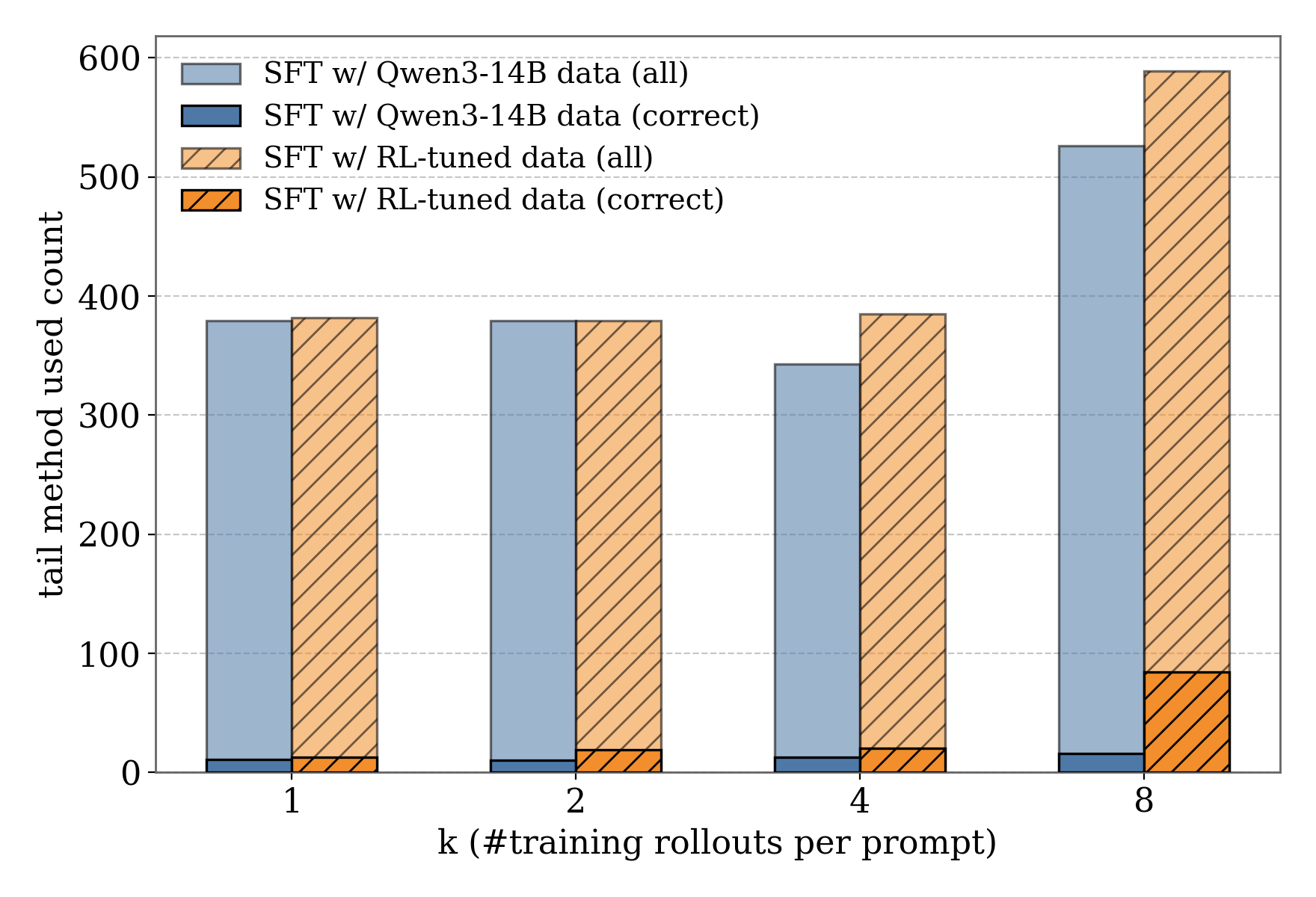}
        \caption{(b) MMLU-Pro+}
    \end{minipage}
    \caption{Tail-method usage statistics. The figures report the usage of LLM-annotated tail methods by \ac{sft}-trained models on MMLU-Pro and MMLU-Pro+. Tail methods are used more frequently (light bars) and can correctly solve more questions (dark bars) in general reasoning tasks (e.g., MMLU-Pro, MMLU-Pro+), as the training dataset encompasses a broader range of methods.}
    \label{fig:tail_helps}
\end{figure}

\paragraph{Tail Method Helps.}
To investigate how tail methods in the training dataset influence generalization to general reasoning tasks, we collect the reasoning rollouts during evaluation from models \ac{sft}-tuned on the four datasets. 
We then prompt Qwen3-32B to identify the methods employed in each rollout and compute the usage statistics on these tasks.
Figure~\ref{fig:tail_helps} shows the usage frequency of the tail methods on two representative tasks, MMLU-Pro~\citep{wang2024mmlupro} and MMLU-Pro+ (excluding math)~\citep{asgari2024mmlupro}, for the models trained on datasets with different $k$.
We report the usage counts both over all evaluation rollouts and over correct rollouts only.

We observe that tail methods are used more frequently as the training-set size scales.
Furthermore, these tail methods are capable of solving more questions as the tail-method coverage increases.
We arrive at the following observation.

\begin{highlightbox}
\textbf{Observation 4:} 
One benefit of increasing the training samples is improved coverage of long-tailed patterns. Training on these rare but informative samples can enhance performance on downstream general reasoning tasks.
\end{highlightbox}

We provide more rollout analysis in Appendix~\ref{app:rollout_analysis} and more discussions about our experimental setup, including the comparison with distillation and the use of incorrect rollouts, in Appendix~\ref{app:discussions}.

\section{A Case Study of Shallow Transformers }

Section~\ref{sec:exp} shows that SFT generalization depends critically on training-data scale, especially the coverage of long-tailed methods. We now provide theoretical grounding for these findings by analyzing the training dynamics of shallow transformers with a simplified data model.

\textbf{Data model.} We first introduce our simplified data model. The question dataset contains $L$ questions. For the $l$-th question, the prompt consists of $N+1$ token embeddings, $P=(X_1^\top,\ldots,X_N^\top,X_q^\top)^\top \in \mathbb R^{(N+1) \times d}$, with the first $N$ token denoted as $\pit=(X_1^\top,\ldots,X_N^\top)$. The row-wise last token embedding $X_q=u_l$ represents the query, where $\mathcal U=\{u_1,\ldots,u_L\}\subset \mathbb R^{1 \times d}$ denotes the candidate set of query embeddings and $d$ is the embedding dimension. The first $N$ tokens are sampled from two types of embeddings: informative embeddings $\mathcal I_l=\{i_{l,1},\ldots,i_{l,K}\}\subset \mathbb R^{1 \times d}$ and context embeddings $\mathcal C=\{c_1,\ldots,c_J\}\subset \mathbb R^{1\times d}$. The informative embeddings represent solution cues in the question, while the context embeddings represent connecting words. For the $l$-th question, each $X_i$ is sampled independently by first choosing
$\mathcal I_l$ or $\mathcal C$ with equal probability, and then sampling
uniformly from the selected set. Thus, prompts across all questions contain tokens from $\mathcal I=\cup_{l=1}^L \mathcal I_l$, $\mathcal C$, and $\mathcal U$. For simplicity, we assume that all embeddings in $\mathcal I\cup\mathcal C\cup\mathcal U$ are orthonormal, which is standard in previous theoretical work~\citep{huang2023context,nichani2024transformers,cheng2025transformers}. We denote by $\bbP_{\pt}$ the distribution over sampled questions and their
corresponding prompt embeddings, induced by the uniform distribution over
questions. Each question $l$ has $K$ correct answers, $\mathcal A_l=\{a_{l,1},\ldots,a_{l,K}\}\subset \mathbb R^{1\times (LK+1)}$, where $a_{l,k}$ corresponds to the informative token $i_{l,k}$. Here, $a_{l,k}$ represents both the final answer and the solving steps implied by the solution cue in $i_{l,k}$. Let $\mathcal A=\cup_{l=1}^L\mathcal A_l$ denote the set of all correct answers. The full answer space is $\bar{\mathcal A}=\mathcal A\cup\{\tilde a\}$, where $\tilde a$ represents all possible wrong answers. Similar to prompt embeddings, we assume that the embeddings in $\bar{\mathcal A}$ are orthonormal, and choose the one-hot embedding of answers as $a_{l,k}=e_{(l-1)K+k}$ and $\tilde a = e_{LK+1}$.

Further, we assume that answers $A$ presented in the \ac{sft} training dataset are long-tailedly distributed.
\begin{assumption}[Long-Tailed Distribution in \ac{sft} Data]\label{Assp: SFT Data} 
For each $l\in[L]$, we index the candidate answers so that $a_{l,1}$ is the unique head answer and $\{a_{l,2},\ldots,a_{l,K}\}$ are tail answers. Let $\mathbb{P}_{\sft}(A=a_{l,k}\mid X_q=u_l)=p_{l,k}$ be the conditional answer distribution of the \ac{sft} data. We assume that, for every $l\in[L]$, $p_{l,1}=\Theta(1)$ and $p_{l,k}=o(1)$ for $k\neq 1$.
\end{assumption}
This assumption captures long-tailedness under mild requirements. Specifically, we only require the head probability to be $\Theta(1)$ and the tail probabilities to be $o(1)$, mimicking the empirical pattern in Figure~\ref{fig:longtail}.

\textbf{Model architecture.} We next introduce the model structure. Specifically, we consider the model with a one-layer attention module:
$F(\pit,X_q;\W) = \Soft(\Soft(X_q \W \pit^\top)\pit W_{\lm}) \in \mathbb{R}^{1\times {(LK+1)}}$, where $\Soft(\cdot)$ denotes the softmax operator applied to a row vector. The inner softmax gives the attention weights, with the query embedding $X_q$ as the query and the prompt embeddings $P$ as both keys and values. The parameter
$\W\in\bbR^{d\times d}$ is optimized to capture the correlation between the query and the keys. The attention output is then mapped by the language-model head $W_{\lm}\in\bbR^{d\times (LK+1)}$ to answer-token logits, and the outer softmax converts these logits into answer probabilities. We assume that $W_{\lm}$ is fixed and inherited from prior knowledge in the pretrained model. Specifically, $W_{\lm}$ stores associative memories between informative tokens and their corresponding answers, as commonly observed in pretrained models~\citep{fang2024alphaedit,meng2022locating,meng2022mass}. We model this as
$$W_{\lm}=c_{\lm}\log(K)\big(\sum_{l=1}^L \sum_{k=1}^K i_{l,k}^\top (a_{l,k}-\sum_{l'\neq l}\sum_{k'=1}^Ka_{l',k'}-\widetilde a )+\sum_{j=1}^J c_j^\top\widetilde a \big)\in \mathbb{R}^{d \times (LK+1)},\text{where}\; c_{\lm} > 0.$$ 
This construction encodes one-to-one associations between informative tokens and correct answers, while associating context tokens with the incorrect answer $\widetilde a$. We index the $k$-th answer of the $l$-th question in $\bar{\calA}$ by the global index $(l-1)K+k$. For simplicity, when there is no ambiguity, we omit the dependence on $P$, $X_q$, and $\W$, and write $F_{l,k}$, or equivalently $F_{(l-1)K+k}$, for the predicted probability of this answer in $\bar{\calA}$.

\textbf{Training recipes.} \ac{sft} and \ac{rl} update the attention parameter $\W$ according to their respective training recipes. Both start from zero initialization, i.e., $\W_\sft^{(0)}=\W_\rl^{(0)}=0_{d\times d}$. \ac{sft} optimizes $\W$ with the following loss:
$$
L_\mathrm{SFT}(W)= -\sum_{l=1}^{L}\sum_{k=1}^{K}\mathbb{E}_{\pit,X_q,A\sim\mathcal{D}_{\mathrm{SFT}}}\left[\mathbbm{1}{\{X_q=u_l,A=a_{l,k}\}}\log(F(\pit,X_q;\W)_{(l-1)K+k})\right].
$$
The expectation is taken with respect to the joint distribution induced by the prompt-embedding distribution $\bbP_{\pt}$ and the conditional answer distribution in Assumption~\ref{Assp: SFT Data}. With learning rate $\eta_{\sft}>0$, \ac{sft} updates the parameter by
$\W_{\sft}^{(t+1)}=\W_{\sft}^{(t)}-\eta_{\sft}
\nabla_{\W}L_{\sft}(\W_{\sft}^{(t)})$. 
As introduced in Section~\ref{sec:prelim}, \ac{rl} updates the model by maximizing the expected reward of model rollouts. For theoretical simplicity, we let \ac{rl} directly optimize the unregularized expected reward, i.e.,
$$
J_\mathrm{RL}(W)= \sum_{l=1}^{L}\sum_{k=1}^K\mathbb{E}_{\pit,X_q}\bigg[\mathbbm{1}\{X_q=u_l\}F(\pit,X_q;\W)_{(l-1)K+k}\bigg].
$$
For each question prompt $P$, the reward is defined as the total probability mass assigned to all correct answers, i.e., $\sum_{k=1}^K F(\pit,X_q;\W)_k$. The parameter is then optimized by gradient ascent: $\W_{\rl}^{(t+1)}=\W_{\rl}^{(t)}+\eta_{\rl}\nabla_{\W}L_{\rl}(\W_{\rl}^{(t)})$. 
For simplicity, we consider only a one-step \ac{sft} and \ac{rl} update, which suffices for the shallow model to reach optimality. This simplification is standard in the learning-dynamics literature~\citep{huang2026learning,wang2025muon,ba2022high}.

\begin{theorem}[In-Distribution Convergence of SFT and RL]\label{thm: indist}
Consider the regime $N \gg \mathrm{poly}(K) \gg \mathrm{polylog}(N)$, $N \gg d$, $K\gg L$. Let $\epsilon=\log(K)/K>0$ denote the error threshold. Then the following results hold with probability at least $1-\exp(-\Omega(\poly(K)))$. 
\begin{itemize}[leftmargin=1em]
    \item Under Assumption~\ref{Assp: SFT Data}, when \ac{sft} tunes the parameter with a learning rate large enough $\eta_\sft = \Theta(LK/p_{l,1})$, after one step gradient update, the learned model achieves high accuracy on the training dataset. Specifically, for the $l$-th question, the wrong answer probability of the learned model is upper-bounded as follows: 
    $
        1-\sum_{k=1}^K F_{(l-1)K+k} \leq \epsilon. 
    $
    \item When \ac{rl} tunes the parameter with learning rates $\eta_{\rl}
    =
    \Theta(\poly(K,\log(1/\epsilon)))$, after one step gradient update, the learned model achieves high accuracy on the training dataset, i.e., for the $l$-th question, the model learns all the correct answers equally, i.e., 
    $
    |F_{(l-1)K+k}-1/K|
    \le
    \epsilon/K,\,
    \forall k\in[K]$. The wrong answer probability is also upper bounded by $\epsilon$.
\end{itemize}
\end{theorem}
These results theoretically show that both \ac{sft} and \ac{rl} can achieve high accuracy on in-distribution problems, i.e., problems covered by the training dataset. This is consistent with our experimental results in Section~\ref{sec:exp}, where \ac{sft} achieves mathematical reasoning ability comparable to \ac{rl} when the two methods are matched by the number of question prompts. This finding is also consistent with \citet{chu2025sft,lu2026does}. In addition, we also show that \ac{rl}-trained models explore and learn a balanced distribution across answers, which will support their generalization ability. 

To assess the generalization ability of models learned by \ac{sft} and \ac{rl}, we evaluate them on \ac{ood} prompts. For each query $X_q=u_l$, an \ac{ood} prompt, denoted by $P_{l,k}^{\ood}$, contains a single informative token type $i_{l,k}$ repeated $N/2$ times, with the remaining $N/2$ positions filled by context tokens, where $k\in[K]$. Thus, $a_{l,k}$ is the unique correct answer for this prompt. For each $l\in[L]$, the \ac{ood} test distribution mixes over $k\in[K]$ with weights $\{\rho_{l,k}\}_{k=1}^K$, where $\sum_{k=1}^K\rho_{l,k}=1$ and $\rho_{l,k}\geq0$. We evaluate a model by its probability of outputting the correct answer:
$
\mathrm{Eval}(\W)
=
L^{-1}\sum_{l=1}^L\sum_{k=1}^K
\rho_{l,k}F(P_{l,k}^{\ood},u_l;\W)_{(l-1)K+k}.
$

\begin{theorem}[\ac{ood} Generalization]\label{Thm: ood generalization}
Denote the trained weights of SFT and RL as \(W_\sft\) and \(W_\rl\), respectively. Let $\mathrm{Eval}_\sft=\mathrm{Eval}(W_\sft),\mathrm{Eval}_\rl=\mathrm{Eval}(W_\rl).$
Then we have:
\begin{itemize}[leftmargin=1em]
    \item (RL always generalizes.)
    For any \ac{ood} mixture \(\{\rho_{l,k}\}_{k=1}^K\), we have $1-\mathrm{Eval}_\rl
        =O(K^{-3}).$
    \item (SFT generalizes on covered methods.)
    If the \ac{ood} mixture distribution is supported on answers whose training probabilities satisfy $p_{l,k}\geq 1/\log K$, then $1-\mathrm{Eval}_{\sft}=O(K^{-3})$. In contrast, if it is supported on answers whose training probabilities satisfy $p_{l,k}\leq 1/K$, then
$\mathrm{Eval}_{\sft}\leq 1/2$.
\end{itemize}
\end{theorem}
First, we show that \ac{rl} generalizes to \ac{ood} prompts, which serve as a stylized analogue of the general reasoning tasks in our experiments. This is consistent with the strong generalization of \ac{rl} observed in prior work~\citep{chu2025sft}. Second, we show that \ac{sft} generalizes to well-covered methods, characterized by $p_{l,k}\geq 1/\log K$. This explains Observations~2 and~4: increasing the data scale covers more methods and raises the empirical frequencies of tail methods in the dataset, as shown in Figure~\ref{fig:tail_helps}.
We provide more discussions on the connection between the empirical findings and the theoretical results in Appendix~\ref{app:inter_theory}
\section{Conclusion}

This work rethinks the claim that ``\ac{rl} generalizes, whereas \ac{sft} memorizes'' from the perspective of \ac{sft} training data. Specifically, we show that both data source and data scale substantially influence \ac{sft} generalization. With a carefully mixed dataset and a rollout budget matched to \ac{rl}, \ac{sft} can even generalize better than \ac{rl}. We further show that this improvement is supported by increased coverage of tail methods, and provide theoretical analysis to justify these insights. A limitation of our work is that we focus on \acp{llm}; extending our analysis to other generative models, such as diffusion models, is left for future work.
\section*{Acknowledgements and Disclosure}
 Zhuoran Yang gratefully acknowledges financial support from UK DSIT (AP-S2-100058) and NSF (DMS 2413243).

\bibliographystyle{unsrtnat}
\bibliography{reference}

\appendix
\section{Related Works}\label{app:relatedworks}

\textbf{Reinforcement Learning with Verifiable Rewards} has substantially improved the reasoning ability of \acp{llm}, as exemplified by the success of DeepSeek-R1~\citep{guo_deepseek-r1_2025}. A central component of this training pipeline is \ac{grpo}~\citep{shao2024deepseekmath}, a group-based policy optimization method that normalizes advantages by the within-group standard deviation. Beyond the original DeepSeek-R1 results~\citep{guo_deepseek-r1_2025}, the effectiveness of \ac{grpo} has been reproduced across models of different scales~\citep{zeng2025simplerlzoo,deepscaler2025,he2025justrlscaling15bllm}. Subsequent works improve \ac{grpo} from several perspectives, including bias correction~\citep{liu2025understanding,yang2026grouprelativeadvantagebiased,yu2026dapo}, sample selection~\citep{xiong2025reinforceadaadaptivesamplingframework,yu2026dapo}, reward design~\citep{tuyls2026representationbased,zhu2026thenegative}, and entropy regularization~\citep{wang2026beyond}. A parallel line of work studies the mechanisms and limits of vanilla \ac{grpo} and \ac{rlvr}. One central question is whether RL-tuned models acquire genuinely new skills or mainly refine existing reasoning trajectories. \citet{yue2026does} argues that \ac{rlvr} sharpens correct reasoning trajectories on math and coding tasks without expanding the model's skill boundary. In contrast, \citet{wen2025reinforcementlearningverifiablerewards} show that, under CoT-\textrm{Pass}@K, which evaluates both final answers and intermediate reasoning steps, \ac{rlvr} can extend the reasoning ability of the base model. Relatedly, \citet{sun2026rl} find that \ac{rlvr}-trained models can solve tasks requiring recombination of learned skills but remain weak on transformative tasks, while \citet{liu2026prorl} show that prolonged \ac{rl} training can uncover reasoning strategies inaccessible to the base model. Other works investigate the generalization of \ac{rlvr} beyond the training topics, with evidence that RL-tuned models can transfer learned abilities to unseen domains~\citep{huan2026doesmath,fu2025srft,chu2025sft}. Finally, \citet{tan2026scalingbehaviorsllmreinforcement} studies the scaling behavior of \ac{grpo} on math tasks, while \citet{cui2025entropymechanismreinforcementlearning,shao2026spuriousrewardsrethinkingtraining,chen2026exploration} show that reasoning performance can improve even without verifiable rewards, for example, through random rewards or entropy minimization under suitable conditions. See \citet{wang2026survey,li2025system,xu2025toward} for the comprehensive surveys. Our work focuses on the generalization ability of \ac{rl} and its comparison with \ac{sft}. Specifically, we use the generalization ability of \ac{rl} as a baseline for evaluating \ac{sft}.

\paragraph{Long-Reasoning \ac{sft}} \hspace{-0.8em} has recently emerged as a simple and effective paradigm for transferring explicit reasoning procedures to \acp{llm} via supervised learning on long chain-of-thought trajectories. Early rationale-supervision methods such as STaR bootstrap reasoning by generating rationales, filtering correct ones, and fine-tuning on the retained traces~\citep{zelikman2022star}. Instruction-tuning and explanation-distillation studies further show that chain-of-thought demonstrations or teacher-generated explanations improve reasoning, as in FLAN-CoT~\citep{chung2024scaling}, Orca~\citep{mukherjee2023orca}, and Orca~2~\citep{mitra2023orca}. In mathematical reasoning, specialized SFT datasets with intermediate derivations or program-aided traces have been developed by MAmmoTH~\citep{yue2023mammoth}, ToRA~\citep{gou2023tora}, and NuminaMath~\citep{li2024numinamath}. More recent o1-style studies show that long reasoning can be elicited from small but carefully curated SFT datasets: s1 uses 1,000 examples~\citep{muennighoff2025s1}, LIMO reports strong reasoning from 817 examples~\citep{ye2025limo}, and \citet{li2025llms} transfer long-chain reasoning through data-efficient \ac{sft} or LoRA on long-CoT demonstrations. Recent work also studies efficiency and length control, including O1-Pruner for reducing redundant reasoning~\citep{luo2025o1} and long-short mixture SFT for mitigating overthinking~\citep{yu2025long}. Together, these works show that long-reasoning \ac{sft} effectively transfers reasoning behaviors from demonstrations, while its success depends on trace quality, reasoning structure, data curation, and inference-length control. Unlike these works, we study the \ac{ood} generalization ability of \ac{sft} on reasoning tasks that differ from the training dataset. We investigate how properties of the training dataset influence this generalization ability.

\paragraph{Understanding \ac{sft} and \ac{rl}} \hspace{-0.8em}  has been an active topic, as both \ac{sft} and \ac{rl} can improve model reasoning but appear to do so through different mechanisms. A central finding is that \ac{sft} often imitates or memorizes supervised demonstrations, whereas \ac{rl} can better elicit generalizable behavior~\citep{chu2025sft,chen2025sft,shenfeld2025rls,huan2026doesmath}. In particular, \citet{huan2026doesmath} compare models trained on math tasks and evaluated on other reasoning tasks, showing that \ac{sft} is less effective than \ac{rl} in improving general reasoning ability. Revisiting this setting, \citet{ren2026rethinking} find that \ac{sft} with long chain-of-thought supervision can improve cross-domain reasoning, with gains depending on training duration, the quality and structure of the training data, and base model capability. Mechanistically, \ac{grpo} appears to amplify existing capabilities with milder parameter changes, while \ac{sft} can overwrite prior skills and degrade out-of-domain performance~\citep{rajani2025scalpel}. This distinction is refined by works showing that \ac{rl} can recover out-of-distribution capability forgotten during \ac{sft}~\citep{jin2025rl}, mitigate forgetting through on-policy, mode-seeking data~\citep{chen2025retaining}, and reshape reasoning trajectories by compressing incorrect paths rather than homogenizing correct ones~\citep{matsutani2025rl}. Other studies qualify the advantage of \ac{rl}, attributing it partly to implicit medium-difficulty data filtering~\citep{lu2026does}, showing its dependence on model scale and task difficulty~\citep{lu2025generalization}, and questioning whether \ac{rlvr} creates new reasoning patterns beyond those already present in the base model~\citep{yue2026does}. High \ac{sft} scores may also poorly predict post-\ac{rl} gains~\citep{kang2025quagmires}, while broader surveys organize \ac{sft}, \ac{rl}, and hybrid post-training methods under a unified comparison framework~\citep{jiang2026supervised}. More recently, on-policy distillation has emerged as an \ac{sft}-style alternative that uses on-policy data to mitigate the forgetting issues of conventional \ac{sft} while improving sample and compute efficiency relative to \ac{rl}~\citep{lu2025onpolicydistillation,chen2025retainingdoingroleonpolicy}. Subsequent self-distillation methods further allow the same model to act as both teacher and student~\citep{zhao2026selfdistilledreasoneronpolicyselfdistillation,hubotter2026reinforcement}. Our work also studies the relationship between \ac{sft} and \ac{rl}. Unlike prior work, however, we show that \ac{sft} can generalize as well as \ac{rl} when trained with an appropriate data source and sufficient data scale.

\section{More Experimental Results and Details}\label{app:exp}

\subsection{Experimental Results on Qwen3-8B-Base}\label{app:qwen3_8b_results}
To better support our claims, we further conduct the experiments on Qwen3-8B-Base with the same setup to reproduce Figure~\ref{fig:plot_3_mix2}: SFT with a matched number of training rollouts as RL. The result is presented in Table~\ref{tab:qwen3_8b}.

\begin{table}[ht]
\centering
\begin{tabular}{lc}
\toprule
\textbf{Method} & \textbf{General Reasoning Average} \\
\midrule
RL-tuned & 42.5 \\
SFT w/RL-tuned & 42.1 \\
Mix-1 & 41.3 \\
Mix-2 & 42.0 \\
Mix-3 & \textbf{42.6} \\
SFT w/Qwen3-14B & 39.7 \\
\bottomrule
\end{tabular}
\caption{General reasoning performance comparison with different mixing ratios on Qwen3-8B-Base.}
\label{tab:qwen3_8b}
\end{table}

We observe a similar trend as that in Figure~\ref{fig:plot_3_mix2}: on general reasoning tasks, with a mixed dataset and a rollout budget matched to RL training, SFT can achieve comparable general reasoning ability as RL. We also observe that the best mixture rate changes to 50\%.

\subsection{\ac{rl} Training Details}\label{app:rl_train_config}
As the original DAPO dataset duplicates each prompt 100 times due to reproduction concerns, we pre-process the dataset and obtain $17391$ unique prompts. We add the following prompt suffix to the input question, as per the standard procedure in the literature~\citep{liu2025understanding,chen2026acereasonnemotron} to encourage reasoning:
\begin{center}
    ``Please reason step by step, and put your final answer within \textbackslash\textbackslash boxed\{\}.''
\end{center}

The parameter used in \ac{rl} training is listed in Table~\ref{tab:rl_hparams}.

\begin{table}[t]
\centering
\caption{Hyperparameters for RL training with \texttt{verl.trainer.main\_ppo}.}
\label{tab:rl_hparams}
\begin{tabular}{lc}
\toprule
\textbf{Hyperparameter} & \textbf{Value} \\
\midrule
Training batch size & 128 \\
PPO mini-batch size & 64 \\
PPO micro-batch size per GPU & 1 \\
Prompt length & 1024 \\
Response length & 8192 \\
Rollout sample number ($n$) & 8 \\
Learning rate & $ 10^{-6}$ \\
Clip ratio high & 0.28 \\
Use KL loss & False \\
Use KL in reward & False \\
Entropy coefficient & 0 \\
Temperature & 1.0 \\
Top-$p$ & 1.0 \\
Enable chunked prefill & False \\
Enforce eager execution & False \\
Free cache engine & False \\
Use remove padding & True \\
Enable gradient checkpointing & True \\
Ulysses sequence parallel size & 1 \\
Actor parameter offload & True \\
Actor optimizer offload & True \\
Reference parameter offload & True \\
Critic warmup & 0 \\
\bottomrule
\end{tabular}
\end{table}

\subsection{\ac{sft} Training Details}\label{app:sft_train_config}
\subsubsection{General \ac{sft} Training Configurations}
The parameter used in \ac{sft} training is listed in Table~\ref{tab:sft_hparams}.

\begin{table}[t]
\centering
\caption{Hyperparameters for supervised fine-tuning (SFT).}
\label{tab:sft_hparams}
\begin{tabular}{lc}
\toprule
\textbf{Hyperparameter} & \textbf{Value} \\
\midrule
DeepSpeed config & \texttt{deepspeed\_z3} \\
Template & \texttt{qwen} \\
Cutoff length & 16000 \\
Per-device train batch size & 1 \\
Gradient accumulation steps & 8 \\
Learning rate & $1.0 \times 10^{-5}$ \\
Number of training epochs & 1.0 \\
Learning rate scheduler & \texttt{cosine} \\
Warmup ratio & 0.1 \\
BF16 & True \\
\bottomrule
\end{tabular}
\end{table}

\subsubsection{Details of the \ac{sft} Training Dataset Size}\label{app:exp_data_scale_detail}
In Section~\ref{sec:scale}, we match the \ac{sft} training dataset size with those during \ac{rl} training. 
Specifically,
\begin{itemize}
    \item When the number of solvable prompts is matched, the corresponding Qwen3-14B dataset contains $9963$ prompt–rollout pairs; and the \ac{rl}-tuned dataset contains $10020$ prompt–rollout pairs.
    \item When the number of correct rollouts per prompt in the SFT training dataset is matched to that observed during RL training, both Qwen3-14B dataset and the \ac{rl}-tuned dataset consist of $39623$ prompt-rollout pairs, respectively.
    \item When the total number of rollouts for each prompt is matched, each of the two datasets contains $39623$ correct training samples with $98641$ prompt-wrong rollout pairs.
\end{itemize}

\subsection{Evaluation Setup Details}\label{app:exp_eval_detail}
For the math ability evaluation, we adopt \textrm{Avg}@32 for AIME24 due to the small sample size, and \textrm{Pass}@1 for the rest of the tasks. To evaluate the generalization ability of the trained models, we evaluate on the following reasoning benchmarks, which are accessible via \citet{eval-harness}: 
\begin{itemize}
    \item GPQA-Diamond~\citep{rein2024gpqa}:  a challenging dataset of 448 high-quality and difficult multiple-choice questions written by domain experts in biology, physics, and chemistry.
    \item HEAD-QA~\citep{vilares-gomez-rodriguez-2019-head}: a healthcare dataset for complex reasoning, which contains questions covering medicine, nursing, psychology, chemistry, pharmacology and biology.
    \item ACPBench~\citep{kokel2025acp}: a benchmark for evaluating the reasoning tasks in the field of planning, consisting of 7 reasoning tasks over 13 planning domains.
    \item ARC~\citep{Clark2018ThinkYH}: it consists of 7787 science exam questions drawn from a variety of sources, including science questions provided under license by a research partner affiliated with AI2. The questions are sorted into a Challenge set of ``hard'' problems and a Easy set of ``easy'' problems. We adopt the challenge set.
    \item BIG-Bench Hard (BBH)~\citep{suzgun-etal-2023-challenging}: a suit of 23 challenging BIG-Bench tasks, where the language models usually perform worse than human-raters.
    \item C-Eval~\citep{huang2023ceval}: a comprehensive Chinese evaluation suite, targeting at the knowledge and reasoning abilities of LLMs within the context of Chinese language and culture.
    \item COPAL-ID~\citep{wibowo-etal-2024-copal}: an Indonesian causal commonsense reasoning dataset that captures local nuances.
    \item GroundCocoa~\citep{kohli-etal-2025-groundcocoa}: a lexically diverse benchmark connecting compositional and conditional reasoning skills to the real-world problem of flight booking.
    \item MMLU-Pro~\citep{wang2024mmlupro}: an enhanced dataset which incorporates more challenging, reasoning-focused questions than the well-known MMLU dataset, and expands the choice set from four to ten options.
    \item MMLU-Pro+~\citep{asgari2024mmlupro}: an enchanced benchmark building upon MMLU-Pro, which is used to assess shortcut learning and higher-order reasoning in LLMs.
    \item OpenBookQA~\citep{mihaylov-etal-2018-suit}: a question-answering dataset which models open book exams and requires external knowledge and reasoning.
    \item PIQA~\citep{Bisk2020}: a physical commonsense reasoning and a corresponding benchmark dataset.
\end{itemize}
As post-training usually is done with templates, we apply the default chat template in \citet{eval-harness} during evaluation.

\subsection{More Evaluation Results for the Data Source Experiment}\label{app:exp_results_data_source_exp}
We provide more detailed evaluation results on the general reasoning tasks of the models trained via mixed dataset in Section~\ref{sec:training_data_matters} in Table~\ref{tab:benchmark_results}.

\begin{table}[htbp]
    \centering
    \vspace{0.5em}
    \scriptsize
    \setlength{\tabcolsep}{4pt}
    \begin{tabular}{l cc | cc>{\columncolor{blue!8}}ccccc}
        \toprule
        \textbf{Benchmark} & Qwen2.5-7B & RL & \makecell{SFT w/\\RL-tuned} & \makecell{Mix\\model 1} & \makecell{Mix\\model 2} & \makecell{Mix\\model 3} & \makecell{Mix\\model 4} & \makecell{Mix\\model 5} & \makecell{SFT w/\\Qwen3-14B} \\
        \midrule
        GPQA-Diamond   & 22.7 & 30.3 & 21.7 & 23.7 & 25.3 & 21.7 & \textbf{28.8} & \underline{27.3} & \underline{27.3} \\
        HEAD-QA        & 33.6 & 33.6 & 32.9 & \textbf{33.3} & \underline{33.1} & 33.0 & 32.1 & 31.6 & 31.4 \\
        ARC            & 42.1 & 42.0 & 40.3 & 40.8 & \underline{41.6} & \textbf{42.9} & 40.3 & 40.0 & 37.8 \\
        ACPBench       & 23.3 & 50.1 & \textbf{58.0} & \underline{56.6} & 53.4 & 50.7 & 43.7 & 46.6 & 45.3 \\
        BIG-Bench Hard & 10.9 & 42.9 & 33.8 & 34.4 & 37.3 & 35.6 & 34.1 & \underline{37.7} & \textbf{42.1} \\
        C-Eval         & 40.1 & 43.6 & 42.4 & 36.4 & 47.8 & 28.2 & \underline{52.9} & \textbf{53.8} & 36.5 \\
        COPAL-ID       & 61.7 & 60.8 & 60.1 & 59.7 & \underline{60.3} & \textbf{61.4} & 59.0 & 59.4 & 59.0 \\
        GroundCocoa    & 34.7 & 34.8 & 35.6 & \underline{36.8} & \textbf{37.0} & 36.7 & 34.6 & 34.2 & 33.2 \\
        MMLU-Pro       & 9.2  & 45.7 & \underline{19.4} & 17.9 & \textbf{21.3} & 3.8  & 0.7  & 1.2  & 1.1 \\
        MMLU-Pro+      & 21.0 & 31.8 & \underline{13.8} & \textbf{15.1} & \textbf{15.1} & 2.5  & 1.6  & 1.1  & 0.8 \\
        OpenBookQA     & 28.6 & 33.2 & 30.6 & \textbf{32.8} & \underline{32.4} & 32.2 & 30.2 & 30.0 & 27.2 \\
        PIQA           & 77.1 & 76.1 & 76.2 & 76.3 & 76.7 & 76.5 & \underline{77.3} & 77.1 & \textbf{77.5} \\
        \midrule
        Average        & 33.8 & 43.8 & \underline{38.7} & \underline{38.7} & \textbf{40.1} & 35.4 & 36.3 & 36.7 & 34.9 \\
        \bottomrule
    \end{tabular}
    \vspace{1em}
    \caption{Performance Comparison Across General Reasoning Benchmarks}
    \label{tab:benchmark_results}
\end{table}

\subsection{Evaluation Results with Standard Error}\label{app:with_std}
Our evaluation pipeline contains \textbf{5 benchmarks} from math domain and \textbf{12 benchmarks} of general reasoning domain, which contain \textbf{enough} prompts to yield accurate evaluation results to support our claims. To better support our results, we aggregate the standard error (stderr) metrics given in the lm-eval-harness results, and report the general reasoning performance when the total rollouts are matched (Figure~\ref{fig:plot_3_mix2}) in Table~\ref{tab:eval_std}.

\begin{table}[ht]
\centering
\begin{tabular}{lc}
\toprule
\textbf{Method} & \textbf{Average score $\pm$ stderr} \\
\midrule
Base & $33.8 \pm 0.41$ \\
RL & $43.8 \pm 0.43$ \\
SFT w/ RL-tuned & $43.7 \pm 0.41$ \\
Mix 1 & $\mathbf{44.4 \pm 0.42}$ \\
Mix 3 & $42.9 \pm 0.41$ \\
Mix 5 & $41.3 \pm 0.42$ \\
SFT w/ Qwen3-14B & $36.7 \pm 0.43$ \\
\bottomrule
\end{tabular}
\caption{Evaluation results with standard error.}
\label{tab:eval_std}
\end{table}

These standard errors quantify evaluation-set uncertainty, but do not capture variation across training seeds due to practical computation constraints. We acknowledge this limitation of our experiments.

\subsection{Ablation Study on the Dataset Scale with Fixed Compute}\label{app:exp_ablation_compute}
In Section~\ref{sec:scale}, we have studied the effect of the \ac{sft} training data scale on the generalization ability of the \ac{sft}-tuned model.
The training compute also scales as more training data is introduced. To dig out the true workhorse behind data scale and compute, we carry out an ablation study.

Specifically, we fix the total compute used for \ac{sft} and vary the composition of the training data.
In the first ablation study, we train the model with \ac{sft} using all $39623$ prompt-correct rollout pairs for $3$ epochs. In this setting, the total compute is approximately aligned ($118869$ vs. $138264$), while the number of unique training samples differs.
In the second set of experiments, we randomly sample $1/k$ fraction of the $8$ rollouts for each prompt, and train the base models via \ac{sft} for $k$ epochs for $k=2,4,8$. In this case, the compute is aligned, but less training data is used.

\begin{figure}[t]
    \centering

    \begin{minipage}[t]{0.49\textwidth}
        \centering
        \includegraphics[width=\linewidth]{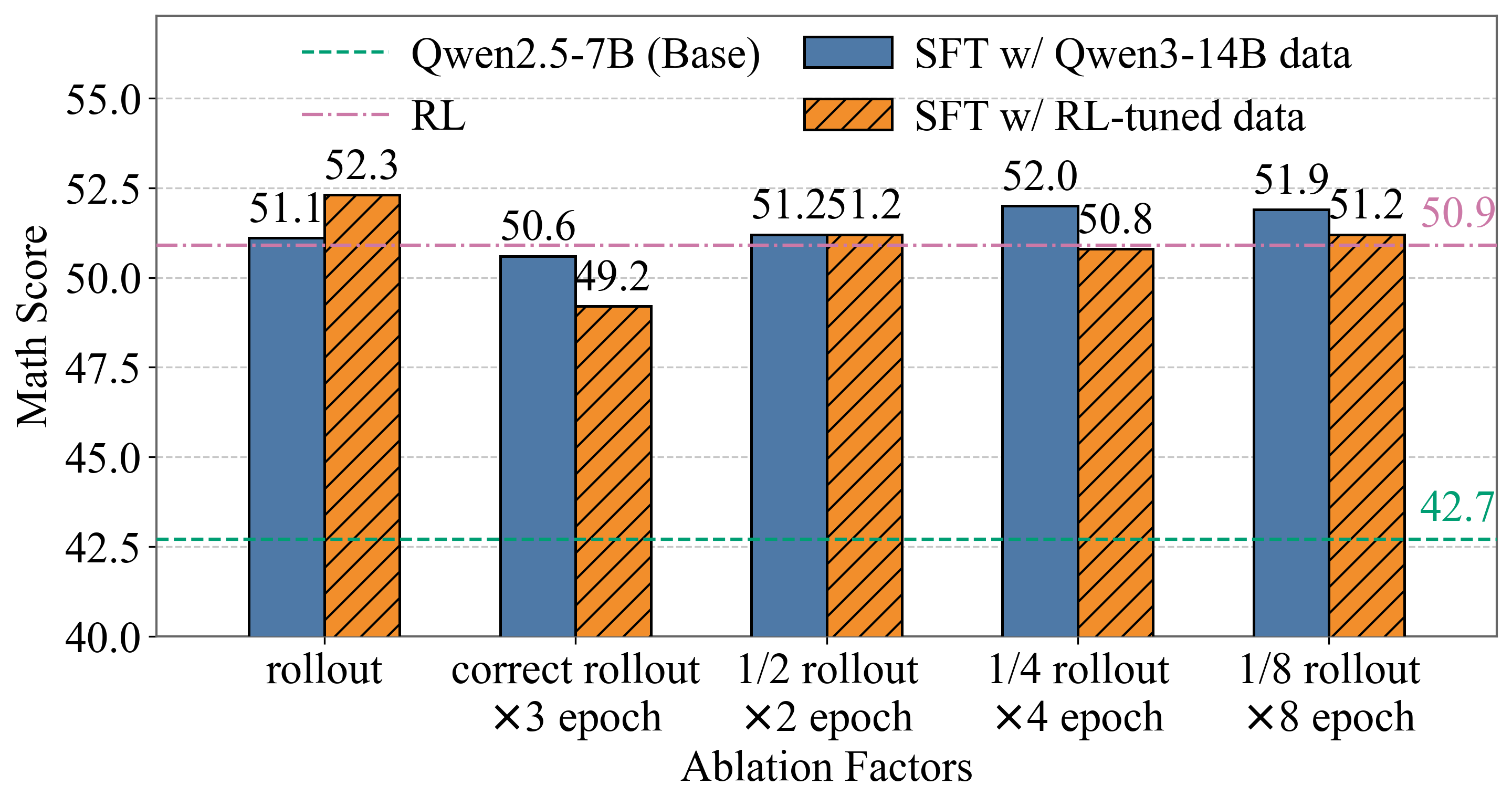}
        \caption*{(a) Math benchmarks}
    \end{minipage}
    \hfill
    \begin{minipage}[t]{0.49\textwidth}
        \centering
        \includegraphics[width=\linewidth]{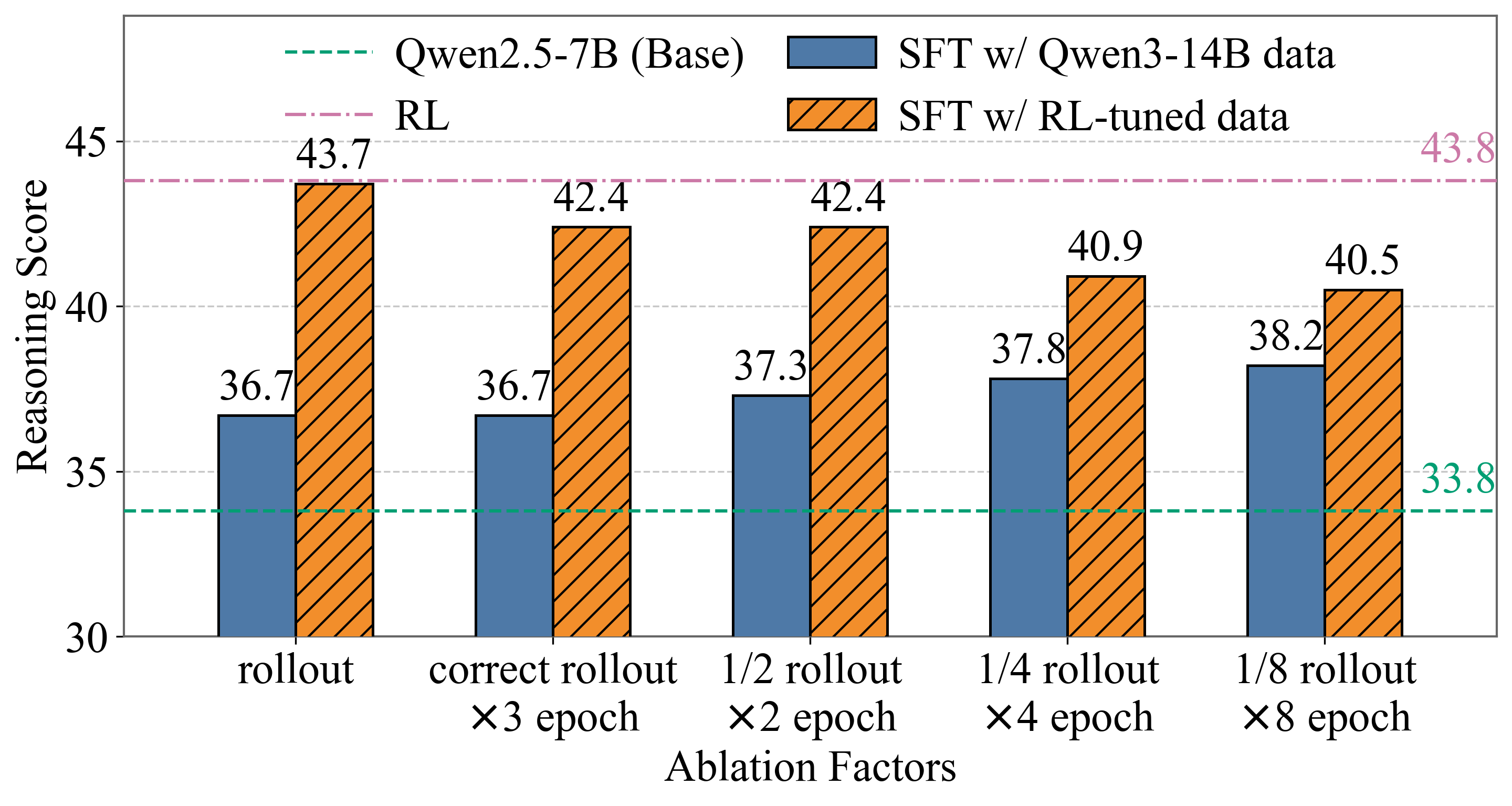}
        \caption*{(b) General reasoning benchmarks}
    \end{minipage}

    \caption{Ablation study on the components of the training dataset with the training compute budget fixed. When the training dataset shrinks, the performance also downgrades.}
    \label{fig:plot_4_ablation}
\end{figure}

The experimental results are presented  in Figure~\ref{fig:plot_4_ablation}.
It indicates using fewer training samples but with the same compute as in \ac{rl} training, the final performance on general reasoning tasks decreases compared to that of using the whole dataset.

We conclude that the training data scale, i.e., the number of unique rollouts, matters more than compute for \ac{sft} training given the same gradient compute budget as \ac{rl} training.

\subsection{Rollout Analysis}\label{app:rollout_analysis}

We list the classified $138$ consolidated methods in Table~\ref{tab:math_methods}.
\begin{table}[htbp]
    \centering
    \caption{Alphabetical Classification of Mathematical Methods} 
    \label{tab:math_methods}
    \vspace{1em}
    \renewcommand{\arraystretch}{1.2} 
    \small 
    \resizebox{\textwidth}{!}{%
        \begin{tabular}{>{\raggedright\arraybackslash}p{0.32\textwidth} >{\raggedright\arraybackslash}p{0.32\textwidth} >{\raggedright\arraybackslash}p{0.32\textwidth}}
            \toprule
            \textbf{Method} & \textbf{Method} & \textbf{Method} \\
            \midrule
            3D geometry \& volumes & equation solving & partial fractions \\
            absolute value methods & estimation \& bounding & partitions \\
            algebraic manipulation & Euler's formula & pattern recognition \\
            angle properties & expected value & percentages \\
            area \& perimeter & exponents \& powers & perfect powers \\
            arithmetic operations & extremal principle & periodicity \\
            asymptotic analysis & factoring & permutations \\
            averages & Fermat's \& Euler's theorems & pigeonhole principle \\
            bijection \& double counting & Fibonacci \& special sequences & polygons \\
            binomial expansion & floor \& ceiling functions & polynomial methods \\
            calculation \& computation & fractions \& ratios & prime numbers \\
            calculus (general) & function analysis & probability \\
            case analysis & functional equations & problem reformulation \\
            centroid \& triangle centers & game theory \& strategy & proof by contradiction \\
            characteristic equation & GCD \& LCM & proof by induction \\
            Chinese Remainder Theorem & generating functions & Pythagorean theorem \\
            circles \& circle theorems & geometry (general) & quadratic methods \\
            classical geometry theorems & graph theory & random walks \& Markov chains \\
            coefficient comparison & greedy algorithms & range \& domain analysis \\
            coloring \& tiling arguments & grid \& lattice methods & rational root theorem \\
            combinations \& binomial coefficients & group theory & rationalization \\
            combinatorial identities & inclusion-exclusion & recurrence relations \\
            combinatorics (general) & inequalities & rings \& fields \\
            completing the square & integer methods & root analysis \\
            complex numbers & integration & search \& traversal \\
            conic sections & interval analysis & sequence analysis \\
            constraint analysis & invariants & series \& convergence \\
            construction & iterative methods & set theory \\
            coordinate geometry & known results \& formulas & similarity \& congruence \\
            counting (general) & limits & simplification \& expansion \\
            counting principles & linear algebra & simulation \& enumeration \\
            critical points \& optimization (calculus) & lines \& slopes & sorting \\
            degree analysis & logarithms & squaring both sides \\
            derivatives & logical deduction & statistics \\
            determinants & magnitude \& norm & substitution \\
            difference of squares & matrices & summation techniques \\
            differential equations & midpoint methods & symmetry \\
            digits \& number representation & modular arithmetic & systems of equations \\
            Diophantine equations & multivariable calculus & transformations \\
            discriminant analysis & number theory (general) & triangle geometry \\
            distance formulas & number-theoretic functions & trigonometric identities \\
            distributions & optimization & trigonometry \\
            divisibility \& divisors & other & unit conversion \\
            dot product \& cross product & p-adic valuations & vectors \\
            dynamic programming & parametric methods & verification \& checking \\
            eigenvalues \& eigenvectors & parity arguments & Vieta's formulas \\
            \bottomrule
        \end{tabular}%
    }
\end{table}

Furthermore, we conducted a controlled experiment for the tail analysis which supports that the tail-coverage does influence the output in a causal way as implied by Figure~\ref{fig:tail_helps}. Specifically, we generate more rollouts with the RL-tuned model, and we use those rollouts which only adopt the methods accounting for the top 80\% as the SFT training dataset. After the same SFT training and processing of the evaluation rollouts, we have the tail method used count for all rollouts and the correct rollouts in Table~\ref{tab:tail_usage}.

\begin{table}[ht]
\centering
\begin{tabular}{lcc}
\toprule
\textbf{Benchmark} & \textbf{SFT w/ less tail} & \textbf{SFT w/ RL-tuned data} \\
\midrule
\multicolumn{3}{l}{\textit{All rollouts}} \\
\hspace{1em} MMLU-Pro  & 455 & 550 \\
\hspace{1em} MMLU-Pro+ & 431 & 589 \\
\midrule
\multicolumn{3}{l}{\textit{Correct rollouts}} \\
\hspace{1em} MMLU-Pro  & 40 & 108 \\
\hspace{1em} MMLU-Pro+ & 21 & 84 \\
\bottomrule
\end{tabular}
\caption{Tail method usage count. With more tail methods in the training dataset, more tail methods are used.}
\label{tab:tail_usage}
\end{table}

With more tail data in the SFT training, the SFT-tuned model tries to use more tail methods during the evaluation. This indicates that the tail-coverage influences the generalization causally.

\section{Notations and Outline of Proofs}
\begin{table}[H]
    \centering
    \begin{tabularx}{\textwidth}{ll}
        \toprule
        \rowcolor{gray}
        \textbf{Notations} & \textbf{Descriptions} \\
        \midrule
        $\pit=(X_1^\top, X_2^\top, \ldots, X_N^\top)^\top$ & The input informative and context tokens.\\
        \midrule
        $S_n=\Soft(X_q\W \pit^\top)_n$ & The attention score from query $X_q$ to token
        $X_n$ at position $n$.\\
        \midrule
        $S_{l,k}=\sum_{n=1}^N\mathbbm{1}{\{X_n=i_{l,k}\}}S_n$ & The attention score from query $u_l$ to informative token
        $i_{l,k}$.\\
        \midrule
        $\widetilde S_m=\sum_{n=1}^N\mathbbm{1}{\{X_n = c_j\}}S_n$ & The attention score from query $X_q$ to context token $c_j$.\\
        \midrule
        $\widetilde S=\sum_{j=1}^J\widetilde S_m$ & The attention score from query $X_q$ to all context token.\\
        \midrule
        $B_{l,k} = u_l \W  i_{l,k}^\top$ & The attention logit from query $u_l$ to informative token
        $i_{l,k}$.\\
        \midrule
        $\widetilde B_{l,j} = u_l \W  c_j^\top$ & The attention logit from query $u_l$ to context token
        $c_j$.\\
        \midrule
        $\beta_{l,k} = -u_l \nabla_\W L_{\mathrm{SFT}}\, i_{l,k}^\top$ & Update of logit from query $u_l$ to informative token
        $i_{l,k}$.\\
        \midrule
        $\widetilde \beta_{l,j} = -u_l \nabla_\W L_{\mathrm{SFT}}\, c_j^\top$ & Update of attention logit from query $u_l$ to context token
        $c_j$.\\
        \midrule
        $ \esb$ (\Cref{lem: concen-P}) & High probability event for token concentration.\\
        \midrule
        $A_l
    =
    \sum_{k=1}^K a_{(l-1)K+k}$ & Class-specific correct-answer indicator vector.\\
    \midrule
    $A_{-l}
    =
    \sum_{l'\neq l}\sum_{k'=1}^Ka_{l',k'}$ & Class-specific incorrect-answer
indicator vector.\\
\midrule
       $\Phi_l
    =
    FA_l^\top$ & Probability assigned to correct answers of class $l$\\
    \midrule
       $\Phi_{-l}
    =
    FA_{-l}^\top $ & Probability assigned to incorrect answers of class $l$\\
        \bottomrule
    \end{tabularx}
    \vspace{3pt}
    \caption{Summary of frequently used notations in proof. We omit $(t)$ here for simplicity.}
    \label{Table: Summary of Notations}
\end{table}
\paragraph{Notations.} We first summarize the frequently used notations in \Cref{Table: Summary of Notations}. In addition, the pre-trained knowledge is stored in the language model head:
\begin{align*}
    W_{\lm}=c_{\lm}\log(K)\big(\sum_{l=1}^L \sum_{k=1}^K i_{l,k}^\top (a_{l,k}-A_{-l}-\widetilde a)+\sum_{j=1}^J c_j^\top\widetilde a \big),
\end{align*}
where $A_{-l}=\sum_{l'\neq l}\sum_{k'=1}^Ka_{l',k'}\in\mathbb{R}^{LK+1}$. In addition, we have for the output as,
\begin{align*}
    F = \Soft\bigg(c_{\lm}\log(K)\bigg(\sum_{k=1}^K S_{l,k}a_{l,k}^\top +(2\widetilde S-1)\widetilde a^\top-(1-\widetilde S) A_{-l}\bigg)\bigg).
\end{align*}

\paragraph{Roadmap of Proof Part.} We prove the SFT part and RL part of \Cref{thm: indist} in Appendix \ref{app:proof_sft} and Appendix \ref{app:proof_rl}, respectively, and prove \Cref{Thm: ood generalization} in Appendix \ref{app: ood}. We next summarize the outline of the proof of the SFT part of \Cref{thm: indist}, the proof of the RL part is similar.

\section{Proofs of SFT}\label{app:proof_sft}
\paragraph{Proof Outline of SFT.} The output $F$ is directly determined by attention scores $S$, while $S$ itself depends on the attention logits $B$. Accordingly, we first formalize the key quantities, namely the attention logits and their update rules, in \Cref{def:sft-logits}, and then analyze their evolution throughout optimization. We compute the gradient in \Cref{lem:sft-gradient} and characterize the logit updates in \Cref{lem:logit-up}. Building on these foundations, we analyze one step of gradient descent in Appendix \ref{app: sft dynamics}, which proves the SFT part of \Cref{thm: indist}.

\subsection{Gradient Computations}\label{app:gradient}
In this section, we first calculate the gradient with respect to $W$ in \Cref{lem:sft-gradient} and characterize the key variables: logits and their updates in \Cref{def:sft-logits} and \Cref{lem:logit-up}. We omit $(t)$ and the subscript $\sft$ in this section when there is no ambiguity and write $L_\sft(W_\sft)$ as $L_\sft$ here for simplicity.
\begin{lemma}[SFT Gradient]\label{lem:sft-gradient}
Recall that $\pit=(X_1^\top, X_2^\top, \ldots, X_N^\top)^\top$. The gradient of the SFT loss function with respect to $W$ is given by 
    \begin{align*}
    \nabla_\W L_{\mathrm{SFT}} = \mathbb{E}_{\pit,X_q,A\sim\mathcal D_{\mathrm{SFT}}}
\!\left[\,
X_q
\Big(
(F-A)W_{\lm}^\top \pit^\top
\big(\diag(S)-S^\top S\big)
\Big)
\pit
\right],
\end{align*}
where \(S=\Soft(X_q^\top \W \pit^\top)\in \mathbb{R}^{N \times 1}\) is the
attention score. The output is
\(F=\Soft(S \pit W_{\lm})\in \mathbb{R}^{1 \times (LK+1)}\).
\end{lemma}
\begin{proof}
For a single sample \((\pit,X_q,A)\), we define
\(s=X_q^\top \W \pit^\top\in \mathbb{R}^{1\times N}\),
\(S=\Soft(s)\in \mathbb{R}^{1\times N}\),
\(z=S P W_{\lm}\in \mathbb{R}^{1\times (LK+1)}\), and
\(F=\Soft(z)\in \mathbb{R}^{1\times (LK+1)}\).
Here \(\pit \in\mathbb{R}^{N\times d}\), \(X_q\in\mathbb{R}^{d\times 1}\), \(\W\in\mathbb{R}^{d\times d}\), and \(W_{\lm}\in\mathbb{R}^{d\times (LK+1)}\).

The single-sample loss is
\[
\ell_\sft(\W;\pit,X_q,A)
=
-\sum_{l=1}^L\sum_{k=1}^K\mathbbm{1}\{A=a_{l,k}\}\log F_{(l-1)K+k}.
\]
The corresponding population loss is
\[
L_{\mathrm{SFT}}(\W)
=
\mathbb{E}_{\pit,X_q,A\sim\mathcal D_{\mathrm{SFT}}}
\big[\ell_\sft(\W;\pit,X_q,A)\big].
\]
We first compute the gradient of single-sample loss, then take the expectation to get the gradient of population loss. We compute the gradient of single-sample loss by applying the chain rule. First, differentiating with respect to the output logits \(z\) and using the standard cross-entropy gradient for softmax gives
\[
g_z = \nabla_z \ell_\sft
=
\sum_{l=1}^L\sum_{k=1}^K\mathbbm{1}\{A=a_{l,k}\}(F-e_{(l-1)K+k})
\in \mathbb{R}^{1\times (LK+1)}.
\]

Next, differentiating with respect to the inner softmax output \(S\), we use
\(z=S P W_{\lm}\) to obtain
\(g_S=\nabla_S \ell_\sft
=g_z W_{\lm}^\top \pit^\top\in \mathbb{R}^{1\times N}\).

We then differentiate through the inner softmax. Since \(S=\Soft(s)\), the
Jacobian of the row-softmax is
\(J_{\Soft}(s)=\diag(S)-S^\top S\in \mathbb{R}^{N\times N}\).
Therefore, the gradient with respect to the attention logits is
\begin{align*}
    g_s
=
g_S\big(\diag(S)-S^\top S\big)
=
\sum_{l=1}^L\sum_{k=1}^K \mathbbm{1}\{A=a_{l,k}\}(F-e_{(l-1)K+k})W_{\lm}^\top \pit^\top
\big(\diag(S)-S^\top S\big)\in \mathbb{R}^{1\times N}.
\end{align*}

Finally, we differentiate with respect to \(\W\). Recall that $s = X_q^\top \W \pit^\top$, whose differential is $\mathrm{d}s = X_q^\top\, \mathrm{d}\W\, \pit^\top$. Hence, we have that 
\begin{align*}
\mathrm{d}\ell_\sft
&\quad=
g_s\, \mathrm{d}s \\
&\quad=
g_s X_q^\top\, \mathrm{d}\W\, \pit^\top \\
&\quad=
\tr\!\big(\pit^\top g_s X_q^\top\, \mathrm{d}\W\big)
=
\tr\!\big((X_q g_s P)^\top \mathrm{d}\W\big).
\end{align*}
Therefore, the single-sample gradient with respect to \(\W\) is
\(\nabla_\W \ell_\sft=X_q\, g_s\, P\in \mathbb{R}^{d\times d}\).
Substituting the expression for \(g_s\) into the single-sample gradient gives
\begin{align*}
  &\nabla_\W \ell_\sft =
\!\sum_{l=1}^L\sum_{k=1}^K\!\mathbbm{1}\{A=a_{l,k}\}\,
X_q
\Big[
(F-e_{(l-1)K+k})W_{\lm}^\top \pit^\top
\big(\diag(S)\!-\!S^\top S\big)
\Big]
\pit.  
\end{align*}

Taking expectation over the SFT data distribution gives the population gradient
\begin{align*}
&\nabla_\W L_{\mathrm{SFT}}(\W)\\
&\quad
=
\mathbb{E}_{\pit,X_q,A\sim\mathcal D_{\mathrm{SFT}}}
\!\left[
\sum_{l=1}^L\sum_{k=1}^K \mathbbm{1}\{A=a_{l,k}\}\,
X_q
\Big(
(F-e_{(l-1)K+k})W_{\lm}^\top \pit^\top
\big(\diag(S)-S^\top S\big)
\Big)
\pit \right]\\
& \quad
=
\mathbb{E}_{\pit,X_q,A\sim\mathcal D_{\mathrm{SFT}}}
\!\left[\,
X_q
\Big(
(F-A)W_{\lm}^\top \pit^\top
\big(\diag(S)-S^\top S\big)
\Big)
\pit
\right],
\end{align*}
where the attention score is \(S=\Soft(X_q^\top \W \pit^\top)\) and the output
is \(F=\Soft(S \pit W_{\lm})\).
This completes the proof of the SFT gradient formula.
\end{proof}

For each question $l$, we define the important logits during training. Since only informative tokens from $\mathcal{I}_l$ and context tokens from $\mathcal{C}$ can occur in the prompt of question $l$, it is enough to track the following variables.
\begin{definition}[Important Logits and Their Update Rate in SFT]\label{def:sft-logits}
For any $l \in [L]$, $k \in [K]$ and $j \in [J]$, for iteration $t\geq 0$, we denote the weight as $\W^{(t)}$ and the loss as $L_{\mathrm{SFT}}^{(t)}=L_{\mathrm{SFT}}(\W^{(t)})$. We define the following quantities:
\begin{align*}
\mathrm{Attention\;Logits:}\;B_{l,k}^{(t)} &= u_l \W^{(t)}  i_{l,k}^\top, \qquad \widetilde B_{l,j}^{(t)} = u_l \W^{(t)}  c_j^\top; \\    
\mathrm{Logits\; Update\;Rate:}\;\beta_{l,k}^{(t)} & = -u_l \nabla_\W L_{\mathrm{SFT}}^{(t)}\, i_{l,k}^\top, \qquad \widetilde\beta_{l,j}^{(t)} = -u_l \nabla_\W L_{\mathrm{SFT}}^{(t)}\, c_j^\top.
\end{align*}
The GD update gives
\begin{align*}
   B_{l,k}^{(t+1)}&= B_{l,k}^{(t)}+\eta \beta_{l,k}^{(t)},\;\widetilde B_{l,j}^{(t+1)}= \widetilde B_{l,j}^{(t)}+\eta \widetilde\beta_{l,j}^{(t)}.
\end{align*}
The initialization $\W^{(0)}=\mathbf{0}_{d\times d}$ also gives $B_{l,k}^{(0)}=\widetilde B_{l,j}^{(0)}=0$.    
\end{definition}
In the remainder of the proof, for simplicity, we omit the dependence on the iteration $t$ when it is clear from the context.
\begin{lemma}[Logits Update Rate]\label{lem:logit-up}
For any $l \in [L]$, $k \in [K]$ and $j \in [J]$, the logit update rates are
\begin{align*}
\beta_{l,k} 
&  = - \frac{c_{\lm}\log(K)}{L}
\mathbb{E}_{\pit}\Big[S_{l,k} \Big((F_{(l-1)K+k}-p_{l,k})-\sum_{k'=1}^K S_{l,k'}(F_{(l-1)K+k'}-p_{l,k'})\\
& 
\qquad \qquad \qquad \qquad \qquad \qquad \qquad \qquad \qquad \qquad \qquad \qquad \quad - (2F_{LK+1}+\Phi_{-l}) \widetilde S
\Big)\Big| X_q = u_l\Big],\\
\widetilde \beta_{l,j} 
& 
=
-\frac{c_{\lm}\log(K)}{L}
\mathbb{E}_{\pit}\!\Big[ \widetilde S_j
\Big((2F_{LK+1}+\Phi_{-l})(1-\widetilde S)-\sum_{k'=1}^K S_{l,k'}(F_{(l-1)K+k'}-p_{l,k'})\Big)\Big| X_q = u_l
\Big].
\end{align*}
\end{lemma}
\begin{proof}
We first simplify the matrix factor appearing in the gradient. The diagonal
softmax term is
\(\pit^\top\diag(S)\pit=\sum_{n=1}^N S_n X_n X_n^\top\).
The rank-one softmax term satisfies
\[
\pit^\top S^\top S P
=
\sum_{n=1}^N\sum_{m=1}^N S_nS_m\,X_n X_m^\top
=
\left(\sum_{n=1}^N S_n X_n\right)\left(\sum_{m=1}^N S_m X_m\right)^\top.
\]
Combining these identities, this factor can be written as
\begin{align*}
\pit^\top(\diag(S)-S^\top S)\pit
&=
\sum_{n=1}^N S_n X_n X_n^\top
-
\sum_{n=1}^N\sum_{m=1}^N S_nS_m\,X_n X_m^\top \\
&=
\sum_{n=1}^N S_n
\left(X_n-\sum_{m=1}^N S_m X_m\right)X_n^\top.
\end{align*}
Substituting $W_{\lm}=c_{\lm}\log(K)\big(\sum_{l=1}^L\sum_{k=1}^K i_{l,k}^\top (a_{l,k}-A_{-l}-\widetilde a)+\sum_{j=1}^J c_j^\top\widetilde a \big)\in \mathbb{R}^{d \times (LK+1)}$ into this factor gives
\begin{align}
    &W_{\lm}^\top \pit^\top(\diag(S)-S^\top S)\pit \nonumber\\
    &\quad = 
    c_{\lm}\log(K)\big(\sum_{l=1}^L\sum_{k=1}^K S_{l,k} (a_{l,k}-A_{-l}-\widetilde a)^\top i_{l,k}+ \tilde a^\top \sum_{n=1}^N\mathbbm{1}{\{X_n \in \mathcal{C}\}}S_nX_n\big)\nonumber\\
    &\qquad \quad 
    - c_{\lm}\log(K)\big(\sum_{l=1}^L\sum_{k=1}^K S_{l,k} (a_{l,k}-A_{-l}-\widetilde a)^\top+\widetilde S \tilde a^\top \big)\big(\sum_{n=1}^N S_nX_n\big). \label{eq: track-1}
\end{align}
Note that the informative-token update rate is
    \begin{align}
\beta_{l,k} 
& = -u_l \nabla_\W L_{\mathrm{SFT}}\, i_{l,k}^\top\nonumber\\
& =
- \mathbb{E}_{\pit,X_q,A\sim\mathcal D_{\mathrm{SFT}}}
\!\left[\,
\mathbbm{1}{\{X_q=u_l\}}
\Big(
(F-A)W_{\lm}^\top \pit^\top
\big(\diag(S)-S^\top S\big)
\Big)
P i_{l,k}^\top
\right]\nonumber\\
&=
- \frac{1}{L} \mathbb{E}_{\pit,A\sim\mathcal D_{\mathrm{SFT}}}
\!\left[
\Big(
(F-A)W_{\lm}^\top \pit^\top
\big(\diag(S)-S^\top S\big)
\Big)
P i_{l,k}^\top \Big | X_q=u_l
\right], \label{eq: track-1-2}
\end{align}
where the last equation follows from $\mathbb{P}(X_q=u_l) =1/L$ for all $l \in [L]$.
Applying \Cref{eq: track-1} to an informative token gives
\begin{align}
W_{\lm}^\top &\pit^\top
\bigl(\diag(S)-S^\top S\bigr)
\pit\,i_{l,k}^\top\nonumber\\
&=
c_{\lm}\log(K)(S_{l,k} (a_{l,k}-A_{-l}-\widetilde a)^\top
-
S_{l,k} \sum_{l'=1}^L\sum_{k'=1}^K S_{l',k'} (a_{l',k'}-A_{-l'}-\widetilde a)^\top
-
S_{l,k} \tilde S\,\tilde a^\top)\nonumber \\
&=
c_{\lm}\log(K) S_{l,k}\left(
(a_{l,k}-A_{-l}-\widetilde a)^\top
-
\sum_{l'=1}^L\sum_{k'=1}^K S_{l',k'} (a_{l',k'}-A_{-l'}-\widetilde a)^\top
-
\tilde S\,\tilde a^\top
\right).\label{eq: track-1-3}
\end{align}
By definition, when \(X_q=u_l\), we have \(S_{l',k'}=0\) for all \(l'\neq l\), since question \(l\) contains informative tokens only from \(\mathcal{I}_l\). Therefore, by plugging \Cref{eq: track-1-3} into \Cref{eq: track-1-2}, the update rate for the informative logit can be written as
\begin{align*}
& \beta_{l,k}=
- \frac{c_{\lm}\log(K)}{L}\mathbb{E}_{\pit,A\sim\mathcal D_{\mathrm{SFT}}}\!\Big[
(F-A)\,
S_{l,k}\\
& \quad\quad\qquad\qquad\qquad
\left(
(a_{l,k}-A_{-l}-\widetilde a)^\top
-
\sum_{k'=1}^K S_{l,k'}(a_{l,k'}-A_{-l}-\widetilde a)^\top
-
\tilde S\,\tilde a^\top
\right)\Big | X_q=u_l
\Big]\\
& \overset{\RM{1}}{=}
- \frac{c_{\lm}\log(K)}{L}
\mathbb{E}_{\pit}\Big[S_{l,k} \\
& \quad\quad\quad\qquad
\mathbb{E}_{A\sim\mathcal D_{\mathrm{SFT}}}\!\left[
(F-A)\left(
a_{l,k}^\top
-
\sum_{k'=1}^K S_{l,k'} a_{l,k'}^\top
-
\widetilde S\,(2\tilde a+A_{-l})^\top
\right)\Big|\pit
\right]\Big|X_q=u_l\Big]\\
& \overset{\RM{2}}{=}
- \frac{c_{\lm}\log(K)}{L}
\mathbb{E}_{\pit}\Big[S_{l,k}\\
& \quad\quad\quad \Big((F_{(l-1)K+k}-p_{l,k})-\sum_{k'=1}^K S_{l,k'}(F_{(l-1)K+k'}-p_{l,k'})- (2F_{LK+1}+\Phi_{-l}) \widetilde S
\Big)\Big|X_q=u_l\Big],
\end{align*}
where $\RM{1}$ follows from applying the tower property, and $\RM{2}$ follows from taking expectation over $A$. For a context token, \Cref{eq: track-1} gives
\begin{align*}
&W_{\lm}^\top \pit^\top
 \bigl(\diag(S)-S^\top S\bigr)
\pit\,c_j^\top\\
&\quad=
c_{\lm}\log(K)(\widetilde S_j \tilde a^\top
-
\widetilde S_j \sum_{l'=1}^L\sum_{k'=1}^K S_{l',k'} (a_{l',k'}-A_{-l'}-\widetilde a)^\top
-
\widetilde S_j \tilde S\,\tilde a^\top) \\
&\quad=
c_{\lm}\log(K) \widetilde S_j\big(
\tilde a^\top
-
\sum_{l'=1}^L\sum_{k'=1}^K S_{l',k'} (a_{l',k'}-A_{-l'}-\widetilde a)^\top
-
\tilde S\,\tilde a^\top
\big).
\end{align*}
Applying the same calculation to the context logit yields
\begin{align*}
\widetilde \beta_{l,j}
&= -u_l \nabla_\W L_{\mathrm{SFT}}\, c_j^\top\\
&=
-\frac{c_{\lm}\log(K)}{L}
\mathbb{E}_{\pit}\!\Big[ \widetilde S_j
\Big((2F_{LK+1}+\Phi_{-l})(1\!-\!\widetilde S)\!-\!\sum_{k'=1}^K S_{l,k'}(F_{(l-1)K+k'}-p_{l,k'})\Big)\Big| X_q = u_l
\Big].
\end{align*}
This completes the proof of the logit update characterization.
\end{proof}

\subsection{Learning Dynamics of SFT}\label{app: sft dynamics}
With the gradient expressions derived in Appendix \ref{app:gradient}, we analyze the
training dynamics of SFT in this section. For each question \(l\in[L]\),
we fix \(l\) and condition on the event \(\{X_q=u_l\}\), and then study the
corresponding training dynamics. It is worth noting that, when \(X_q=u_l\), we
have \(S_{l',k'}=0\) for all \(l'\neq l\), since question \(l\) contains
informative tokens only from \(\mathcal I_l\).
For simplicity, we consider only one step.
We take a relatively large learning rate
\(\eta_{\sft}^{(0)}=KLc^{(\mathrm I)}/p_{l,1}\),
where \(c^{(\mathrm I)}>0\) is a sufficiently large constant.
Since \(p_{l,1}=\Omega(1)\gg L/K\), the head-answer token
receives the dominant update. Tail answers receive smaller positive updates.

After one gradient-descent update, the attention score concentrates on the informative
token corresponding to the correct answer, namely \(i_{l,1},\cdots,i_{l,K}\). 

\begin{lemma}[Initialization Variable]\label{lem:phase1-start-var}
At initialization \(t=0\), for all \(k\in[K]\) and \(j\in[J]\), the key
variables satisfy the following estimates.
\begin{itemize}[leftmargin=8pt]
    \item \textbf{Attention Logits.}
    The initialized logits are \(B_{l,k}^{(0)}=\widetilde B_{l,j}^{(0)}=0\).

    \item \textbf{Attention Scores.}
    Suppose \(\pit\in\esb\) satisfies the high-probability event in
    \Cref{lem: concen-P}. Under this event, the initial attention scores satisfy
    \begin{align}
       S_{l,k}^{(0)}
       &=
       \frac{1}{2K}
       \left(
       \!\!1\!+\!O\left(\frac{(K+J)^3}{N}\right)
       \right),
       \widetilde S_j^{(0)}
       =
       \frac{1}{2J}
       \left(
       \!\!1\!+\!O\left(\frac{(K+J)^3}{N}\right)
       \right),
       \widetilde S^{(0)}
       =
       \frac12
       \left(
       \!\!1\!+\!O\left(\frac{(K+J)^3}{N}\right)
       \right).
       \label{Eq: S0}
    \end{align}

    \item \textbf{Outputs.}
    Suppose \(\pit\in\esb\) satisfies the high-probability event in
    \Cref{lem: concen-P}. For any \(l'\neq l\), the initial outputs are
    \begin{align*}
    F^{(0)}_{(l-1)K+k}
    =
    \frac{1}{K+L}
    \left(
    1+O\left(\frac{\log K}{K}\right)
    \right),
    F^{(0)}_{LK+1}
    =
    \frac{1}{K+L}
    \left(
    1+O\left(\frac{\log K}{K}\right)
    \right).
    \end{align*}
    For other question $l' \neq l$, the other-question output probability is $F^{(0)}_{(l'-1)K+k}=
    (
    1+O((\log K)/K)
    )/(K(K+L)).$
    Summing the same-question and other-question probabilities yields
    \[
        \Phi_l^{(0)}
        =
        \frac{K}{K+L}
        \left(
        1+O\left(\frac{\log K}{K}\right)
        \right),
        \qquad
        \Phi_{-l}^{(0)}
        =
        \frac{L-1}{K+L}
        \left(
        1+O\left(\frac{\log K}{K}\right)
        \right).
    \]

    \item \textbf{Attention Logits Update.}
    For every \(k\in[K]\), the informative-token logit update is
    \[
    \beta_{l,k}^{(0)}
    =
    \frac{\log K}{LK}
    \left[
    p_{l,k}
    +
    \frac{L-1}{2(K+L)}
    -
    \frac{L}{2K(K+L)}
    +
    o\left(\frac1K\right)
    \right].
    \]
    For the context token, the corresponding update is
    \[
    \widetilde\beta_{l,j}^{(0)}
    =
    -
    \frac{\log K}{LJ}
    \left[
    \frac{L+1}{2(K+L)}
    +
    \frac{L}{2K(K+L)}
    +
    o\left(\frac1K\right)
    \right]
    \le 0.
    \]
\end{itemize}
\end{lemma}

\begin{proof}
The attention-logit result follows directly from the initialization
\(\W^{(0)}=\mathbf 0_{d\times d}\). The attention-score estimates follow from
\Cref{lem: concen-P}.

We next compute the output probabilities. Conditional on \(X_q=u_l\), the logits of output
before softmax are
\[
    2\log K
    \left(
    \sum_{k=1}^K S_{l,k}a_{l,k}^\top
    +
    (2\widetilde S-1)\widetilde a^\top
    -
    (1-\widetilde S)A_{-l}
    \right),
\]
where \(A_{-l}=\sum_{l'\neq l}\sum_{k'=1}^K a_{l',k'}\). At initialization,
the attention-score estimates are
\(S_{l,k}^{(0)}=(1+o(1))/(2K)\) and
\(\widetilde S^{(0)}=(1+o(1))/2\). These estimates imply the
same exponential factor corresponding to different informative tokens
\(K^{2S_{l,k}^{(0)}}=1+O\left((\log K)/K\right)\) and the context
exponential factor \(K^{2(2\widetilde S^{(0)}-1)}=1+o(1)\). The other-question
exponential factor is \(K^{-2(1-\widetilde S^{(0)})}=K^{-1}(1+o(1))\).
The softmax denominator is therefore calculated as
\begin{align*}
    & \sum_{k=1}^K K^{2S_{l,k}^{(0)}}  + K^{2(2\widetilde S^{(0)}-1)} + (L-1)K\cdot K^{-2(1-\widetilde S^{(0)})}\\
    & \qquad 
    =  K\big(1+O\big(\frac{\log K}{K}\big)\big)
    +
    1
    + o(1)
    +
    (L-1)K\cdot K^{-1}(1+o(1))\\
    & \qquad
    =
    K+L+O(\log K).
\end{align*}
Dividing by this denominator yields the output probabilities. The estimates for
\(\Phi_l^{(0)}\) and \(\Phi_{-l}^{(0)}\) follow by summing the corresponding
output probabilities.

We now compute the logit updates. With \(c_{\lm}=2\), \Cref{lem:logit-up} gives
\begin{align*}
\beta_{l,k}^{(0)}
&=
-\frac{2\log K}{L}
\mathbb E_{\pit}
\Bigg[
S_{l,k}^{(0)}
\Bigg(
F_{(l-1)K+k}^{(0)}-p_{l,k}
-
\sum_{k'=1}^K
S_{l,k'}^{(0)}
\big(
F_{(l-1)K+k'}^{(0)}-p_{l,k'}
\big)
\\
&\hspace{12em}
-
\big(
2F_{LK+1}^{(0)}+\Phi_{-l}^{(0)}
\big)
\widetilde S^{(0)}
\Bigg)
\Bigg|X_q=u_l
\Bigg].
\end{align*}
The initialization estimates for the attention scores are
\(S_{l,k}^{(0)}=(1+o(1))/(2K)\) and
\(\widetilde S^{(0)}=(1+o(1))/2\). The corresponding output estimates are
\(F_{(l-1)K+k}^{(0)}
=F_{LK+1}^{(0)}
=(1+o(1))/(K+L)\).
Together, these estimates give the term
\[
\sum_{k'=1}^K
S_{l,k'}^{(0)}
\big(
F_{(l-1)K+k'}^{(0)}-p_{l,k'}
\big)
=
\frac1{2K}
\left(
\frac{K}{K+L}-1
\right)(1+o(1))
=
-\frac{L}{2K(K+L)}(1+o(1)).
\]
The effect from incorrect answers is
\(\big(2F_{LK+1}^{(0)}+\Phi_{-l}^{(0)}\big)\widetilde S^{(0)}
=(L+1)(1+o(1))/(2(K+L))\).
Substituting these estimates into the expression for \(\beta_{l,k}^{(0)}\) gives
\begin{align*}
\beta_{l,k}^{(0)}
&=
-\frac{2\log K}{L}\cdot \frac1{2K}
\left[
\frac1{K+L}
-
p_{l,k}
+
\frac{L}{2K(K+L)}
-
\frac{L+1}{2(K+L)}
+
o\left(\frac1K\right)
\right]
\\
&=
\frac{\log K}{LK}
\left[
p_{l,k}
+
\frac{L-1}{2(K+L)}
-
\frac{L}{2K(K+L)}
+
o\left(\frac1K\right)
\right].
\end{align*}

For the context logit, \Cref{lem:logit-up} gives
\begin{align*}
\widetilde\beta_{l,j}^{(0)}
&=
-\frac{2\log K}{L}
\mathbb E_{\pit}
\Bigg[
\widetilde S_j^{(0)}
\Bigg(
\big(
2F_{LK+1}^{(0)}+\Phi_{-l}^{(0)}
\big)
(1-\widetilde S^{(0)})
\\
&\hspace{12em}
-
\sum_{k'=1}^K
S_{l,k'}^{(0)}
\big(
F_{(l-1)K+k'}^{(0)}-p_{l,k'}
\big)
\Bigg)
\Bigg|X_q=u_l
\Bigg]
\\
&=
-
\frac{\log K}{LJ}
\left[
\frac{L+1}{2(K+L)}
+
\frac{L}{2K(K+L)}
+
o\left(\frac1K\right)
\right].
\end{align*}
This completes the proof of the initialization estimates.
\end{proof}

We next calculate the variables after one step of gradient descent.

\begin{lemma}[Variables after One Step Gradient Descent]\label{lem:sft-phase1-end-var}
Under the same condition as \Cref{lem:phase1-start-var}, suppose
\(\pit\in\esb\) and choose the first-step learning rate
\(\eta_{\sft}^{(0)}=KLc^{(\mathrm I)}/p_{l,1}\).
Furthermore, for all question $l \in [L]$, we assume $c^{(\mathrm I)}$ is large enough so that $c^{(\mathrm I)}
    K
    [
    (L+1)/(2(K+L))
    +
    L/(2K(K+L))
    ]
    \ge 2Jp_{l,1}, c^{(\mathrm{I})}(1-(\max_{k\geq 2} p_{l,k})/p_{l,1}) \gg 1$.
After one gradient-descent update, the key variables for all questions $l$ satisfy the following
estimates.
\begin{itemize}[leftmargin=8pt]
    \item \textbf{Attention Logits.} The logit corresponding to the head answer
    satisfies \(B_{l,1}^{(1)}=c^{(\mathrm I)}\log K(1+o(1))\).
    For tails \(2\le k\le K\), the logits satisfy $B_{l,k}^{(1)}
        =
        c^{(\mathrm I)}\log K\left(p_{l,k}
+
(L-2)/(2(K+L))
+
O\left(1/K\right)\right)/p_{l,1}$.
    The context logit satisfies
    $\widetilde B_{l,j}^{(1)}
        =
        -
        c^{(\mathrm I)}K\log K
        \left[
        (L+2)/(2(K+L))
        +
        O\left(1/K\right)
        \right]/Jp_{l,1}$.

    \item \textbf{Attention Scores.}
    The logits imply that the context-token contribution is lower order, so the attention scores satisfy
    \[
        S_{l,1}^{(1)}
        =
        1-O\left(K^{1-c^{(\mathrm I)}(1-\max_{k\geq 2} p_{l,k}/p_{l,1}) }\right),
        \qquad
        S_{l,k}^{(1)}
        =
        K^{-c^{(\mathrm I)}(1-\max_{k\geq 2} p_{l,k}/p_{l,1})+o(1)},
        \quad k\ge2.
    \]
    The context attention satisfies
    \(\widetilde S_j^{(1)}=o(S_{l,k}^{(1)})\) and
    \(\widetilde S^{(1)}=o(1)\).

    \item \textbf{Outputs.}
    Under the language model head with \(c_{\lm}=2\), the head-correct answer prediction
    satisfies \(1-F_{(l-1)K+1}^{(1)}=(1+o(1))/K\).
    For every non-head same-question correct answer \(k\ge2\), including both covered
    tails and uncovered answers,
    \(F_{(l-1)K+k}^{(1)}=K^{-2+o(1)}\).
    The remaining incorrect answer output probabilities satisfy
    \(F_{LK+1}^{(1)}=K^{-4+o(1)}\) and
    \(F_{(l'-1)K+k}^{(1)}=K^{-4+o(1)}\) for all \(l'\neq l\) and \(k\in[K]\).
\end{itemize}
\end{lemma}

\begin{proof}
From \Cref{lem:phase1-start-var} and the choice
\(\eta_{\sft}^{(0)}=KLc^{(\mathrm I)}/p_{l,1}\), the head-answer logit is
\(B_{l,1}^{(1)}
=B_{l,1}^{(0)}
+\eta_{\sft}^{(0)}\beta_{l,1}^{(0)}
=c^{(\mathrm I)}\log K(1+o(1))\).
For covered tails \(2\le k\le K\), we obtain
\[
    B_{l,k}^{(1)}
    =
    \frac{c^{(\mathrm I)}\log K}{p_{l,1}}\left(p_{l,k}
+
\frac{L-2}{2(K+L)}
+
O\left(\frac1K\right)\right).
\]
The estimate for \(\widetilde B_{l,j}^{(1)}\) follows similarly by applying
\Cref{lem:phase1-start-var} to the context-token update.

Because \(p_{l,1}\gg L/K\), the non-head logits are of lower order than
\(B_{l,1}^{(1)}\), and therefore
\[
    B_{l,1}^{(1)}-\max_{k\ge2}B_{l,k}^{(1)}
    =
    c^{(\mathrm I)}\log (K) \cdot \bigg(1-\frac{\max_{k\geq 2} p_{l,k}}{p_{l,1}}+o(1)\bigg).
\]
The context-token contribution to the attention denominator is also lower order
than the non-head informative-token contribution. Applying the high-probability
count bounds in \Cref{lem: concen-P}, we obtain
\[
    S_{l,1}^{(1)}
    =
    \frac{K^{c^{(\mathrm I)}(1-\frac{\max_{k\geq 2} p_{l,k}}{p_{l,1}})+o(1)}}
    {K^{c^{(\mathrm I)}(1-\frac{\max_{k\geq 2} p_{l,k}}{p_{l,1}})+o(1)}+K+o(1)}.
\]
The non-head attention scores satisfy, for every \(k\ge2\),
\[
    S_{l,k}^{(1)}
    =
    \frac{1+o(1)}
    {K^{c^{(\mathrm I)}(1-\frac{\max_{k\geq 2} p_{l,k}}{p_{l,1}})+o(1)}+K}.
\]
The gap between the context logits and the largest non-head informative logit satisfies
\[
    \widetilde B_{l,j}^{(1)}-\max_{k\ge2}B_{l,k}^{(1)}
    =
    \frac{c^{(\mathrm I)}\log K}{p_{l,1}}(-\frac{K(L+2)}{2J(K+L)}-{\max_{k\geq 2} p_{l,k}}-\frac{L-2}{2(K+L)}+o(\frac{1}{K})).
\]
This gap implies the context-attention bound
\begin{align*}
    \widetilde S^{(1)} = \frac{K^{\frac{c^{(\mathrm I)}}{p_{l,1}}(-\max_{k\geq 2} p_{l,k}-\frac{L+2}{K+L})+o(\frac{1}{K})}}
    {K^{c^{(\mathrm I)}(1-\frac{\max_{k\geq 2} p_{l,k}}{p_{l,1}})+o(1)}+K+o(1)} = K^{\frac{c^{(\mathrm I)}}{p_{l,1}}(-\max_{k\geq 2} p_{l,k}-\frac{L+2}{K+L})+o(\frac{1}{K})}\max_{k\ge 2}S_{l,k}^{(1)},
\end{align*}
which proves the lemma's claim on attention scores.

The remaining output estimates follow from the same softmax calculation used in the initialization step of \Cref{lem:phase1-start-var}. The head and non-head correct answer factors are
\[
    K^{2S_{l,1}^{(1)}}=K^{2-o(1)},
    \qquad
    K^{2S_{l,k}^{(1)}}=1+o(1),
    \quad k\ge2.
\]
The context and other-question factors are
\[
    K^{2(2\widetilde S^{(1)}-1)}=K^{-2+o(1)},
    \qquad
    K^{-2(1-\widetilde S^{(1)})}=K^{-2+o(1)}.
\]
The head-answer probability is therefore
\(F_{(l-1)K+1}^{(1)}=1-(1+o(1))/K\).
The non-head correct answer probabilities satisfy, for every \(k\ge2\),
\(F_{(l-1)K+k}^{(1)}=K^{-2+o(1)}\).
The wrong-answer and other-question estimates follow similarly, completing the proof.
\end{proof}

\begin{proof}[Proof of \Cref{thm: indist}]
    \Cref{lem:sft-phase1-end-var} directly gives the residual same-question mass estimate
    \begin{align*}
    1-    \Phi_l^{(1)}
    =
    1-\sum_{k=1}^K F_{(l-1)K+k}^{(1)}
    =
    LK^{-3+o(1)}+O\left(\frac{L}{K^3}\right).
    \end{align*}
    This completes the SFT part of the proof.
\end{proof}

\subsection{Useful Concentration Bound}
Recall that, given a prompt $\pit = (X_1^\top, X_2^\top, \ldots, X_N^\top)^\top$, the tokens $X_n$ are independent and identically distributed. As in \cite{huang2024context,cheng2026demystifying}, for any question $l \in [L]$, the occurrence counts of the \((l,k)\)-th informative token and the \(j\)-th context token in $\pit$, denoted by $|\cV_{l,k}|$ and $|\widetilde \cV_j|$, follow a multinomial distribution. We use this multinomial concentration property to define a high-probability event for the prompt.

We first introduce the following tail bound for multinomial distributions.
\begin{lemma}[Tail Bound of Multinomial Distribution~\citep{devroye1983equivalence}]\label{app: lem:tail}
 Let $\left(X_1, \cdots, X_K\right)$ be a multinomial $\left(N, p_1, \cdots, p_K\right)$ random vector. For all $\varepsilon \in(0,1)$ and all $K$ satisfying $ K/ N \leq \varepsilon^2 / 20$, we have
$$
P\left(\sum_{i=1}^K\left|X_i-\mathbb{E}\left(X_i\right)\right|>N \varepsilon\right) \leq 3 \exp \left(-N \varepsilon^2 / 25\right).
$$
\end{lemma}

\begin{lemma}[High-probability Event for $\pit$]\label{lem: concen-P}
  For any question $l \in [L]$, suppose that the probability of $l,k$-th informative token $p_{l,k}=\Theta\left({1}/{2K}\right)$ for any $k\in[K]$, and the probability of context token $\widetilde p_j=\Theta\left({1}/{2J}\right)$ for any $j\in[J]$ and  $(K+J)^3\ll  N$. For some constant  $\sqrt{{40 (K+J)^3}/{N}}\geq c\geq \sqrt{{20 (K+J)^3}/{N}} $, define
\begin{align*}
   \esb=\Big\{\pit|X_q=u_l: \text{ for all } l\in[L], \;
   &
   |\cV_{l,k}|\in \Big[p_{l,k} N-\frac{c N}{K+J}, p_{l,k} N+\frac{c N}{K+J}\Big]\text{ for }k\in[K]\\
   & 
   |\widetilde\cV_j|\in \Big[\widetilde p_j N-\frac{c N}{K+J}, \widetilde p_j N+\frac{c N}{K+J}\Big]\text{ for }j\in[J]
   \Big\}. 
\end{align*}
Under these assumptions, the prompt belongs to this event with high probability, i.e., 
\(\mathbb{P}(\pit\in\esb )
\geq 1-3 \exp \left(-c^2 N/(25 L(K+J)^2)\right)\).
Let $\Lba_{l,k}=p_{l,k} K-cK/(K+J)$, $\Uba_{l,k}= p_{l,k} K+cK/(K+J)$, $\widetilde\Lba_j=\widetilde p_j J-cJ/(K+J)$, and $\widetilde\Uba_j= \widetilde p_j J+cJ/(K+J)$. Since $p_{l,k}=\Theta\left({1}/{K}\right)$ and $\widetilde p_j=\Theta\left({1}/{J}\right)$, these lower and upper constants are all of constant order. Therefore, for any $\pit\in\esb$, we have $|\cV_{l,k}|\in[{\Lba_{l,k}N}/{K},{\Uba_{l,k}N}/{K}]=\Theta({N}/{K})$ and $|\widetilde\cV_j|\in[{\widetilde\Lba_jN}/{J},{\widetilde\Uba_jN}/{J}]=\Theta({N}/{J})$. The constant $c$ can be chosen so that $\Lba_{l,k}> 0$ for all $k\in[K],l \in [L]$ and $\widetilde\Lba_j> 0$ for all $j\in[J]$.

\end{lemma}

\section{Proofs of RL}\label{app:proof_rl}

\subsection{Gradient Computation for RL}\label{app:rl-gradient}
In this section, we analyze the training dynamics of RL. The proof follows a similar structure to the SFT analysis in Appendix \ref{app:proof_sft}, but the dynamics are different in an important way. In SFT, the correct answer distribution is long-tailed, and therefore, the head answer receives a much larger initial positive update. In contrast, the RL objective rewards every \emph{correct} answer symmetrically. This symmetry makes all informative tokens receive nearly the same positive update, while the context tokens, which correspond to the incorrect answer, receive a negative update.

Throughout this section, we fix a question $l\in[L]$ and condition on $X_q=u_l$. Similar to SFT, we write \(\pit=(X_1,\ldots,X_N)\in\mathbb R^{N\times d}\) for the prompt tokens, and define \(S=\Soft(X_q^\top \W \pit^\top)\in\mathbb R^{1\times N}\) and \(F=\Soft(S\pit W_{\lm})\in\mathbb R^{1\times (LK+1)}\).
Similar to SFT, conditional on $X_q=u_l$, the output can be written as
\[
    F
    =
    \Soft\left(
    c_{\lm}\log K
    \left(
    \sum_{k=1}^K S_{l,k}a_{l,k}^\top
    +
    \widetilde S\widetilde a^\top
    -
    (1-\widetilde S)
    \sum_{l'\neq l}\sum_{k'=1}^K a_{l',k'}^\top
    \right)
    \right).
\]
First, we calculate the gradient as follows.
\begin{lemma}[RL Gradient]\label{lem:rl-gradient}
The gradient of the RL objective with respect to $\W$ is
\[
    \nabla_\W J_{\mathrm{RL}}
    =
    \mathbb E_{\pit,X_q}
    \!\left[
    \sum_{l=1}^L
    \mathbbm 1\{X_q=u_l\}
    X_q
    \Big(
        \big(F\odot(A_l-\Phi_l\mathbf 1_{LK+1})\big)
        W_{\lm}^\top
        \pit^\top
        \big(\diag(S)-S^\top S\big)
    \Big)
    \!\pit
    \!\right]\!,
\]
where $\mathbf 1_{LK+1}\in\mathbb R^{1\times (LK+1)}$ is the all-one vector and
$\odot$ denotes entrywise product.
\end{lemma}

\begin{proof}

For a single sample $(\pit,X_q)$, define the intermediate variables \(s = X_q^\top \W \pit^\top \in \mathbb R^{1\times N}\) and \(S = \Soft(s)\in \mathbb R^{1\times N}\), and define the output-logit variables \(z = S\pit W_{\lm} \in \mathbb R^{1\times (LK+1)}\) and \(F = \Soft(z)\in \mathbb R^{1\times (LK+1)}\). Here $\pit\in\mathbb R^{N\times d}$, $X_q\in\mathbb R^{d\times 1}$, $\W\in\mathbb R^{d\times d}$, and $W_{\lm}\in\mathbb R^{d\times (LK+1)}$.

For each question $l\in[L]$, define the question-specific correct-answer indicator vector \(A_l=\sum_{k=1}^K a_{(l-1)K+k}\in \mathbb R^{1\times (LK+1)}\). The total probability assigned to the correct answers to question $l$ is \(\Phi_l=\sum_{k=1}^K F_{(l-1)K+k}=F A_l^\top\). Conditional on $X_q=u_l$, the single-sample RL objective is \(j_{\mathrm{RL}}(\W;\pit,X_q)=\Phi_l=\sum_{k=1}^K F_{(l-1)K+k}\).
Therefore, the population RL objective is 
\(J_{\mathrm{RL}}(\W)
=\mathbb E_{\pit,X_q}
\left[
\sum_{l=1}^L
\mathbbm 1\{X_q=u_l\}\Phi_l
\right]\). 
We compute the gradient for one sample $(\pit,X_q)$ conditional on $X_q=u_l$. The unconditional gradient follows by multiplying by $\mathbbm 1\{X_q=u_l\}$ and taking expectation over $l$.

We first differentiate with respect to the output logits $z$. Recall that \(F=\Soft(z)\) and \(\Phi_l=\sum_{k=1}^K F_{(l-1)K+k}\). For any coordinate $m\in[LK+1]$, the softmax derivative is
\((\partial F_r)/(\partial z_m)
=F_r\big(\mathbbm 1\{r=m\}-F_m\big)\).
Therefore, the derivative of the correct-answer probability with respect to each output logit is
\[
\begin{aligned}
    \frac{\partial \Phi_l}{\partial z_m}
    &=
    \sum_{k=1}^K
    \frac{\partial F_{(l-1)K+k}}{\partial z_m} \\
    &=
    \sum_{k=1}^K
    F_{(l-1)K+k}
    \big(
        \mathbbm 1\{(l-1)K+k=m\}-F_m
    \big) \\
    &=
    F_m\mathbbm 1\{m\in\{(l-1)K+1,\ldots,lK\}\}
    -
    F_m
    \sum_{k=1}^K F_{(l-1)K+k} \\
    &=
    F_m
    \left(
        (A_l)_m
        -
        \Phi_l
    \right).
\end{aligned}
\]
Equivalently, the output-logit gradient in row-vector form is
\(g_z=\nabla_z j_{\mathrm{RL}}
=F\odot(A_l-\Phi_l\mathbf 1_{LK+1})
\in\mathbb R^{1\times (LK+1)}\).

Next, we differentiate with respect to the attention vector $S$. The output logits satisfy \(z=S\pit W_{\lm}\), so the gradient with respect to the attention vector is
\(g_S=\nabla_S j_{\mathrm{RL}} =g_z W_{\lm}^\top \pit^\top\in\mathbb R^{1\times N}\). 
Furthermore, we differentiate through the inner softmax. The inner-softmax relation is \(S=\Soft(s)\), so the gradient with respect to the attention logits is \(g_s=\nabla_s j_{\mathrm{RL}}
=g_S\big(\diag(S)-S^\top S\big)\in\mathbb R^{1\times N}\).
Substituting the expression for $g_S$ gives the explicit attention-logit gradient
\[
    g_s
    =
    \big(F\odot(A_l-\Phi_l\mathbf 1_{LK+1})\big)
    W_{\lm}^\top
    \pit^\top
    \big(\diag(S)-S^\top S\big).
\]

We then differentiate with respect to $\W$. Recall that \(s=X_q^\top \W \pit^\top\), whose corresponding differential is \(\mathrm{d}s=X_q^\top\,\mathrm{d}\W\,\pit^\top\). Together, these identities give the single-sample gradient
\(\nabla_\W j_{\mathrm{RL}}=X_q g_s\pit\).
Substituting the expression for $g_s$ gives the explicit single-sample gradient
\[
    \nabla_\W j_{\mathrm{RL}}
    =
    X_q
    \Big(
        \big(F\odot(A_l-\Phi_l\mathbf 1_{LK+1})\big)
        W_{\lm}^\top
        \pit^\top
        \big(\diag(S)-S^\top S\big)
    \Big)
    \pit.
\]

Finally, taking expectation over $(\pit,X_q)$ and summing over the possible questions gives
\begin{align*}
    & \nabla_\W J_{\mathrm{RL}}(\W)
    =
    \mathbb E_{\pit,X_q}
    \left[
    \sum_{l=1}^L
    \mathbbm 1\{X_q=u_l\}
    X_q
    \Big(
        \big(F\odot(A_l-\Phi_l\mathbf 1_{LK+1})\big)
        W_{\lm}^\top
        \pit^\top
        \big(\diag(S)-S^\top S\big)
    \Big)
    \pit
    \right].   
\end{align*}
This completes the proof of RL gradient.
\end{proof}

Similar to the SFT analysis, we overload the notation and define the important logits and their corresponding update rates for RL.
\begin{definition}[Important Logits and Their Update Rate in RL]\label{def:rl-logits}
For any $l \in [L]$, $k \in [K]$ and $j \in [J]$, for iteration $t\geq 0$, we denote the weight as $\W^{(t)}$ and the loss as $J_{\mathrm{RL}}^{(t)}=J_{\mathrm{RL}}(\W^{(t)})$. We define the following  quantities 
\begin{align*}
\mathrm{Attention\;Logits:}\;B_{l,k}^{(t)} &= u_l \W^{(t)}  i_{l,k}^\top, \qquad \widetilde B_{l,j}^{(t)} = u_l \W^{(t)}  c_j^\top; \\    
\mathrm{Logits\;Update\;Rate:}\; \beta_{l,k}^{(t)} & = u_l \nabla_\W J_{\mathrm{RL}}^{(t)}\, i_{l,k}^\top, \qquad \widetilde\beta_{l,j}^{(t)} = u_l \nabla_\W J_{\mathrm{RL}}^{(t)}\, c_j^\top.
\end{align*}
The GD update gives
\begin{align*}
   B_{l,k}^{(t+1)}&= B_{l,k}^{(t)}+\eta_\rl \beta_{l,k}^{(t)},\quad
   \widetilde B_{l,j}^{(t+1)}= \widetilde B_{l,j}^{(t)}+\eta_\rl \widetilde\beta_{l,j}^{(t)}.
\end{align*}
The initialization $\W^{(0)}=\mathbf{0}_{d\times d}$ also gives $B_{l,k}^{(0)}=\widetilde B_{l,j}^{(0)}=0$.    
\end{definition}

\begin{lemma}[Logits Update Rate for RL]\label{lem:rl-logit-up}
For any $l\in[L]$, $k\in[K]$ and $j\in[J]$, the RL update rates of the attention logits can be written as
\begin{align*}
\beta_{l,k}
&=
\frac{c_{\lm}\log(K)}{L}
\mathbb{E}_{\pit}\Big[
S_{l,k}
\Big(
(1-\Phi_l)
\Big(
F_{(l-1)K+k}
-
\sum_{k'=1}^K S_{l,k'}F_{(l-1)K+k'}
\Big)
\\
&\hspace{7em}
+
\Phi_l(2F_{LK+1}+\Phi_{-l})\widetilde S
\Big)
\Big|X_q=u_l
\Big],
\\
\widetilde\beta_{l,j}
&=
-\frac{c_{\lm}\log(K)}{L}
\mathbb{E}_{\pit}\Big[
\widetilde S_j
\Big(
(1-\Phi_l)\sum_{k'=1}^K S_{l,k'}F_{(l-1)K+k'}
\\
&\hspace{7em}
+
\Phi_l(2F_{LK+1}+\Phi_{-l})(1-\widetilde S)
\Big)
\Big|X_q=u_l
\Big].
\end{align*}
\end{lemma}

\begin{proof}
The proof follows the same calculation as \Cref{lem:logit-up}, but with the RL output-gradient vector
\(g_z=F\odot(A_l-\Phi_l\mathbf 1_{LK+1})\).
Recall from \Cref{lem:rl-gradient} that, conditional on $X_q=u_l$,
\[
    \nabla_\W J_{\mathrm{RL}}
    =
    \frac1L
    \mathbb E_{\pit}\!\left[
    u_l^\top
    \Big(
    g_z W_{\lm}^\top \pit^\top
    (\diag(S)-S^\top S)
    \Big)
    \pit
    \Big|X_q=u_l
    \right].
\]
The language-model head is
\[
    W_{\lm}
    =
    c_{\lm}\log(K)
    \left(
    \sum_{l'=1}^L\sum_{k'=1}^K
    i_{l',k'}^\top(a_{l',k'}-A_{-l'}-\widetilde a)
    +
    \sum_{j=1}^J c_j^\top\widetilde a
    \right).
\]
Substituting this matrix into the shared gradient factor gives
\begin{align}
&W_{\lm}^\top\pit^\top(\diag(S)-S^\top S)\pit
\nonumber\\
&\quad=
c_{\lm}\log(K)
\left(
\sum_{l'=1}^L\sum_{k'=1}^K
S_{l',k'}(a_{l',k'}-A_{-l'}-\widetilde a)^\top i_{l',k'}
+
\widetilde a^\top
\sum_{n=1}^N\mathbbm 1\{X_n\in\mathcal C\}S_nX_n
\right)
\nonumber\\
&\quad\qquad
-
c_{\lm}\log(K)
\left(
\sum_{l'=1}^L\sum_{k'=1}^K
S_{l',k'}(a_{l',k'}-A_{-l'}-\widetilde a)^\top
+
\widetilde S\widetilde a^\top
\right)
\left(
\sum_{n=1}^N S_nX_n
\right).
\label{eq:rl-track-new-WO}
\end{align}
When $X_q=u_l$, we have $S_{l',k'}=0$ for all $l'\neq l$. The informative-token factor is therefore
\begin{align*}
&W_{\lm}^\top\pit^\top(\diag(S)-S^\top S)\pit\, i_{l,k}^\top
\\
&\quad=
c_{\lm}\log(K)S_{l,k}
\left(
(a_{l,k}-A_{-l}-\widetilde a)^\top
-
\sum_{k'=1}^K S_{l,k'}(a_{l,k'}-A_{-l}-\widetilde a)^\top
-
\widetilde S\widetilde a^\top
\right).
\end{align*}
The identity $\sum_{k'=1}^K S_{l,k'}=1-\widetilde S$ simplifies this factor to
\[
c_{\lm}\log(K)S_{l,k}
\left(
a_{l,k}^\top
-
\sum_{k'=1}^K S_{l,k'}a_{l,k'}^\top
-
\widetilde S A_{-l}^\top
-
2\widetilde S\widetilde a^\top
\right).
\]
The required inner products with the output gradient vector $g_z=F\odot(A_l-\Phi_l\mathbf 1_{LK+1})$ are
\(g_za_{l,k}^\top=(1-\Phi_l)F_{(l-1)K+k}\),
\(g_zA_{-l}^\top=-\Phi_l\Phi_{-l}\), and
\(g_z\widetilde a^\top=-\Phi_lF_{LK+1}\).
Substituting these inner products gives the informative-token update rate
\begin{align*}
\beta_{l,k}
&=
\frac{c_{\lm}\log(K)}{L}
\mathbb{E}_{\pit}\Big[
S_{l,k}
\Big(
(1-\Phi_l)
\Big(
F_{(l-1)K+k}
-
\sum_{k'=1}^K S_{l,k'}F_{(l-1)K+k'}
\Big)
\\
&\hspace{8em}
+
\Phi_l(2F_{LK+1}+\Phi_{-l})\widetilde S
\Big)
\Big|X_q=u_l
\Big].
\end{align*}

For the context token $c_j$, the corresponding factor is
\begin{align*}
&W_{\lm}^\top\pit^\top(\diag(S)-S^\top S)\pit\, c_j^\top
\\
&\quad=
c_{\lm}\log(K)\widetilde S_j
\left(
\widetilde a^\top
-
\sum_{k'=1}^K S_{l,k'}(a_{l,k'}-A_{-l}-\widetilde a)^\top
-
\widetilde S\widetilde a^\top
\right)
\\
&\quad=
c_{\lm}\log(K)\widetilde S_j
\left(
-\sum_{k'=1}^K S_{l,k'}a_{l,k'}^\top
+
(1-\widetilde S)A_{-l}^\top
+
2(1-\widetilde S)\widetilde a^\top
\right).
\end{align*}
Taking inner product with $g_z$ gives
\begin{align*}
\widetilde\beta_{l,j}
&=
-\frac{c_{\lm}\log(K)}{L}
\mathbb{E}_{\pit}\Big[
\widetilde S_j
\Big(
(1-\Phi_l)\sum_{k'=1}^K S_{l,k'}F_{(l-1)K+k'}
\\
&\hspace{8em}
+
\Phi_l(2F_{LK+1}+\Phi_{-l})(1-\widetilde S)
\Big)
\Big|X_q=u_l
\Big].
\end{align*}
This proves the lemma.
\end{proof}

\subsection{Learning Dynamics of RL}
With the gradient expressions derived in Appendix \ref{app:rl-gradient}, we analyze the training dynamics of RL in this section. For each question class $l\in[L]$, we fix $l$ and condition on $\{X_q=u_l\}$. Similar to SFT training, when $X_q=u_l$, only informative tokens from $\mathcal I_l$ and context tokens from $\mathcal C$ appear in the prompt, so $S_{l',k'}=0$ for all $l'\neq l$.

Different from SFT, the RL objective is symmetric over all correct answers to the same question. All informative tokens, therefore, receive the same leading positive update, while context tokens receive a negative update. This symmetry means that RL does not concentrate attention on the head answer; instead, it learns all correct informative tokens nearly uniformly.

Under the language-model head with $c_{\lm}=2$, the head matrix is
\[
    W_{\lm}
    =
    2\log(K)
    \left(
    \sum_{l'=1}^L \sum_{k'=1}^K
    i_{l',k'}^\top(a_{l',k'}-A_{-l'}-\widetilde a)
    +
    \sum_{j=1}^J c_j^\top \widetilde a
    \right).
\]
Conditional on $X_q=u_l$, the output can be written as
\[
    F
    =
    \Soft\left(
    2\log K
    \left(
    \sum_{k=1}^K S_{l,k}a_{l,k}^\top
    +
    (2\widetilde S-1)\widetilde a^\top
    -
    (1-\widetilde S)A_{-l}
    \right)
    \right),
\]
where \(A_{-l}=\sum_{l'\neq l}\sum_{k'=1}^K a_{l',k'}\) is the other-question answer vector.

We first calculate the important estimates at initialization.
\begin{lemma}[Initialization Variable for RL]\label{lem:rl-phase1-start-var}
At initialization $t=0$, for all $k\in[K]$ and $j\in[J]$, the key variables satisfy the following estimates.
\begin{itemize}[leftmargin=8pt]
    \item \textbf{Attention Logits.} The initialized logits are
    \(B_{l,k}^{(0)}=\widetilde B_{l,j}^{(0)}=0\).

    \item \textbf{Attention Scores.}
    Suppose $\pit\in\esb$ satisfies the high-probability event in
    \Cref{lem: concen-P}. Under this event, the initial token-level attention
    scores satisfy
    \(S_{l,k}^{(0)}
    =\left(1+O\left((K+J)^3/N\right)\right)/(2K)\) and
    \(\widetilde S_j^{(0)}
    =\left(1+O\left((K+J)^3/N\right)\right)/(2J)\).
    The total context attention at initialization is
    \(\widetilde S^{(0)}
    =\left(1+O\left((K+J)^3/N\right)\right)/2\).

    \item \textbf{Outputs.}
    Suppose $\pit\in\esb$. The same-question and context output probabilities satisfy
    \[
    \begin{aligned}
        F_{(l-1)K+k}^{(0)}
        &=
        \frac{1}{K+L}
        \left(1+O\left(\frac{\log K}{K}\right)\right),\quad
        F_{LK+1}^{(0)}
        =
        \frac{1}{K+L}
        \left(1+O\left(\frac{\log K}{K}\right)\right).
    \end{aligned}
    \]
    For every $l'\neq l$, the other-question output probabilities satisfy
    \(F_{(l'-1)K+k}^{(0)}
    =\left(1+O\left((\log K)/K\right)\right)/(K(K+L))\).
    Summing the same-question and other-question output probabilities gives
    \[
    \begin{aligned}
        \Phi_l^{(0)}
        &=
        \frac{K}{K+L}
        \left(1+O\left(\frac{\log K}{K}\right)\right),\quad
        \Phi_{-l}^{(0)}
        =
        \frac{L-1}{K+L}
        \left(1+O\left(\frac{\log K}{K}\right)\right).
    \end{aligned}
    \]

    \item \textbf{Attention Logits Update.}
    For every $k\in[K]$ and $j\in[J]$, the initial logit updates satisfy
    \(\beta_{l,k}^{(0)}
    =\Theta\left((L+1)\log K/(LK^2)\right)>0\) and
    \(\widetilde\beta_{l,j}^{(0)}
    =-\Theta\left((L+1)\log K/(LJK)\right)<0\).
\end{itemize}
\end{lemma}

\begin{proof}
The estimates for attention logits and attention scores, and the calculations of the output probabilities, follow from $\W^{(0)}=\mathbf 0_{d\times d}$ and \Cref{lem: concen-P}, which is similar to the proof of \Cref{lem:phase1-start-var}. It remains to compute the attention logits update.

Then \Cref{lem:rl-logit-up} directly gives the initialization update rates
\(\beta_{l,k}^{(0)}
=\Theta\left((L+1)\log K/(LK^2)\right)>0\) and
\(\widetilde\beta_{l,j}^{(0)}
=-\Theta\left((L+1)\log K/(LJK)\right)<0\).
This completes the proof.
\end{proof}
If the one-step RL learning rate is
\(\eta_{\rl}^{(0)}
=2LK^2c_\epsilon/(L+1)=\Theta\left(LK^2/(L+1)\right)\),
where $c_\epsilon>0$ is a sufficiently large constant. More precisely, we choose $\eta_{\rl}^{(0)}$ so that
\(\eta_{\rl}^{(0)}\beta_{l,k}^{(0)}
=c_\epsilon\log K(1+o(1))\).
After the one-step update, the attention weights on context tokens become negligible, while the attention weights on the $K$ informative tokens remain nearly uniform and large enough as in the following lemma.

\begin{lemma}[Variables after One Step Update for RL]
\label{lem:rl-phase1-end-var}
Under the same condition as \Cref{lem:rl-phase1-start-var}, set
\(b_K=(L+1)\log K/(LK^2)\). Then
\Cref{lem:rl-phase1-start-var} gives
\(\beta_{l,k}^{(0)}=b_K(1+o(1))\) and
\(\widetilde \beta_{l,j}^{(0)}
=-Kb_K(1+o(1))/J\). For any target accuracy \(\epsilon=O(\log(K)/K)\), choose a constant \(c_\epsilon\) satisfying
\(c_\epsilon
\ge J\cdot
\log(C_0\log K/\epsilon)/(K\log K)\), where $C_0>0$ is a sufficiently large universal constant. Set the learning rate to
\(\eta_{\rl}^{(0)}
=2c_\epsilon\log K/b_K
=2LK^2c_\epsilon/(L+1)\).
After one RL update, for every $k\in[K]$ and $j\in[J]$, the key variables satisfy the following estimates.
\begin{itemize}[leftmargin=8pt]
    \item \textbf{Attention Logits.} The logits satisfy
    \(B_{l,k}^{(1)}
    =c_\epsilon\log K(1+o(1))\) and
    \(\widetilde B_{l,j}^{(1)}
    =-(K/J)c_\epsilon\log K(1+o(1))\).

    \item \textbf{Attention Scores.} The informative-token attention scores satisfy $
        \max_{k\in[K]}
        \left|
        S_{l,k}^{(1)}-1/K
        \right|
        \le
        \epsilon/K.$
    The context-token attention scores satisfy
    \(\widetilde S_j^{(1)}
    =O\left(K^{-c_\epsilon K/J}/J\right)\) and
    \(\widetilde S^{(1)}\le\epsilon/(C_0\log K)\).

    \item \textbf{Outputs.} The correct-answer output probabilities satisfy 
    \begin{align*}
        \max_{k\in[K]}
        \left|
        F_{(l-1)K+k}^{(1)}-1/K
        \right|
        \le
        C/K
        \left(
        \epsilon+L/K^2+1/K^3
        \right),
    \end{align*}
    for some universal constant $C>0$. The total probability of correct answers satisfies
    \(\Phi_l^{(1)}
    \ge 1-C\left(\epsilon+L/K^2+1/K^3\right)\).
    The incorrect-answer probabilities satisfy
    \(F_{LK+1}^{(1)}=O(K^{-3})\) and
    \(\Phi_{-l}^{(1)}=O\left(L/K^2\right)\).
\end{itemize}
\end{lemma}

\begin{proof}
\Cref{lem:rl-phase1-start-var} and \Cref{lem:rl-logit-up} give the
initialization-scale updates
\(\beta_{l,k}^{(0)}=b_K(1+o(1))\) and
\(\widetilde \beta_{l,j}^{(0)}
=-(K/J)b_K(1+o(1))\),
where the common scale is \(b_K=(L+1)\log K/(LK^2)\).

The chosen learning rate is \(\eta_{\rl}^{(0)}=2c_\epsilon\log K/b_K\).
With this choice, the informative logits satisfy
\(B_{l,k}^{(1)}
=B_{l,k}^{(0)}
+\eta_{\rl}^{(0)}\beta_{l,k}^{(0)}
=c_\epsilon\log K(1+o(1))\) for all \(k\in[K]\). The context logits satisfy
\(\widetilde B_{l,j}^{(1)}
=\widetilde B_{l,j}^{(0)}
+\eta_{\rl}^{(0)}\widetilde\beta_{l,j}^{(0)}
=-Kc_\epsilon\log K(1+o(1))/J\) for all \(j\in[J]\).

We next control the attention scores. Since all informative logits are equal up to lower-order terms and every context logit is smaller than the informative logits by the gap \(\left(1+K/J\right)c_\epsilon\log K\), the total context attention mass satisfies
\(\widetilde S^{(1)}=O\left(K^{-c_\epsilon K/J}\right)\).
The choice of $c_\epsilon$ gives the context-attention bound
\(\widetilde S^{(1)}
\le K^{-c_\epsilon K/J}
\le\epsilon/(C_0\log K)\).
Since the remaining attention mass is distributed symmetrically over the $K$ informative tokens, each informative-token attention score satisfies
\(S_{l,k}^{(1)}
=(1-\widetilde S^{(1)})(1+o(1))/K\) for all \(k\in[K]\).
Hence, its deviation from the uniform attention score is bounded by
\(\max_{k\in[K]}|S_{l,k}^{(1)}-1/K| \le\epsilon/K\).

It remains to control the output probabilities. With $c_{\lm}=2$, each correct-answer output probability can be written as
\[
F_{(l-1)K+k}^{(1)}
=
\frac{
K^{2S_{l,k}^{(1)}}
}{
\sum_{k'=1}^K K^{2S_{l,k'}^{(1)}}
+
K^{2(2\widetilde S^{(1)}-1)}
+
\sum_{l'\neq l}\sum_{k'=1}^K
K^{-2(1-\widetilde S^{(1)})}
}.
\]
The attention bound gives
\(S_{l,k}^{(1)}
=1/K+O\left(\epsilon/K\right)\) and
\(\widetilde S^{(1)}
\le\epsilon/(C_0\log K)\).
Therefore, the same-question exponential factors satisfy
\[
    K^{2S_{l,k}^{(1)}}
    =
    \exp\left(2S_{l,k}^{(1)}\log K\right)
    =
    1+O\left(\frac{\log K}{K}\right)+O(\epsilon)=1+O(\epsilon),
\]
where the last step uses $\epsilon = O(\log(K)/K)$ and absorbs constants into the $O(\cdot)$ term. The context exponential factor satisfies
\[
    K^{2(2\widetilde S^{(1)}-1)}
    =
    K^{-2}\exp\left(4\widetilde S^{(1)}\log K\right)
    =
    O(K^{-2}).
\]
The other-question exponential factor satisfies
\[
    K^{-2(1-\widetilde S^{(1)})}
    =
    K^{-2}\exp\left(2\widetilde S^{(1)}\log K\right)
    =
    O(K^{-2}).
\]
The resulting softmax denominator is
\[
\begin{aligned}
&\sum_{k'=1}^K K^{2S_{l,k'}^{(1)}}
+
K^{2(2\widetilde S^{(1)}-1)}
+
\sum_{l'\neq l}\sum_{k'=1}^K
K^{-2(1-\widetilde S^{(1)})}
\\
&\qquad =
K(1+O(\epsilon))
+
O(K^{-2})
+
O\left(\frac{L}{K}\right).
\end{aligned}
\]
Dividing the same-question numerator by this denominator gives
\(F_{(l-1)K+k}^{(1)}
=\left(1+O\left(\epsilon\right)\right)/K\).
This output estimate implies
\[
    \max_{k\in[K]}
    \left|
    F_{(l-1)K+k}^{(1)}-\frac1K
    \right|
    \le
    \frac{C\epsilon}{K},
\]
for some constant $C$. Because of the term $(\log K)/K$, the smallest useful choice of $\epsilon$ is of order $O((\log K)/K)$. With this choice, the context-output probability is
\[
    F_{LK+1}^{(1)}
    =
    \frac{K^{2(2\widetilde S^{(1)}-1)}}
    {K(1+O(\epsilon))+O(K^{-2})+O(L/K)}
    =
    O(K^{-3}).
\]
For answers to other questions, each output probability is also $O(K^{-3})$. Since there are $(L-1)K$ such answers,
\(\Phi_{-l}^{(1)}=O\left(L/K^2\right)\).
Finally, the total correct-answer probability is
\[
    \Phi_l^{(1)}
    =
    1-F_{LK+1}^{(1)}-\Phi_{-l}^{(1)}
    \ge
    1
    -
    O\big(\frac{L}{K^2}+\frac{1}{K^3}\big) \geq 1- \epsilon,
\]
This completes the proof.
\end{proof}

\begin{proof}[Proof of \Cref{thm: indist}]
    Following from \Cref{lem:rl-phase1-end-var}, we prove \Cref{thm: indist} immediately. 
\end{proof}

\section{Proof of \Cref{Thm: ood generalization}}\label{app: ood}
\begin{proof}
First, for \textbf{SFT}, the one-step dynamics in
\Cref{lem:sft-phase1-end-var} give the following informative-token and
context-token logits for any \(k\in[K]\):
\begin{align*}
    B_{l,k}^\sft
    &=
    \frac{c^{(\mathrm I)}\log K}{p_{l,1}}
    \left(
        p_{l,k}
        +
        \frac{L-2}{2(K+L)}
        +
        o\left(\frac{1}{K}\right)
    \right), \\
    \widetilde B_{l,j}^\sft
    &=
    -
    \frac{c^{(\mathrm I)}K\log K}{Jp_{l,1}}
    \left(
        \frac{L+2}{2(K+L)}
        +
        o\left(\frac{1}{K}\right)
    \right).
\end{align*}
If \(p_{l,k}\geq 1/\log K\), then the attention assigned to the informative
token is bounded below while the total context attention is bounded above:
\begin{align*}
    S_{l,k}^\sft
    &=
    \frac{(N/2)\exp(B_{l,k}^\sft)}
    {(N/2)\exp(B_{l,k}^\sft)
    +\sum_{j=1}^J |\widetilde{\mathcal V}_j|
    \exp(\widetilde B_{l,j}^\sft)}  \\
    &\geq
    \frac{\exp\!\left(c^{(\mathrm I)}/p_{l,1}\right)}
    {\exp\!\left(c^{(\mathrm I)}/p_{l,1}\right)+1},
    \qquad
    \widetilde S^\sft
    \leq
    \frac{1}
    {\exp\!\left(c^{(\mathrm I)}/p_{l,1}\right)+1}.
\end{align*}
Therefore, the output probability assigned to answer \(k\) by the SFT-trained
model can be written as
\begin{align*}
    F_{l,k}^\sft
    &=
    \frac{K^{2S_{l,k}^\sft}}
    {K^{2S_{l,k}^\sft}
    +(L-1)K\cdot K^{-2S_{l,k}^\sft}
    +K^{2(2\widetilde S^\sft-1)}} \\
    &=
    \frac{K^{4S_{l,k}^\sft-1}}
    {K^{4S_{l,k}^\sft-1}
    +(L-1)
    +K^{1-2S_{l,k}^\sft}}.
\end{align*}
This expression, together with the lower bound on \(S_{l,k}^\sft\), gives the
following bound on the SFT prediction error:
\begin{align*}
    1-F_{l,k}^\sft
    \leq
    \frac{L}{K^{4S_{l,k}^\sft-1}}
    =
    O\left(
        K^{
        1-
        \frac{4\exp(c^{(\mathrm I)}/p_{l,1})}
        {\exp(c^{(\mathrm I)}/p_{l,1})+1}
        }
    \right)
    =
    O\left(\frac{1}{K^3}\right),
\end{align*}
where the last step follows because \(c^{(\mathrm I)}\) can be chosen
sufficiently large.

If \(p_{l,k}\leq 1/K\), then the SFT attention mass is balanced between this tail
informative token and the context tokens, with
\(S_{l,k}^\sft=\widetilde S^\sft=1/2\). With these attention scores,
the same softmax formula gives
\begin{align*}
    F_{l,k}^\sft
    &=
    \frac{K^{2S_{l,k}^\sft}}
    {K^{2S_{l,k}^\sft}
    +(L-1)K\cdot K^{-2S_{l,k}^\sft}
    +K^{2(2\widetilde S^\sft-1)}} \\
    &=
    \frac{K}{K+(L-1)+K}
    \leq
    \frac{1}{2}.
\end{align*}
This proves the claimed behavior for SFT.

On the other hand, for \textbf{RL}, \Cref{lem:rl-phase1-end-var} gives the
symmetric informative-token logits and the context-token logits for all
\(k\in[K]\):
\begin{align*}
    B_{l,k}^\rl
    &=
    c_\epsilon \log K(1+o(1)), \\
    \widetilde B_{l,j}^\rl
    &=
    -\frac{K}{J}c_\epsilon \log K(1+o(1)).
\end{align*}
Thus, the RL attention score for each informative token is
\(S_{l,k}^\rl
=K^{c_\epsilon}/(
K^{-\frac{K}{J}c_\epsilon}+K^{c_\epsilon})\).
Using this attention score in the output softmax gives
\begin{align*}
    F_{l,k}^\rl
    &=
    \frac{K^{2S_{l,k}^\rl}}
    {K^{2S_{l,k}^\rl}
    +(L-1)K\cdot K^{-2S_{l,k}^\rl}
    +K^{2(2\widetilde S^\rl-1)}} \\
    &=
    \frac{K^{4S_{l,k}^\rl-1}}
    {K^{4S_{l,k}^\rl-1}
    +(L-1)
    +K^{1-2S_{l,k}^\rl}}.
\end{align*}
The resulting output-probability expression yields the following RL prediction
error bound:
\begin{align*}
    1-F_{l,k}^\rl
    \leq
    \frac{L}{K^{4S_{l,k}^\rl-1}}
    =
    O\left(
        K^{
        1-
        \frac{4K^{c_\epsilon}}
        {K^{-\frac{K}{J}c_\epsilon}+K^{c_\epsilon}}
        }
    \right)
    =
    O\left(\frac{1}{K^3}\right),
\end{align*}
where the last step follows because \(c_\epsilon\) can be chosen sufficiently
large. Taking the sum over \(k\) and \(l\) completes the proof.
\end{proof}

\section{Discussions}\label{app:discussions}

\subsection{Comparison between Our Setup with Distillation}\label{app:comp_distill}
We clarify our \ac{sft} experimental setup with respect to distillation in this subsection.

We would like to clarify the terminology.
\begin{itemize}[leftmargin=2em]
    \item \ac{sft} refers to fine-tuning an LLM on a curated dataset, which may be constructed in various ways—for example, using responses generated by other models, annotations collected from humans, or data filtered and processed through carefully designed pipelines. 
    \item Distillation~\citep{hinton2015distill}, by contrast, specifically transfers knowledge from a teacher model to a student model on the desired task using the teacher’s outputs, probability distributions, or intermediate representations.
\end{itemize}

According to these terminologies, these two notions share overlap procedures during implementation, e.g., both methods adopt a stronger model to generate the training data.
We believe our methods are better characterized as \ac{sft} for the following reasons:
\begin{itemize}[leftmargin=2em]
    \item We explicitly filter and control the composition of the dataset according to different criteria. In Section~\ref{sec:scale}, we construct the fine-tuning dataset by matching the numbers of correct and incorrect rollouts or prompts. Consequently, the filtered dataset does not follow the original output distribution of the model that generated the rollouts.
    \item Our experiments do not directly learn the target task from the training datasets. Most distillation works directly tune LLMs on the target tasks. However, in our work, we conduct \ac{sft} on the math task and evaluate it on the general reasoning tasks that do not involve math.
    \item We follow the common terminology used in this community. For example, \citet{ye2025limo} constructs its SFT dataset using responses generated by DeepSeek-R1, DeepSeek-R1-Distill-Qwen-32B, and QwQ-32B, and then performs \ac{sft} on the curated dataset. Similarly, \citet{muennighoff2025s1} collects responses from Qwen2.5-32B-Instruct and uses them for \ac{sft}, while \citet{huan2026doesmath} construct an SFT dataset using responses from Qwen3-32B-Instruct. Since these papers characterize their training procedures as SFT, we follow the same terminology.
\end{itemize}

\subsection{Ablation on the Incorrect Rollouts}

The \ac{sft} loss is applied to the entire trace during \ac{sft}, including the wrong rollouts in Section~\ref{sec:scale}. We find that the wrong reasoning rollouts can still be helpful:
\begin{itemize}[leftmargin=2em]
    \item \textbf{Most rollouts with wrong final answers contain partially correct reasoning steps, which potentially benefit SFT.} Specifically, we adopt the capable Qwen3-32B and Qwen2.5-32B-Instruct to investigate the correctness of the intermediate traces of wrong rollouts. We randomly sample 10,000 wrong rollouts from each dataset and prompt each model to investigate the correctness of the intermediate traces. The models output JSON files which give $\{$yes, mostly yes, partial, no, unclear$\}$ to grade the correctness of the intermediate traces, where $\{$yes, mostly yes$\}$ support the claim “intermediate reasoning traces are correct”. The final supportive rates are 0.91 for Qwen3 and 0.82 for Qwen2.5. To do a manual validation, we randomly sample and evaluate 100 rollouts from each dataset labeled by each model. We find that Qwen3-32B made 3/200 wrong labels and Qwen2.5 made 15/200 wrong labels. 
    To conclude, most of the intermediate traces are correct and helpful, and SFT on these rollouts can bring improvements.
    \item \textbf{The existing works report similar findings.} \citet{tian2026learningmistakesnegativereasoning} and \citet{li2023turning} find that the negative reasoning traces can help with general reasoning tasks. In particular, \citet{tian2026learningmistakesnegativereasoning} directly apply SFT to complete incorrect trajectories and find consistent OOD gains over positive-only SFT across multiple model scales, attributing the effect to valid intermediate structure, greater trajectory diversity, and reduced overfitting. This provides evidence that these wrong rollouts are still useful.
    \item We conduct further \ac{sft} with \textbf{more unique correct rollouts} and it shows performance similar to SFT w/ RL-tuned (rollout) on general reasoning tasks. Specifically, we adopt the RL-tuned model to generate more rollouts and replace approximately 10,000 wrong rollouts in the existing training dataset. The evaluation result is as in Table~\ref{tab:morecorrect}.
    \begin{table}[ht]
        \centering
        \begin{tabular}{lc}
        \toprule
        \textbf{Model} & \textbf{General Reasoning Average} \\
        \midrule
        RL-tuned & 43.8 \\
        SFT w/RL-tuned & 43.7 \\
        SFT w/RL-tuned (more unique rollouts) & 43.4 \\
        \bottomrule
        \end{tabular}
        \caption{SFT with more unique correct rollouts share similar performance with SFT w/ RL-tuned with total number of rollouts matched.}
        \label{tab:morecorrect}
    \end{table}
It suggests the existing incorrect rollouts are not more harmful than the additional correct rollouts under this setting, in terms of the general reasoning ability.

\end{itemize}

\subsection{Interpretation of the Theoretical Results}\label{app:inter_theory}

We would like to clarify how the following insights derived from our proofs are connected to the experimental findings.

In the experiments, we find that with the same rollout budget, SFT generalizes as well as RL. Moreover, matching the number of rollouts rather than the number of prompts substantially increases the coverage of tail reasoning methods in the SFT dataset.

Our proof explains this phenomenon from an optimization perspective. Both head and tail methods are encoded through outer-product gradient updates, a mechanism closely related to associative memory~\citep{nichani2024transformers}. Under a long-tailed distribution, frequent head methods receive substantially stronger updates, whereas rare tail methods receive fewer updates and are therefore associated with weaker memories and smaller logits. Consequently, gradient descent is more likely to underweight tail methods on OOD inputs. In our setting, RL produces a more balanced distribution of reasoning methods (see Table~\ref{tab:rl_coverage}), resulting in more uniform memory allocation and improved OOD generalization.

\begin{table}[ht]
\centering
\begin{tabular}{lc}
\toprule
\textbf{Setting} & \textbf{Method coverage} \\
\midrule
Rollout during RL & 137 \\
RL-tuned dataset (rollout=8) & 132 \\
RL-tuned dataset (rollout=1) & 101 \\
\bottomrule
\end{tabular}
\caption{Method coverage of the training datasets. RL produces a more balanced rollout dataset during training.}
\label{tab:rl_coverage}
\end{table}


\newpage
\section*{NeurIPS Paper Checklist}

\begin{enumerate}

\item {\bf Claims}
    \item[] Question: Do the main claims made in the abstract and introduction accurately reflect the paper's contributions and scope?
    \item[] Answer: \answerYes{} 
    \item[] Justification: We rethink ``SFT memorizes, RL generalizes'' from the data-centric view, both empirically and theoretically.

\item {\bf Limitations}
    \item[] Question: Does the paper discuss the limitations of the work performed by the authors?
    \item[] Answer: \answerYes{} 
    \item[] Justification: We have discussed the limitations in the conclusion section. We primarily study the data's effect on LLMs, while its effect on Multi-modal models is a limitation.

\item {\bf Theory assumptions and proofs}
    \item[] Question: For each theoretical result, does the paper provide the full set of assumptions and a complete (and correct) proof?
    \item[] Answer: \answerYes{} 
    \item[] Justification: We have stated clearly about the assumptions we used and provided a complete proof.

    \item {\bf Experimental result reproducibility}
    \item[] Question: Does the paper fully disclose all the information needed to reproduce the main experimental results of the paper to the extent that it affects the main claims and/or conclusions of the paper (regardless of whether the code and data are provided or not)?
    \item[] Answer: \answerYes{} 
    \item[] Justification: We have provided the SFT and RL training configurations, the dataset we used during the experiments.

\item {\bf Open access to data and code}
    \item[] Question: Does the paper provide open access to the data and code, with sufficient instructions to faithfully reproduce the main experimental results, as described in supplemental material?
    \item[] Answer: \answerYes{} 
    \item[] Justification: We provide the experimental script in the submission.

\item {\bf Experimental setting/details}
    \item[] Question: Does the paper specify all the training and test details (e.g., data splits, hyperparameters, how they were chosen, type of optimizer) necessary to understand the results?
    \item[] Answer: \answerYes{} 
    \item[] Justification: We specified the data, hyperparameters, optimizers in our setups.

\item {\bf Experiment statistical significance}
    \item[] Question: Does the paper report error bars suitably and correctly defined or other appropriate information about the statistical significance of the experiments?
    \item[] Answer: \answerYes{} 
    \item[] Justification: Our evaluation pipeline contains 5 benchmarks from math domain and 12 benchmarks of general reasoning domain, which contain enough prompts to yield an accurate evaluation results to support our claims.

\item {\bf Experiments compute resources}
    \item[] Question: For each experiment, does the paper provide sufficient information on the computer resources (type of compute workers, memory, time of execution) needed to reproduce the experiments?
    \item[] Answer: \answerYes{} 
    \item[] Justification: In the main paper and Appendix~\ref{app:exp}, we have reported the model size we used, the dataset size, and training configurations, which are enough to derive the compute resources required to reproduce the results.
    
\item {\bf Code of ethics}
    \item[] Question: Does the research conducted in the paper conform, in every respect, with the NeurIPS Code of Ethics \url{https://neurips.cc/public/EthicsGuidelines}?
    \item[] Answer: \answerYes{} 
    \item[] Justification: We conducted the research and conform, in every respect, with the NeurIPS Code of Ethics.

\item {\bf Broader impacts}
    \item[] Question: Does the paper discuss both potential positive societal impacts and negative societal impacts of the work performed?
    \item[] Answer: \answerNA{} 
    \item[] Justification: Our paper mainly focuses on the technical contributions and does not have direct societal impact.
    
\item {\bf Safeguards}
    \item[] Question: Does the paper describe safeguards that have been put in place for responsible release of data or models that have a high risk for misuse (e.g., pre-trained language models, image generators, or scraped datasets)?
    \item[] Answer: \answerNA{} 
    \item[] Justification: Our paper does not pose such risk.

\item {\bf Licenses for existing assets}
    \item[] Question: Are the creators or original owners of assets (e.g., code, data, models), used in the paper, properly credited and are the license and terms of use explicitly mentioned and properly respected?
    \item[] Answer: \answerYes{} 
    \item[] Justification: We have cited the codes, models, datasets we used during the experiments.

\item {\bf New assets}
    \item[] Question: Are new assets introduced in the paper well documented and is the documentation provided alongside the assets?
    \item[] Answer: \answerYes{} 
    \item[] Justification: We submitted the script for training during submission.

\item {\bf Crowdsourcing and research with human subjects}
    \item[] Question: For crowdsourcing experiments and research with human subjects, does the paper include the full text of instructions given to participants and screenshots, if applicable, as well as details about compensation (if any)? 
    \item[] Answer: \answerNA{} 
    \item[] Justification: Our research does not involve human subjects.

\item {\bf Institutional review board (IRB) approvals or equivalent for research with human subjects}
    \item[] Question: Does the paper describe potential risks incurred by study participants, whether such risks were disclosed to the subjects, and whether Institutional Review Board (IRB) approvals (or an equivalent approval/review based on the requirements of your country or institution) were obtained?
    \item[] Answer: \answerNA{} 
    \item[] Justification: Our research does not involve human subjects.

\item {\bf Declaration of LLM usage}
    \item[] Question: Does the paper describe the usage of LLMs if it is an important, original, or non-standard component of the core methods in this research? Note that if the LLM is used only for writing, editing, or formatting purposes and does \emph{not} impact the core methodology, scientific rigor, or originality of the research, declaration is not required.
    \item[] Answer: \answerYes{} 
    \item[] Justification: We prompted Qwen3-32B model to label the math method used in the responses generated by the trained model and we listed the labeled methods in the list in the appendix.

\end{enumerate}

\end{document}